\documentclass[10pt,journal,compsoc]{IEEEtran}
\usepackage[latin9]{inputenc}
\usepackage{color}
\usepackage{verbatim}
\usepackage{booktabs}
\usepackage{amsmath}
\usepackage{amssymb}
\usepackage{graphicx}
\usepackage{subscript}
\usepackage{afterpage}

\makeatletter

\providecommand{\tabularnewline}{\\}

\usepackage{times}
\usepackage{epsfig}
\usepackage{graphicx}
\usepackage{amsmath}
\usepackage{amssymb}
\usepackage{xcolor}

\usepackage[T1]{fontenc}
\usepackage{color}
\usepackage{url}
\usepackage[ruled,vlined]{algorithm2e}
\usepackage{microtype}
\usepackage{multirow}
\usepackage{array}
\usepackage{float}
\usepackage{wrapfig}
\usepackage{ragged2e}
\usepackage{parskip}

\usepackage{mathtools}
\usepackage{widetext}
\usepackage{balance}
\usepackage{xpatch}

\usepackage{url}% simple URL typesetting
\usepackage{booktabs}% professional-quality tables
\usepackage{amsfonts}% blackboard math symbols
\usepackage{nicefrac}% compact symbols for 1/2, etc.
\usepackage{subcaption}
\usepackage{xspace}
\usepackage{comment}

\newcommand{\ie}{\emph{i.e.}\xspace}
\newcommand{\eg}{\emph{e.g.}\xspace}

\newcommand{\Eq}{Eq.\xspace}
\newcommand{\Fig}{Fig.\xspace}

\newcommand{\Tab}{Tab.\xspace}

\newcommand{\etc}{\emph{etc.}\xspace}

\newcommand{\revise}[1]{\textcolor{black}{#1}}

\usepackage{wrapfig}
\usepackage[normalem]{ulem}
\usepackage[breaklinks=true,colorlinks=false,bookmarks=false]{hyperref}
\ifCLASSOPTIONcompsoc
  \usepackage[nocompress]{cite}
\else
  \usepackage{cite}
\fi

\usepackage{enumitem}
\setlist[itemize]{leftmargin=15pt}

\makeatother

\begin{document}
\setlength{\parskip}{0pt}
\bstctlcite{reflib}
\title{How Trustworthy are 
%the Existing 
% the word existing in the sentence sounds like we are bagging the existing work
Performance Evaluations for Basic Vision Tasks?}
\author{Tran Thien Dat Nguyen$^{*}$, Hamid Rezatofighi$^{*}$, Ba-Ngu Vo, Ba-Tuong Vo, Silvio Savarese, and Ian Reid 
\thanks{$*$ authors have equal contribution.} 
\thanks{T.T.D. Nguyen, B.-N. Vo and B.-T. Vo are with School of Electrical Engineering, Computing and Mathematical Sciences, Curtin University, Australia (emails: t.nguyen1@curtin.edu.au, $\{$ba-ngu.vo, ba-tuong.vo$\}$@curtin.edu.au.). H. Rezatofighi is with Faculty of Information Technology, Monash University, Australia (email: hamid.rezatofighi@monash.edu). S. Savarese is with Stanford University, USA (email: ssilvio@stanford.edu). I. Reid is with School of Computer Science, University of Adelaide, Australia (email: ian.reid@adelaide.edu.au).\protect}}
\makeatletter 
\let\@oldmaketitle\@maketitle
\renewcommand{\@maketitle}{\@oldmaketitle
\setcounter{figure}{0}    
\centering \includegraphics[width=\textwidth] 
{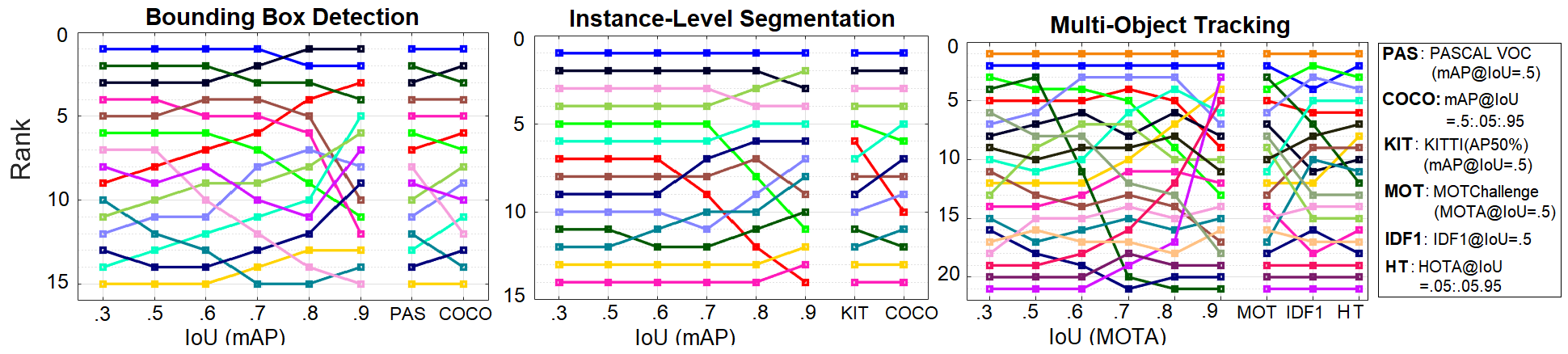} \captionof{figure}{Rankings of some established algorithms (details given in \Fig \ref{fig:real_ranks}) on public datasets and challenge benchmarks, according to various (performance) criteria. The Bounding Box Detection and Instance-Level Segmentation tasks are evaluated on the COCO validation dataset, while the Multi-Object Tracking task is evaluated on the MOT17 training dataset. For a given task, each ranked algorithm is represented by a unique color. The plots show the ranking varies across IoU thresholds, with some algorithms switching from high to  low ranks, and vice-versa. The algorithms are also ranked differently on different benchmarks. Such ranking variability begs the question of how trustworthy are the evaluations by these criteria. %on the trustworthiness of these criteria.
}\label{fig:teaser}\vspace{-10pt}}
\IEEEtitleabstractindextext{
\begin{abstract}
This paper examines performance evaluation criteria for basic vision tasks involving sets of objects namely, object detection, instance-level segmentation and multi-object tracking. The rankings of algorithms by an existing criterion can fluctuate with different choices of parameters, \eg Intersection over Union (IoU) threshold, making their evaluations unreliable. More importantly, there is no means to verify whether we can trust the evaluations of a criterion. This work suggests a notion of trustworthiness for performance criteria, which requires (i) robustness to parameters for reliability, (ii) contextual meaningfulness in sanity tests, and (iii) consistency with mathematical requirements such as the metric properties. We observe that these requirements were overlooked by many widely-used criteria, and explore alternative criteria using metrics for sets of shapes. We also assess all these criteria based on the suggested requirements for  trustworthiness. 
\end{abstract}
}
\maketitle

\section{Introduction }

\IEEEPARstart{I}{n} addition to technological developments, performance
evaluation is indispensable to the advancement of machine vision.
It is difficult to envisage how improvements or advances can be demonstrated
without performance evaluation. In this work we restrict ourselves
to \textit{basic vision tasks} involving sets of objects, namely object
detection, instance-level segmentation, and multi-object tracking,
where several benchmarks have been proposed to evaluate their performance, see for example~\cite{Everingham:2012:VOC,coco,cordts2016cityscapes,MOTChallenge:arxiv:2015,MOTChallenge:arxiv:2016,dendorfer2019cvpr19,Geiger:2012:AWR,002,bernardin2008evaluating,012,borji2015salient}. 

% \revise{Further, in certain applications, practitioners need the performance criteria to be sensitive to certain aspects of the evaluation, \eg localization or cardinality accuracy. However, under the scope of this work we focus on the evaluation tasks, where the users do not give specific reference to any aspects of the performance evaluation.}

Given the importance of performance evaluation, its consistency and
rigor have not received proportionate attention in computer vision.
The standard practice is to rank the solutions according to certain
criteria based on their outputs or \textit{predictions/estimates}
on prescribed datasets \cite{Everingham:2012:VOC,coco,MOTChallenge:arxiv:2015}.
In general, these criteria aim to capture the similarities/dissimilarities
between the \textit{predictions} and prescribed \textit{references},
with higher similarities (lower dissimilarities) indicating better
performance. \revise{In practice, performance criteria are chosen, largely,
via intuition (\eg see \cite{bernardin2008evaluating,coco,kirillov2019panoptic}),
while formal consideration on fairness or consistency is overlooked.}

While the widely-used performance criteria for basic vision tasks
are important to the progress of the field, there are a number of
drawbacks. 
\begin{itemize}
\item Firstly, the rankings by these criteria \revise{may fluctuate} with the choice of parameters
(\eg IoU-thresholds as shown in \Fig \ref{fig:teaser}). Hence, their
evaluations are dubious because tuning of parameters could shift
low-ranking predictions to high-ranking ones, and vice-versa. Note
that the widely-used 0.5 IoU-threshold is rather arbitrary, and there
are no formal justifications for its preference over other choices
\cite{Everingham:2012:VOC,dollar2011pedestrian,coco}. 
\item Secondly, while these criteria are formulated based on intuition and
intent, there is no principled framework to assess how meaningful
their evaluations actually are, or how well they capture the intent
of the evaluation exercise. \vspace{3pt}
\item Thirdly, in basic vision tasks, exact or ground truths are not available
as references, and it is assumed that high similarities with approximate
truths (acquired \eg via annotations) imply high similarities with
ground truths. However, this is not the case as demonstrated in Section \ref{subsec:Mathematical-Consistency} (Figs. \ref{fig:cascaded_boxes-1},
\ref{fig:TAP_exp},  \ref{fig:apprx_traditional}).
Consequently, there is no assurance that high-ranking predictions
actually perform better than low-ranking ones, which undermines the
whole purpose of performance evaluation. 
\end{itemize}
In view of such drawbacks, the ensuing scientific questions are: what
would a trustworthy performance criterion entail, and how to formulate
trustworthy performance evaluation strategies?

This paper suggests a formalism for the trustworthiness of performance
criteria, and provides an independent assessment of some widely-used
criteria in basic computer vision tasks together with criteria borrowed
from point pattern theory. \revise{In particular, this formalism is stipulated as a set of guidelines, whereby a trustworthy performance criteria is required to be:} 

\begin{enumerate}[label=\textit{\textbf{(\roman*)}}]   
\item robust to variations in parameters for reliability;    
\item meaningful in \textit{sanity tests} - systematically constructed test scenarios with pre-determined rankings to capture the intent of the evaluation;  
\item \revise{mathematically consistent - suitable analytical properties \eg metric properties.}
\end{enumerate}Noting that the above requirements were overlooked in widely-used
criteria, such as F1, log-Average Miss Rate (log-AMR), mean Average
Precision, Multi-Object Tracking Accuracy (MOTA), IDF1, and Higher Order Tracking Accuracy (HOTA), we explore
some alternative performance criteria for object detection, instance-level
segmentation, and multi-object tracking. These alternative criteria
are \emph{(mathematical) metrics} for sets of shapes, which integrate
point pattern metrics with shape metrics. We also assess the trustworthiness
of these metrics (and the above criteria) via the suggested requirements. 

\section{Related Work\label{sec:related_work}}

Several performance evaluation methods have been proposed for the
basic vision tasks of object detection, instance-level segmentation,
and multi-object tracking.

\textbf{Intersection over Union (IoU) and Generalized-IoU (GIoU)}
is the most commonly used family of similarity measures between two
arbitrary shapes. IoU captures the similarity of the objects under
comparison by a normalized measure based on the overlap in areas (or
volumes) of the regions they occupy. This construction makes IoU scale-invariant,
and hence the defacto base-similarity measure of many performance
criteria. However, IoU is insensitive to the shape and proximity of
non-overlapping shapes. To this end, a generalization that covers
non-overlapping shapes, namely Generalized IoU (GIoU), was proposed
in \cite{005}. 

\textbf{Performance evaluations for object detection and instance-level
segmentation} consider the similarity (or dissimilarity) between the
reference and predicted sets of bounding boxes or masks. Popular performance
criteria are based on the notion of \textit{true positives}, determined
by matching predictions with references such that the IoU (or GIoU)
value between them is larger than a specified threshold, usually 0.5
\cite{Everingham:2012:VOC,dendorfer2019cvpr19,Geiger:2012:AWR}. Note
that, the subset of true positives is dependent on the choice of thresholds.
The (subset of) \textit{false positives} is then defined to be the
prediction set excluding all true positives. Similarly, the (subset
of) \textit{false negatives} (or misses) is the truth set excluding
all true positives.

\emph{F1-score \cite{borji2015salient}} \revise{is one of the simplest similarity measure for object detections}, where the predictions
are sets of bounding box coordinates with no confidence scores nor
category labels, \eg salient object detection \cite{borji2015salient}.
F-measure captures the similarity with the harmonic mean of precision
(the ratio of true positives to predictions) and recall (the ratio
of true positives to truths). \revise{Specifically, let $FP$ be the number of false positives, $FN$ the number of false negatives
and $TP$ the number of true positives. Then the precision $(P)$, recall
$(R)$ and F1 are are defined respectively as 
\[
P=\frac{TP}{TP+FP}\textrm{  ,  }R=\frac{TP}{TP+FN}\textrm{  ,  } F1=2\times\frac{P\times R}{P+R}
\]}

\emph{Average Precision (AP) and mean AP (mAP) \cite{Everingham:2012:VOC,coco}}
are perhaps the most popular performance criteria for single-category
and multi-category label object detection/instance-segmentation, respectively.
When predictions include confidence scores, true positives are determined
by a non-optimal greedy assignment strategy that matches (with references)
those with higher confidence scores first \cite{Everingham:2012:VOC,coco}.
Precision and recall can be expressed as a curve generated from different
confidence threshold values. \revise{Let $p$ denote the precision in order of confidence scores, and
$r$ denote the recall. Then, the AP score is defined as the area under the $p(r)$
curve, \ie 
\begin{equation*}
AP=\int_{0}^{1}p(r)dr.\label{eq:exact_AP}
\end{equation*}} 
\revise{In practice, this area is approximated by summing over a finite set of recall points \cite{Everingham:2012:VOC,coco}. Given $N$ selected recall points $r_{1},...,r_{N}$
such that $r_{n}<r_{n+1}$, $\forall n<N$, the approximate AP score
is: 
\begin{equation*}
\widetilde{AP}=\sum_{n=1}^{N-1}(r_{n+1}-r_{n})\widetilde{p}(r_{n+1}),
\end{equation*}
where $\widetilde{p}(r)$ is the approximation of $p(r)$ such that
$\widetilde{p}(r)=\max_{\widetilde{r}\geq r}p(\widetilde{r})$.}

For multi-category label predictions,
the mean AP (mAP) over all categories is used. \revise{Conversely,  the  MS  COCO Benchmark challenge \cite{coco} averages mAP across multiple IoU thresholds to reward detector with higher localization accuracy}

\emph{Log-average miss rate (log-AMR) \cite{dollar2011pedestrian}}
is another popular performance criterion for object detection. Given
the reference-prediction matches as per AP, the miss rate (MR) is
plotted against the false positives per image (FPPI) rate. Similar
to AP, log-AMR approximates the area under the MR-FPPI curve from
a finite number of samples. \revise{For a miss rate $m$ and FPPI rate $f$
(sorted in the order of the prediction score), the log-AMR is given by
\begin{equation*}
AMR=\exp\left(\frac{1}{N}\sum_{n=1}^{N}\ln\left(m(f_{n})\right)\right),
\end{equation*}
where $f_{1},...,f_{N}$ are the sampled FPPI rates.}

\textbf{Performance evaluations for multi-object tracking} consider
the similarity/disimilarity between sets of reference and predicted
tracks. Performance criteria usually rely on IoU or Euclidean distance
to match reference tracks with predicted tracks, at each time step
\cite{bernardin2008evaluating}, or on the entire duration \cite{002}.
Other performance criteria such as trajectories-based measures \cite{003},
configuration distance and purity measure \cite{012}, or global mismatch
error \cite{013} were also developed based on similar constructions. A criterion based on high order matching is also
recently proposed in \cite{HOTA}.%

\emph{Multi-Object Tracking Accuracy (MOTA) \cite{bernardin2008evaluating}}
is based on pairing, at each frame, reference and predicted objects
within a separation threshold. From this pairing, the mismatch error
that captures label inconsistency is the total number of times that
track identities are switched. The MOTA score is defined as one minus
the normalized (by the total number of reference tracks) sum of mismatch
error, and the total number (over all frames) of false positives and
false negatives. \revise{Specifically, given $FP_{t}$, $FN_{t}$, $IDSW_{t}$ and $GT_{t}$, which
are respectively the number of false positives, false negatives, ID
switches and ground truth track instances at time $t$, the MOTA score
is given by \cite{bernardin2008evaluating}: 
\begin{equation*}
MOTA=1-\frac{\sum_{t}FP_{t}+FN_{t}+IDSW_{t}}{\sum_{t}GT_{t}}.
\end{equation*}}

\emph{IDF1 \cite{002}} is based on pairing reference tracks to predicted
tracks so as to minimize the sum of, false positives and false negatives
from each pair, for a given distance/IoU threshold. Dummy trajectories
are used to account for the cardinality mismatch between the reference
and predicted sets. From the optimal pairing, the IDPrecision, IDRecall,
and subsequently IDF1 scores are given by the total number of false
positives and false negatives of the pairs. \revise{The IDF1 score is defined as:
\begin{equation*}
IDF1=\frac{2IDTP}{2IDTP+IDFP+IDFN},
\end{equation*}
where $IDTP$, $IDFP$ and $IDFN$ are respectively the numbers of
true positive ID, false positive ID and false negative ID.}

\emph{Higher Order Tracking Accuracy (HOTA) \cite{HOTA}} is designed
to evaluate the long-term high-order association between predictions
and references. In particular, HOTA measures the degree of alignment
between trajectories and matching detections given the matches. Relying on thresholds to declare
matches, the score is first evaluated over a set of localization thresholds $\alpha$,
\revise{
\begin{equation*}
HOTA^{(\alpha)}=\sqrt{\frac{\sum_{c\in\{TP\}}\mathcal{A}(c)}{|TP|+|FN|+|FP|}},
\end{equation*}
where:
\[
\mathcal{A}(c)=\frac{|TPA(c)|}{|TPA(c)|+|FNA(c)|+|FPA(c)|};
\]
$TP$, $FN$, and $FP$ are, respectively, the sets of true positives, false negatives
and false positives for all predicted and ground truth instances; $TPA(c)$, $FNA(c)$, and $FPA(c)$ are, respectively, the sets of true positive associations, false negative associations and false
positive associations for a given $c$, see \cite{HOTA} for details.}

% \begin{align*}
% TPA(c) & =\{k\},\\
%  & k\in\{TP|pID(k)=pID(c)\land gID(k)=gID(c)\},\\
% FNA(c) & =\{k\},\\
%  & k\in\{TP|pID(k)\neq pID(c)\land gID(k)=gID(c)\}\\
%  & \cup\{FN|gID(k)=gID(c)\},\\
% FPA(c) & =\{k\},\\
%  & k\in\{TP|pID(k)=pID(c)\land gID(k)\neq gID(c)\}\\
%  & \cup\{FP|pID(k)=pID(c)\},
% \end{align*}

% $pID(c)$ and $gID(c)$ are respectively the IDs of the prediction
% and ground truth instances given a match $c$.} 
The final score is then obtained
via marginalizing out the thresholds. In this work, we use the term ``HOTA'' to refer to the thresholding version of the measure while the marginalized score will be treated independently for consistent comparison with other performance criteria. 
% \emph{Higher Order Tracking Accuracy (HOTA) \cite{HOTA}} is designed
% to evaluate the long-term high-order association between predictions
% and references. In particular, HOTA measures the degree of alignment
% between trajectories and matching detections. The final score includes
% the averaged amount of alignment over all matching detections and
% the penalty of unmatched detections. Relying on thresholds to declare
% matches, the score is first evaluated over a set of localization thresholds, then the final score is obtained
% via marginalizing out the thresholds.

\section{Guidelines for Performance Criteria\label{sec:guideline}}

A performance criterion quantifies (by a numerical value) the similarity/dissimilarity
of the output of an algorithm to a nominal reference. For basic vision
tasks, namely detection, instance-level segmentation and multi-object
tracking, our interest lies not only in the dissimilarity between
two shapes, but dissimilarity between two (finite) sets of shapes.
This dissimilarity measure can be constructed in many ways, from hand-crafted
criteria based on intuition to using actual human assessments, each
with its own merits and drawbacks. Regardless of its conception, the
fundamental question is: how can we trust that a performance criterion
does what we expect it to do?

This section attempts to answer the above question by suggesting guidelines
for certifying trustworthiness of criteria based on the notions of
\emph{reliability}, \emph{meaningfulness}, and \emph{mathematical
consistency}. Specifically, a trustworthy criterion must be reliable,
meaningful and mathematically consistent. In the following, we discuss
the meaning and rationale of these concepts. 

\begin{figure}[t]
\centering{}\includegraphics[width=0.45\textwidth]{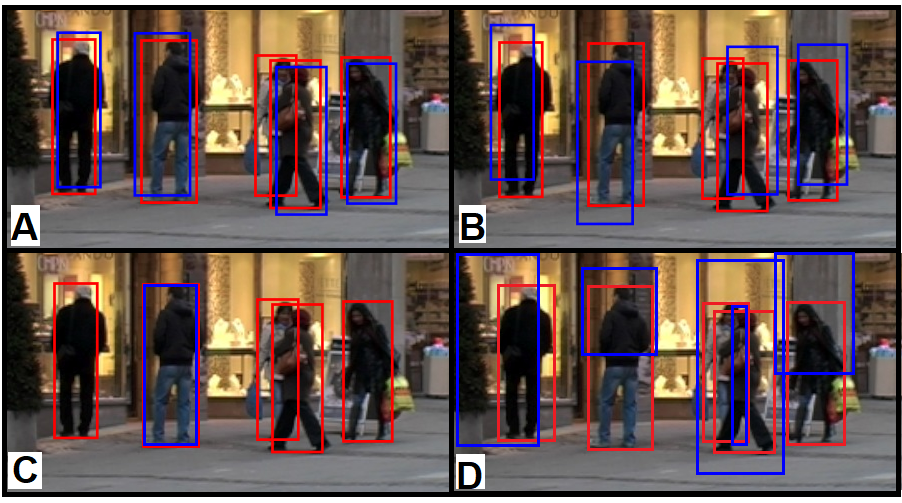}
\caption{\revise{People detection, with red/blue boxes representing truths/predictions.
A and B detect 4 out of 5 people, with 0.8 and 0.55 IoU per person,
respectively. C detects one person perfectly out of 5 people. D detects
5 out of 5 people but with an IoU of 0.3 per person. For criteria
based on IoU thresholding, \eg F1-score: (i) A is indistinguishable
from B, if the commonly used IoU threshold of 0.5 is applied;
(ii) C can rank above A and B if a high IoU threshold (above 0.8) is chosen; (iii) D is the worst detector at IoU threshold above 0.3 but becomes the best detector if an IoU threshold below 0.3 is
selected.} \label{fig:sanity_test_sample_1} }
\end{figure}

\subsection{Reliability\label{subsec:Reliability}}

\revise{The rankings produced by a performance criterion should be robust to variations of the parameters, \eg the IoU thresholds in \Fig \ref{fig:teaser}. Intuitively, a criterion whose rankings are independent of the parameters is more robust than one whose rankings wildly fluctuate with variation of the parameters. More specifically, for a reliable criterion we expect that a small change in parameter values will not result in a drastic change in rankings.  For example, in \Fig \ref{fig:sanity_test_sample_1}
for an IoU threshold below 0.8, detector C has the worst performance among A, B, and C. However, when the threshold is above 0.8 (no matter how small above 0.8), C becomes the best detector. Similarly, if a threshold above 0.3 is chosen, D is the worst detector. However, D becomes the best detector when the threshold is below 0.3 (no matter how small below 0.3). Such sensitivity may allow dubious promotion of certain solutions via
parameter tuning.} Averaging the evaluation score over a set of thresholds (\eg mAP implementation in COCO multi-object detection challenge~\cite{coco})
may lead to even larger ranking discrepancies for criteria with higher
parameter sensitivity, although averaging the score over the a wider range of thresholds seems to improve the ranking performance (as indicated by our experiment). However, the problem with this strategy is its sensitivity to how the averaging is implemented, \ie the parameters of the averaging implementation.

\subsection{Meaningfulness\label{subsec:Meaningfulness}}

\begin{figure}[b]
\centering{}\includegraphics[width=0.45\textwidth]{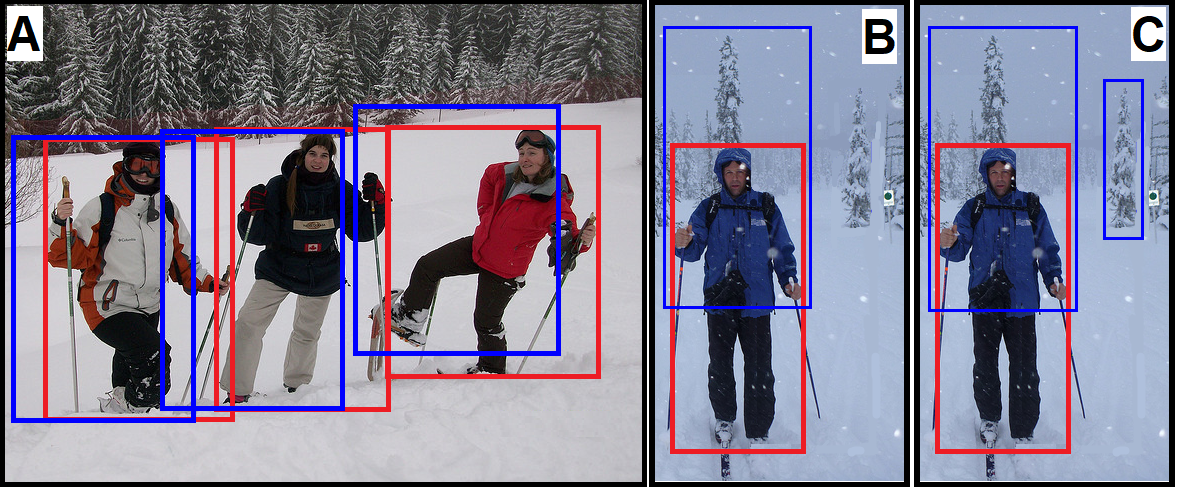}
\caption{Scene A: a correct prediction that there are 3 people in the scene,
with an accuracy of 0.75 IoU per person (red/blue boxes represent
truths/predictions). Scene B: A correct prediction that there is only
person in the scene, with 0.3 IoU accuracy. Scene C is formed by adding
a spurious detection to B. Any meaningful criterion should rank A
above B, and B above C.\label{fig:OSPA_intuition-1} }
\end{figure}

Reliability alone does not guarantee that a criterion is \emph{meaningful},
\ie captures the intent of the performance evaluation exercise. Consider
\eg the people detection task in \Fig \ref{fig:OSPA_intuition-1},
where: detector A correctly detected all 3 people with a small error
for each person; detector B correctly detected the only person but
incurs a large error, and detector C has the same output as B with
an additional spurious positive. Unequivocally, the detection performance
of A is better than B, which, in turn, is better than C. Any performance
criteria that proclaim otherwise are not meaningful.

Given that there are no analytical means in the computer vision literature
for ensuring meaningfulness of performance criteria, the best option
is to consider \emph{experimental validation}--a common practice
in the empirical sciences. This approach tests the criteria on a series
of scenarios (real or simulated) to verify corroboration with the
intent of the performance evaluation. The better the criteria fare,
and the more extensive the test scenarios, the more trust we have
in their meaningfulness when applied to real data.

A popular experimental validation strategy is to use humans to evaluate
whether the performance criteria are meaningful \cite{trackingthetracker,kirillov2019panoptic}.
However, this practice inherently suffers from a number of drawbacks.
Firstly, human evaluation is not scalable, and can only be applied
to evaluate a small number of scenarios. Hence, extensive validation
on complex scenarios involving multiple error sources, large number
of objects, and large datasets is not feasible. Secondly, human evaluation
is subjective and invariably leads to inconsistencies due to differences
in expertise, experience and capability. For example, in object detection
one prediction set may contain more false positives/negatives while
another set has more severe localization error. In this case, human
judgment can be subjective and assessments by different humans can
be inconsistent with one another. Finally, humans are not capable
of differentiating small differences in performance, and thus unable
to assess the granularity of the criteria.

\begin{figure*}[ht]
\begin{centering}
\includegraphics[width=0.87\textwidth]{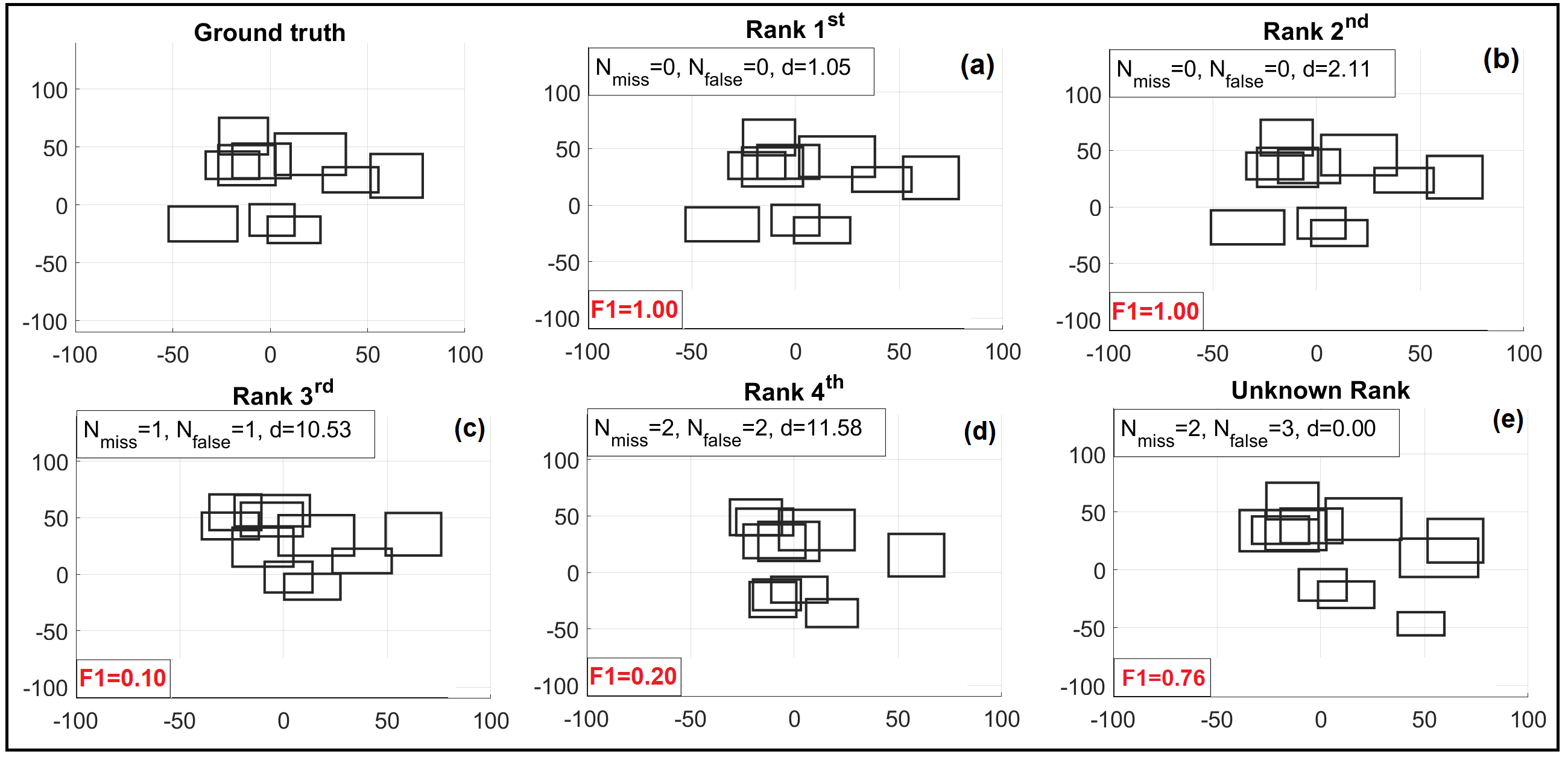}
\par\end{centering}
\caption{Visual demonstration on the concept of parameters characterizing the
perturbation in the sanity test where N\protect\textsubscript{miss},
N\protect\textsubscript{false} are the number of missed
and false objects, d is the dislocation of centroid (Euclidean
distance); F1 is the F1 score at IoU=0.5. Prediction (\textbf{a})
is very competitive compared to prediction (\textbf{b}) and F1 criterion
cannot distinguish their performances. It is uncertain to tell if
(\textbf{c}) is better than (\textbf{d}) by visual inspection but
it is clear via parameters characterizing the perturbation; however,
F1 criterion produces incorrect ranking order for this pair. It is
uncertain to rank (\textbf{e}) among other predictions via either
visualization or parameters characterizing the perturbation hence
we need to solely rely on performance criteria to rank the predictions.\label{fig:sanity_demo}}
\end{figure*}

\subsubsection{Santity Testing}

\emph{Our suggestion for assessing meaningfulness} is to systematically
construct a series of sanity tests, consisting of scenarios with pre-determined
prediction rankings, based on the intent of the performance evaluation
exercise (\eg the edge-cases in \Fig \ref{fig:OSPA_intuition-1}),
and verify whether the criterion's rankings corroborate the pre-determined
rankings. A criterion that does not corroborate the pre-determined
rankings cannot provide meaningful evaluation. On the other hand,
the better the corroboration with the pre-determined rankings, the
more confidence/trust we have in its ability to provide meaningful
performance evaluation in practice. This strategy allows extensive
validation involving multiple error sources, large number of objects,
and large dataset. 

Suppose that the sources of errors for the application can be identified,
\eg false negatives/positives, location/shape errors, \etc.
\begin{itemize}
\item \emph{First}, we generate/use a number of reference sets based on
typical data from the application.\vspace{3pt}
\item \emph{Second}, we generate a number of prediction sets with pre-determined
performance ranking by perturbing the reference sets with simulated
errors. Predictions generated from small perturbations are ranked
higher than those generated from large perturbations. A prediction
with lower rank can be generated from a given prediction by perturbing
it with additional sources of error, see \eg scenarios B and C in
\Fig \ref{fig:OSPA_intuition-1}. This strategy enables the generation
of complex scenarios with a combination of error sources and large
number of objects, where the pre-determined rankings might not be
obvious to the human eye, thereby enabling extensive validation not
achievable with human evaluation. \vspace{3pt}
\item \emph{Third}, we rank the generated predictions according to the criterion
under investigation, and determine how meaningful it is by measuring
the ranking discrepancy or error (with respect to the pre-determined
rankings). For a given a collection of predictions, we measure the
ranking error of a criterion by the \emph{Kendall-tau} distance between
its own ranking and the pre-determined ranking. \revise{This distance (also
called bubble-sort distance), is a well-established (mathematical)
metric for measuring dissimilarity between two rankings by counting
the number of pairwise disagreements between two ranking lists \cite{kendall_tau_dist} and has been widely used in the literature (see \cite{whyrarnkings2018Maier,multires2012Liu,imgretarget2010Rubinstein,ObjAssess2014Hsu,Norefret2016Ma} for examples)}.
The smaller the ranking error, the better the criterion corroborates
the intent of the performance evaluation\footnote{\revise{Other distances such as Manhattan distance and Spearman correlation (in distance form) also show almost identical behaviors to the Kendall-tau distance in our experiments}}.
\end{itemize}
\Fig \ref{fig:sanity_demo} shows a single trial of the proposed
sanity test. The performance of predictions (a) and (b) are almost
impossible for humans to distinguish via visual inspection. In contrast,
from the parameters characterizing the perturbations (the dislocation
magnitude that the experimenter prescribes), it is clear that prediction
(a) is better than (b). Similarly, without any context, it is not
clear how we would rank the performance of predictions (c) and (d)
due to the complexity of the scene. However, based on the prescribed
magnitude of dislocation, number of misses, falses, it is clear that (c)
is better than (d). If a performance criterion corroborates well with
a series of predetermined rankings, we would have more trust in its
ability to capture the intent of the evaluation in ambiguous 
scenarios such as (e), where the ordering of the perturbation
parameters provide no information to rank the predictions.

We stress that no performance criteria in
the literature are guaranteed to provide meaningful evaluations in
general (whether real or simulated). Moreover, there are no analytical
implements nor frameworks to assess how meaningful criteria are. Our
proposed methodology offers a sensible and pragmatic way to address
the meaningfulness of criteria in the context of performance evaluation.

\subsection{Mathematical Consistency\label{subsec:Mathematical-Consistency}}

\begin{figure}[h!]
\centering{}\includegraphics[width=0.18\textwidth]{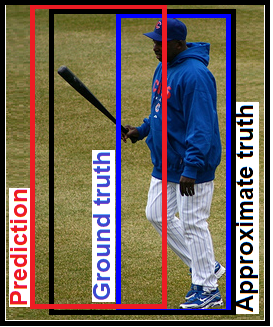}
\caption{For an IoU threshold of 0.5, the Prediction is ``closest'' to the
Approximate truth ($F1=1$), which is ``closest'' to Ground truth
($F1=1$). Thus, the Prediction should be ``close'' to Ground truth,
but it is as ``far'' as possible from Ground truth ($F1=0$)!\label{fig:cascaded_boxes-1} }
\end{figure}

Relying purely on intuitive indicators is not adequate for rigorous
scientific performance evaluation. This is especially true in basic
vision problems, where \textit{ground truths} are not available (except
for simulated data) and only \textit{approximate truths} can be used.
Keeping in mind that approximate truths are acquired through some
measurement processes, \eg manual annotation (which is rather subjective)
and differ from the ground truth, a performance criterion only captures
the similarity/dissimilarity between the predictions and approximate
truths. It is implicitly assumed that the similarity/dissimilarity
measure is \emph{mathematically consistent} in the following sense:
suppose that the approximate truth is \textquotedblleft close\textquotedblright{}
(\ie highly similar) to the ground truth, then being \textquotedblleft close\textquotedblright{}
to the approximate truth means being \textquotedblleft close\textquotedblright{}
to the ground truth. However, this assumption does not necessarily
hold even for similarity/dissimilarity between two shapes, let alone
two sets of shapes, as illustrated in \Fig \ref{fig:cascaded_boxes-1}.
According to the F1 criteria, even though the prediction is \textquotedblleft closest\textquotedblright{}
(indicated by the best F1 score) to the approximate truth, which in
turn is \textquotedblleft closest\textquotedblright{} to the ground
truth, it bears no similarity with the ground truth whatsoever (zero
F1 score). Thus, without mathematical consistency, the best possible
predictions according to a criterion could be the furthest (most dissimilar)
from the truth.

\begin{figure}
\begin{centering}
\includegraphics[width=0.45\textwidth]{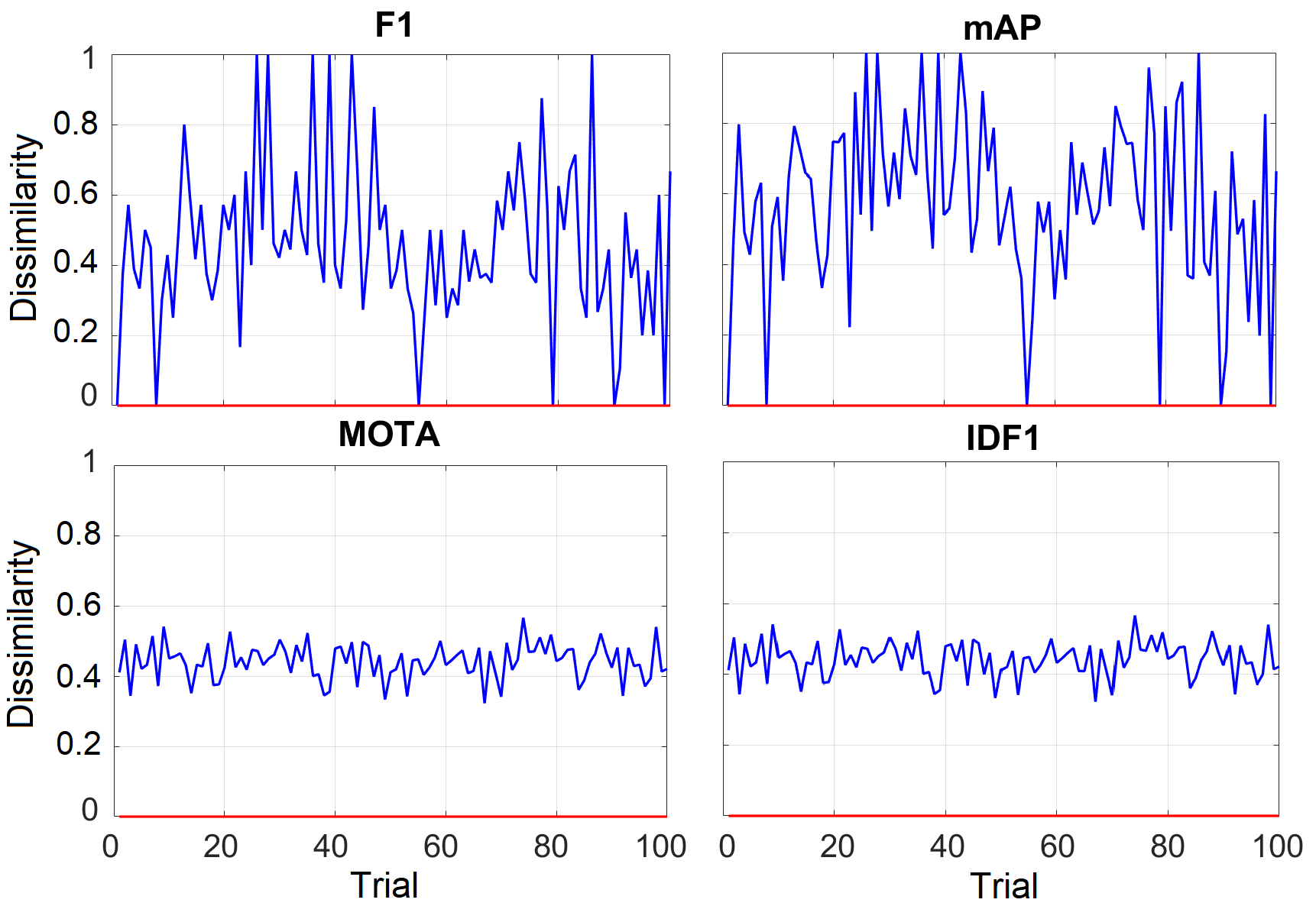}
\par\end{centering}
\caption{\revise{\textbf{(Red)} Dissimilarity between ground truth and approximate
truth. \textbf{(Blue)} Dissimilarity between approximate truth
and prediction. Both are normalized against the dissimilarity between ground truth and prediction. Note that the normalized the dissimilarity between ground truth and prediction is 1. Hence for a consistent criterion the blue lines should be close to 1 (assuming the red line is close to 0).}
%but this is not the case even though the red line is exactly zero. 
\label{fig:TAP_exp}}
\vspace{-0.3cm}
\end{figure}

To further illustrate the role of mathematical consistency in
prediction errors for basic vision tasks, we simulated ground
truths, and approximate truths/predictions by perturbing ground truths
with small/large random dislocations, and consider the $\text{F1}$
and $\text{mAP}$ dissimilarity measures, \ie $1-\text{F1}$ and
$1-\text{mAP}$ (the mAP score is calculated by assuming there is
only one class, and the confidence score is 0.9 for all predictions). \revise{The red curve in \Fig \ref{fig:TAP_exp} indicates zero dissimilarity between ground truth and approximate truth, while
the blue curve shows that the normalized (prediction) error measured from approximate truth is not close to 1 (the normalized prediction error measured from ground truth). This demonstrates large discrepancies between the (prediction) errors measured from ground truth and that measured from approximate truth, even though
there is no dissimilarity between these truths.}

% large discrepancies between the (prediction) errors measured from ground truth and approximate truth, even though
% there is no dissimilarity between ground truth and approximate truth.}
% The dissimilarity between ground truth and approximate truth, and the dissimilarity between approximate truth and prediction, both normalized against the dissimilarity between ground truth and prediction, are plotted in \Fig \ref{fig:TAP_exp}

\begin{figure*}
\centering{}\includegraphics[width=0.95\textwidth]{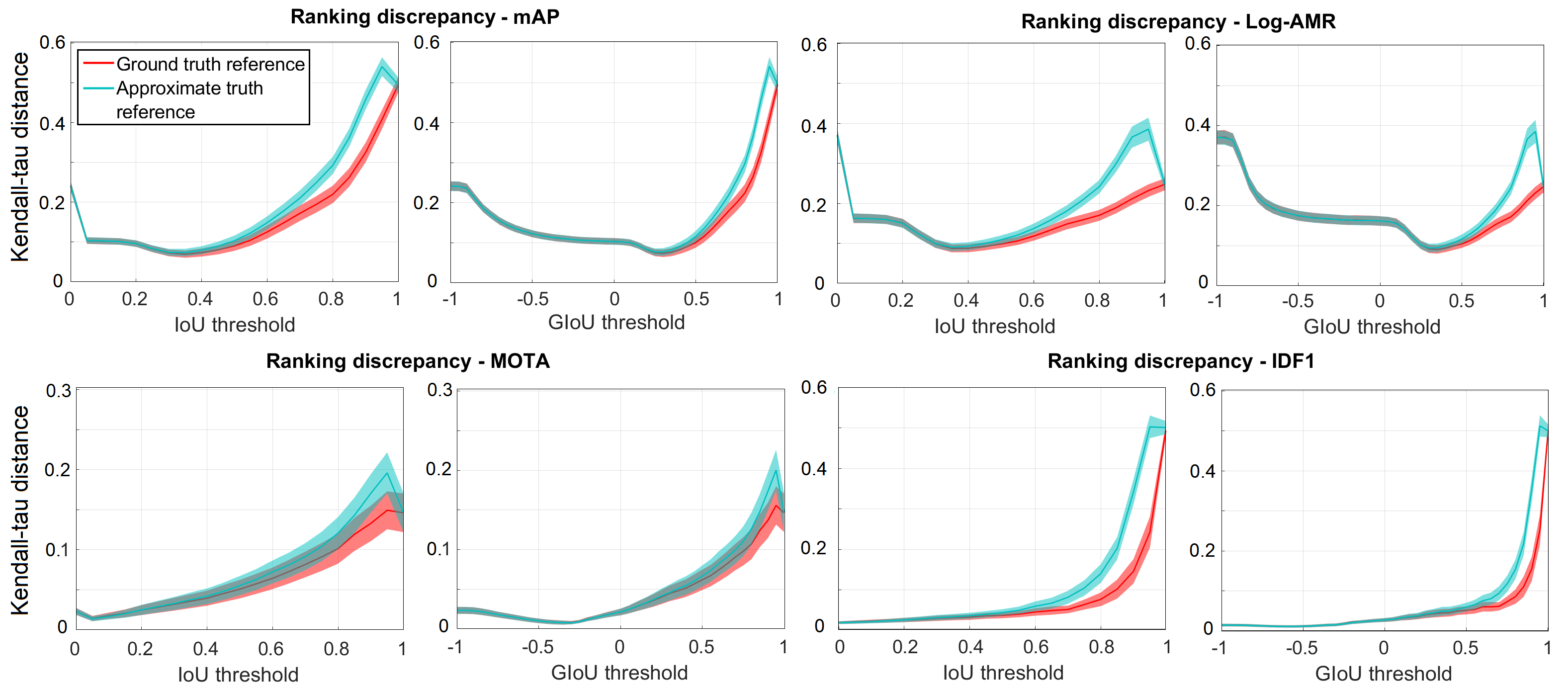}
\caption{Monte Carlo means of normalized Kendall-tau ranking errors (from the true ranking)
for various traditional criteria at different thresholds with ground
truth and approximate truth reference sets,
in \textbf{detection test (top row)} and \textbf{tracking test (bottom
row)}. Shaded area around each curve indicates 0.2-sigma bound. \label{fig:apprx_traditional}}
\end{figure*}

To illustrate the effect of mathematical consistency on performance
rankings, we generate the ground truth and prediction sets for the
multi-class multi-object detection and multi-object tracking tests
(by introducing perturbations to the ground truth). The true ranking
order of the predictions are known (via the severity of the perturbations).
Sets of approximate truth are also generated from the ground truth
sets by perturbing the bounding boxes with small random dislocations
\revise{(the minimum allowable IoU index between ground truth and approximate
truth is 90\%)}. \Fig \ref{fig:apprx_traditional} plots the normalized
Kendall-tau distance between the true ranking vectors and the evaluated
ranking vectors using ground truth references and approximate truth
references, for a number of traditional criteria. Observe that at high IoU and GIoU thresholds, the (Kendall-tau) ranking
error is substantially higher with approximate truth reference compared
to ground truth reference. Thus, in practice where only the approximate
truths are available,  mathematically inconsistent criteria may
not provide fair evaluations because being close to the approximate truth does not mean much. 

One way to ensure mathematical consistency is to consider \textit{(mathematical)
metrics}--dissimilarity measures with certain mathematical properties.
Specifically, a function $d:\mathcal{S\times}\mathcal{S}\rightarrow[0,\infty)$
is called a metric (or distance function) on the space $\mathcal{S}$,
if for all $x,y,z\in\mathcal{S}$ it satisfies: \vspace{3pt}
\begin{enumerate}
\item (Identity) $d(x,y)=0$ if and only if $x=y$ ; \vspace{3pt}
\item (Symmetry) $d(x,y)=d(y,x)$ ; \vspace{3pt}
\item (Triangle inequality) $d(x,z)\leq d(x,y)+d(y,z)$.\vspace{3pt}
\end{enumerate}
The \emph{triangle inequality} warrants mathematical consistency,
\ie if the prediction $z$ is ``close'' to the approximate truth
$y$, and assuming that the approximate truth $y$ is ``close''
to the ground truth $x$, then the triangle inequality asserts that
the prediction $z$ is also ``close'' to the ground truth $x$.
Violating the triangle inequality results in the inconsistencies of
the performance criteria depicted in \Fig \ref{fig:TAP_exp}. It
is also important to note that without the \emph{Identity} property,
imperfect predictions can have the same rank as the perfect prediction.
Violation of this property can result in the inability to distinguish
relatively clear performance differences, as illustrated in our earlier
discussion on \Fig \ref{fig:sanity_test_sample_1}.

\textbf{\emph{Remark}}: All criteria discussed in Section \ref{sec:related_work}
are not mathematically consistent because they rely on thresholding
the base-similarity/dissimilarity to determine the number of true
positives (that solely define the criteria). In fact, (the dissimilarity
forms of) these criteria violate the \emph{Triangle Inequality} and
\emph{Identity} property which is shown in the following 1-D counter
example. Let $\{x\}$ and $\{y\}$ denote the reference set and prediction
set (in the case of multi-object tracking $x$ and $y$ would represent
tracks with unit-length). Given a threshold $\theta>0$, (keeping
in mind that these sets are singletons) the number of true positives
is given by the indicator function $\mathbf{1}(|x-y|\leq\theta)$
(which equals 1 if $|x-y|\leq\theta$, and 0 otherwise). Despite differences
amongst the criteria in Section \ref{sec:related_work}, we can abstract
that any dissimilarity measure $d(\{x\},\{y\})$ of a criterion is
a function of only $\mathbf{1}(|x-y|\leq\theta)$, since number of
false positives and false negatives also depend on this value. More
concisely, $d(\{x\},\{y\})=D(\mathbf{1}(|x-y|\leq\theta))$, where
$D$ is a function such that: $D(1)=0$ (because $d(\{x\},\{x\})=0$
and $d(\{x\},\{x\})=D(1)$); and $D(0)>0$ (because if $D(0)=0$,
then $d(\{x\},\{y\})=0,$ for all $x$, $y$, making this a trivial
criterion). Now, the dissimilarity measure $d$ violates the \emph{Triangle
Inequality} because $d(\{x-0.6\theta\},\{x+0.6\theta\})=D(0)>0$,
but $d(\{x-0.6\theta\},\{x\})+d(\{x\},\{x+0.6\theta\})=D(1)+D(1)=0$.
It also violates the \emph{Identity} property because $\{x\}\neq\{x+0.6\theta\}$
but $d(\{x\},\{x+0.6\theta\})=D(1)=0$.

\section{Metric Performance Criteria \label{sec:metrics}}

\begin{figure}[!th]
\begin{centering}
\includegraphics[width=0.45\textwidth]{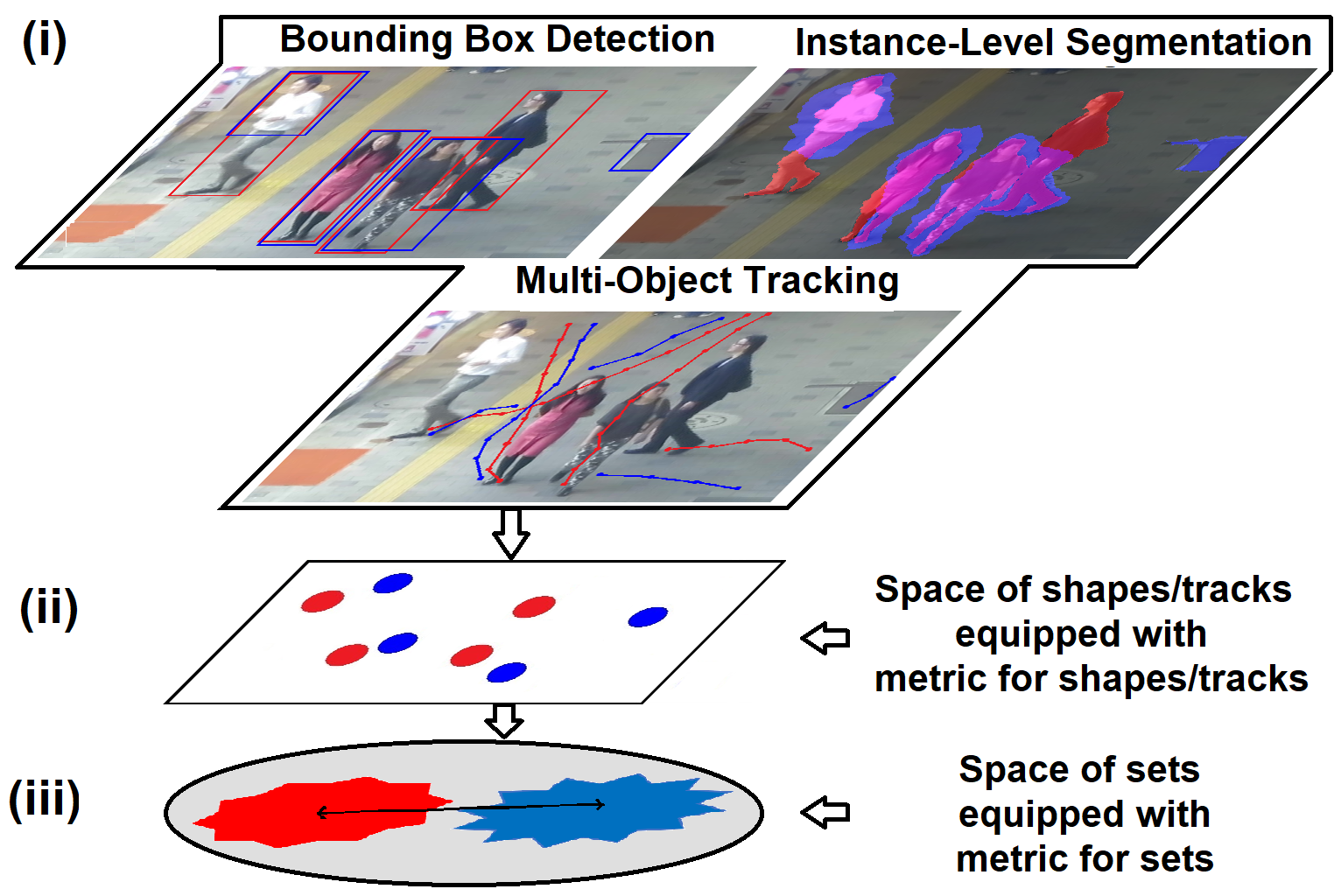} 
\par\end{centering}
\caption{(i) Truths (red) and predictions (blue) for bounding box detection,
instance-level segmentation, and multi-object tracking tasks; (ii)
Conceptualization (for all three tasks) of truths and predictions
as two sets of points in an abstract space; (iii) Dissimilarity of
these sets is measured by the (set) distance between them. \label{fig:demo_set_dist}}

\vspace{-1em}
\end{figure}

Fundamentally, performance evaluations for all three basic vision
tasks in this work can be cast in terms of measuring the dissimilarity
between two sets of shapes (see \Fig \ref{fig:demo_set_dist}). To
ensure mathematical consistency, we seek dissimilarity measures that
avoid the notion of true positives--the source of unreliability and
mathematical inconsistency. In this section, we explore (mathematical)
metrics or distances between two sets of shapes. This is accomplished
by using suitable metrics for shapes (Section \ref{subsec:ShapeMetric})
as the base-distance to construct a number of metrics for sets of
shapes from various point pattern metrics (Section \ref{subsec:SetMetric}). 

\subsection{Metrics for Shapes \label{subsec:ShapeMetric} }

For any two arbitrary shapes $x,y$, the Intersection over Union (IoU)
similarity index is given by $IoU(x,y)={\lambda(x\cap y)}/{\lambda(x\cup y)}\in[0,1]$,
where $\lambda(\cdot)$ denotes hyper-volume. For convex shapes, the
Generalized IoU index is given by $GIoU(x,y)=IoU(x,y)-{\lambda(C(x\cup y)\setminus\left(x\cup y\right))}/{\lambda(C(x\cup y))}$,
where $C(x\cup y)$ is the convex hull of $x\cup y$ \cite{005}.
Note that unlike $IoU(x,y)$, $GIoU(x,y)\in[-1,1]$. For arbitrary
shapes, the definition of GIoU is given in the supplementary section
of \cite{005}. As the defacto base-similarity measure for many performance
criteria, IoU/GIoU is a natural base-distances between shapes, required
to construct distances between sets of shapes. The metric forms of
IoU and GIoU, respectively are $\underline{d}_{IoU}(x,y)=1-IoU(x,y)$
and $\underline{d}_{GIoU}(x,y)=\frac{1-GIoU(x,y)}{2}$ \cite{005},
which are indeed metrics bounded by 1.

%\noindent 
\textbf{\emph{IoU/GIoU extension for shapes with confidence
score}}: The IoU/GIoU distance can also be extended to accommodate
basic vision solutions that attach to each shape a confidence score.
Note that such scores can be normalized to the interval $(0,1]$ since
reference shapes have maximum confidence scores of one. To determine
the IoU/GIoU distance between shapes with confidence scores, we take
the Cartesian products of the shapes with their corresponding confidence
scores to form augmented shapes in a higher dimensional space, and
then compute the IoU/GIoU distance between these augmented shapes.

\subsection{Metrics for Sets of Shapes \label{subsec:SetMetric} }

Our interest is the distance between two point patterns (or finite
subsets) of a metric space $(\mathcal{\mathbb{W}},\underline{d})$,
where $\underline{d}:\mathcal{\mathcal{\mathbb{W}}\times}\mathcal{\mathbb{W}}\rightarrow[0,1]$
denotes the\emph{ base-distance} between the elements of $\mathcal{\mathbb{W}}$.
Specifically, $\mathcal{\mathbb{W}}$ is the space of arbitrary/convex
shapes and the base-distance $\underline{d}$ is the IoU/GIoU distance. 

One option is to consider classical set distances such as \textit{Chamfer}
\cite{chamfer}, \textit{Hausdorff} \cite{hausdorff1967} and \textit{Earth
Mover Distance (EMD)} \cite{EMD1998} (or Wasserstein distance \cite{wasserstein1970}
of order one).

\revise{The Hausdorff distance between two non-empty point patterns $X$ and
$Y$ of $\mathcal{\mathbb{W}}$ is defined by \cite{hausdorff1967,BN-Hoffman}
\begin{equation}
d_{\mathtt{H}}(X,Y)=\max\left\{ \max_{x\in X}\min_{y\in Y}\underline{d}(x,y),\max_{y\in Y}\min_{x\in X}\underline{d}(x,y)\right\} .\label{eq:Hausdorff_dist}
\end{equation}
This metric was traditionally used as a measure of dissimilarity between
binary images. It gives a good indication of the dissimilarity in
the visual impressions that a human would typically perceive between
two binary images.}

\revise{In general, the Wasserstein distance (also known as Mallows distance)
of order $p\geq1$ between two non-empty point patterns $X=\{x_{1},...,x_{m}\}$
and $Y=\{y_{1},...,y_{n}\}$ is defined by \cite{wasserstein1970,BN-Hoffman}
\begin{equation}
d_{\mathtt{W}}^{(p)}(X,Y)=\min_{C}\left(\sum_{i=1}^{m}\sum_{j=1}^{n}c_{i,j}\underline{d}\left(x_{i},y_{j}\right)^{p}\right)^{\frac{1}{p}},\label{eq:OMAT-dist}
\end{equation}
where $C=\left(c_{i,j}\right)$ is an $m\times n$ transportation
matrix, \ie, the entries $c_{i,j}$ are non-negative, each row sum
to $1/m$, and each column sum to $1/n$. The order $p$ in the Wasserstein
distance plays the same role as the order of the $\ell_{p}$-distance
for vectors, which is usually assumed to be 1 or 2 in most applications.}

\revise{For an IoU/GIoU base-distance, which is a ratio of hyper-volumes,
the Wasserstein distance of order $1$ has a more natural interpretation
than its higher order counterparts. This special case is commonly
known as the EMD. If we consider the sets
$X$ and $Y$ as collections of earth piles and suppose that the cost
of moving a mass of earth over a distance is given by the mass times
the distance. Then EMD can be considered as
the minimum cost needed to build one collection of earth piles from
the other.}

\revise{Note that, in general, the Hausdorff and Wasserstein metrics are not
defined when either of the set is empty. This is problematic for performance
evaluation because it is not uncommon for the prediction set or reference
set to be empty. However, when $\underline{d}$ is bounded by 1 (as
per the IoU/GIoU distance), this problem can be resolved (while observing
the metric properties) by defining $d_{\mathtt{H}}(X,Y)=d_{\mathtt{W}}^{(p)}(X,Y)=1$
if one of the set is empty, and $d_{\mathtt{H}}(\emptyset,\emptyset)=d_{\mathtt{W}}^{(p)}(\emptyset,\emptyset)=0$.}

\revise{The Hausdorff and Wasserstein metrics are constructed for arbitrary sets and
probability distributions}. Thus, whether they capture the intent of performance
evaluation in basic vision tasks, remain to be verified. The intent
behind the performance criteria discussed in Section \ref{sec:related_work}
is to capture the dislocation and cardinality error. What these criteria
have in common is the pairing of predicted and reference points so
as to minimize the sum of base-distances between the pairs, either
by greedy assignment or optimal assignment. Despite differences amongst
various criteria, the dislocation is determined from the matched pairs
(those with base-distances below a threshold), and the cardinality
error from unmatched elements, which are then combined to produce
a normalized or averaged score.

An alternative to classical set distances is to find a metric that
captures the above intent. Instead of thresholding the base-distance
between the pairs to determine true positives, which violates the
metric properties, we can capture the same intent simply by adding
the minimum sum of base-distances (representing dislocation) with
the number of unpaired elements (representing cardinality error),
and normalize by the total number of pairs and unpaired elements.
Simply put, this is the best-case per-object dislocation and cardinality
error, \ie for $X=\{x_{1},...,x_{m}\}$ and $Y=\{y_{1},...,y_{n}\}$,
\begin{align}
d_{\mathtt{O}}(X,Y)= & \frac{1}{n}\left(\min_{\pi\in\Pi_{n}}\sum_{i=1}^{m}\underline{d}\left(x_{i},y_{\pi(i)}\right)+\left(n-m\right)\right),\label{eq:OSPA-dist}
\end{align}
 if $n\geq m>0$, where $\Pi_{n}$ is the set of all permutations
of $\left\{ 1,2,...,n\right\} $, additionally: $d_{\mathtt{O}}(X,Y)=d_{\mathtt{O}}(Y,X)$,
if $m>n>0$; $d_{\mathtt{O}}(X,Y)=1$, if one of the set is empty;
and $d_{\mathtt{O}}(\emptyset,\emptyset)=0$. This normalized error
is indeed the \textit{Optimal Sub-Pattern Assignment (OSPA)} metric
\cite{OSPA1}, which can be computed efficiently in polynomial time
via optimal assignment algorithms.

\revise{Note that, although the current formulation of the metric is suitable for generic evaluation tasks where no preference is given to cardinality or localization, the original OSPA metric (see appendix Section 3.2) allows such emphasis via a cut-off parameter. Recently, an attempt to distinguish the false positives and false negatives components of cardinality error in OSPA, called  the \textit{Deficiency Aware Sub-pattern Assignment (DASA)} metric, has been introduced in  \cite{DASA2018oksuz,LRP-ECCV18,oksuz2021metric}.}

\subsection{Metrics for Sets of Tracks
\begin{figure}[t!]
\protect\centering{}\protect\includegraphics[width=0.35\textwidth]{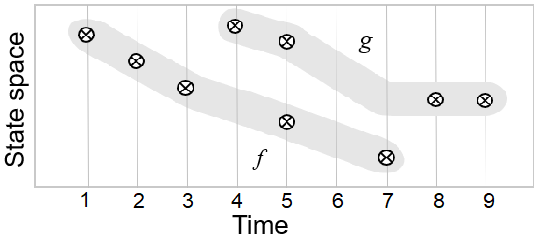}
\protect\caption{Two fragmented tracks $f$ and $g$ in a 1-D state space. Note that
at k = 6 both tracks are undefined (or non-existent).\label{fig:def_tracks} }
\protect
\end{figure}
}

For performance evaluation of multi-object tracking, the metrics for
sets of shapes discussed earlier are not directly applicable because
a track cannot be treated as a shape or a set of shapes due to the
temporal ordering of its constituents. A \textit{track} in a metric
space $(\mathcal{\mathbb{W}},\underline{d})$ and discrete-time window
$\mathbb{T}$, is defined as a mapping $f:\mathbb{T}\mapsto\mathbb{W}$
\cite{OSPA2}. Its \textit{domain} $\mathcal{D}_{f}\subseteq\mathbb{T}$,
is the set of time instants when the object/track has a state in $\mathbb{W}$.
This definition accommodates the so-called fragmented tracks, \ie
tracks with domains that are not intervals, see \Fig\ref{fig:def_tracks}
for visualization in a 1-D state-space.

A meaningful distance between two sets of tracks requires a meaningful
base-distance between two tracks. The most suitable for multi-object
tracking is the time-averaged OSPA distance over instants when at
least one of the tracks exists \cite{OSPA2}, \ie for two tracks
$f$ and $g$ 
\begin{align}
\underline{\widetilde{d}}\left(f,g\right)= & \sum_{t\in\mathcal{D}_{f}\cup\mathcal{D}_{g}}\!\frac{d_{\mathtt{O}}\left(\left\{ f\left(t\right)\right\} ,\left\{ g\left(t\right)\right\} \right)}{\left|\mathcal{D}_{f}\cup\mathcal{D}_{g}\right|},\label{e:Track_Distance-1}
\end{align}
if $\mathcal{D}_{f}\cup\mathcal{D}_{g}\neq\emptyset$, where $\left|\cdot\right|$
denotes cardinality, and $\underline{\widetilde{d}}\left(f,g\right)=0$,
if $\mathcal{D}_{f}\cup\mathcal{D}_{g}=\emptyset$. For example, the
distance between the tracks in Fig. \ref{fig:def_tracks} is the average
OSPA distance between them over all instances in $\{1,...,9\}$ except
for $k=6$, the instance when both tracks are undefined. The distance
$\underline{\widetilde{d}}$ is indeed a metric \cite{OSPA2} bounded
by 1. 

Using the Hausdorff, EMD, and OSPA metrics, respectively, with base-distance
$\underline{\widetilde{d}}$, yield the Hausdorff($\underline{\widetilde{d}}$),
EMD($\underline{\widetilde{d}}$), and OSPA($\underline{\widetilde{d}}$)
distances between two sets of tracks. The latter is called OSPA\textsuperscript{(2)}
(since $\underline{\widetilde{d}}$ is constructed from OSPA) and
can be interpreted as the time-averaged per-track error. OSPA\textsuperscript{(2)}
takes into account errors in localization, cardinality, track fragmentation
and identity switching \cite{OSPA2}. A dropped track that later regained
with the same identity incurs a smaller penalty than if it were regained
with a different identity.

\textbf{\emph{Remark}}: The Hausdorff, EMD, and OSPA metrics (with
both base-distances $\underline{d}$ and $\underline{\widetilde{d}}$)
above are mathematically consistent (by default) and reliable (no
parameters). How meaningful they are will be examined in Section \ref{sec:exp_sanity},
while further discussions can be found in Section 3 of the appendix. 

In contrast to the inconsistencies of various criteria shown in \Fig
\ref{fig:TAP_exp}, \revise{the (prediction) errors measured from ground truth and approximate truth are similar (for the $\text{OSPA}$ and $\text{Hausdorff}$ metrics) given the difference between ground truth and approximate truth is small (the
same observation holds for the EMD).} Moreover,
compared to the criteria in \Fig \ref{fig:apprx_traditional},
\Tab \ref{tab:apprx_metric} shows that for metric criteria, the
differences in (Kendall-tau) ranking errors between ground truth reference
and approximate truth reference are negligible.

\begin{figure}
\begin{centering}
\includegraphics[width=0.45\textwidth]{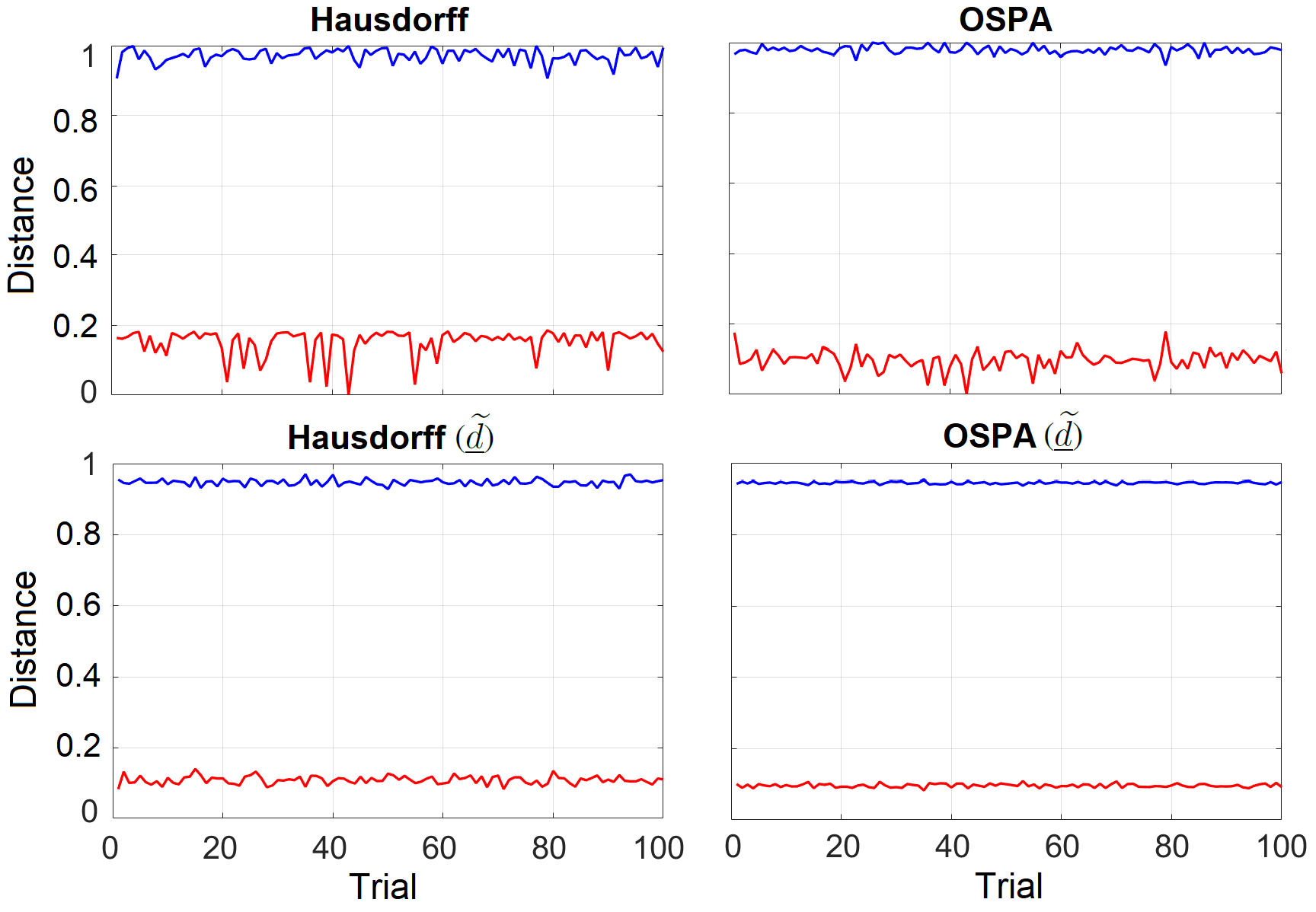}
\par\end{centering}
\caption{\revise{\textbf{(Red)} Dissimilarity between ground truth and approximate
truth. \textbf{(Blue)} Dissimilarity between approximate truth
and prediction. Both are normalized against the dissimilarity between ground
truth and prediction. The results are based on the
same dataset as per \Fig \ref{fig:TAP_exp}. In contrast to \Fig \ref{fig:TAP_exp}, the blue lines
are close to 1, which demonstrates the consistency
of metric criteria .}\label{fig:TAP_met_exp}}
\end{figure}

\begin{table*}
\centering{}\caption{Monte Carlo means (and standard deviations) of normalized Kendall-tau ranking
errors for various metric criteria with ground truth and approximate
truth reference sets (using the same dataset as per Fig. \ref{fig:apprx_traditional}). \revise{The Kendall-tau errors between rankings based on ground truth and approximate truth are similar.}
\label{tab:apprx_metric}}
\begin{tabular}{ccccccc}
\toprule 
\multicolumn{7}{c}{\textbf{\scriptsize{}Multi-Class Multi-Object Detection: Normalized
Kendall-tau ranking error (in units of $\mathbf{10^{-2}}$)}}\tabularnewline
\midrule 
 & \multicolumn{2}{c}{\textbf{\scriptsize{}Hausdorff}} & \multicolumn{2}{c}{\textbf{\scriptsize{}EMD}} & \multicolumn{2}{c}{\textbf{\scriptsize{}OSPA}}\tabularnewline
\midrule 
 & \textbf{\scriptsize{}IoU}  & \textbf{\scriptsize{}GIoU}  & \textbf{\scriptsize{}IoU}  & \textbf{\scriptsize{}GIoU}  & \textbf{\scriptsize{}IoU}  & \textbf{\scriptsize{}GIoU}\tabularnewline
\midrule 
\textbf{\scriptsize{}Ground truth reference}  & {\scriptsize{}$7.75\thinspace(4.62)$} & {\scriptsize{}$9.36\thinspace(4.79)$} & {\scriptsize{}$4.16\thinspace(3.64)$} & {\scriptsize{}$5.45\thinspace(4.33)$} & {\scriptsize{}$3.06\thinspace(3.35)$} & {\scriptsize{}$4.20\thinspace(3.80)$}\tabularnewline
\midrule 
\textbf{\scriptsize{}Approximate truth reference}  & {\scriptsize{}$7.94\thinspace(4.58)$} & {\scriptsize{}$9.48\thinspace(4.80)$} & {\scriptsize{}$4.48\thinspace(3.66)$} & {\scriptsize{}$5.61\thinspace(4.25)$} & {\scriptsize{}$3.37\thinspace(3.34)$} & {\scriptsize{}$4.41\thinspace(3.92)$}\tabularnewline
\midrule 
\multicolumn{7}{c}{\textbf{\scriptsize{}Multi-Object Tracking: Normalized Kendall-tau
ranking error (in units of $\mathbf{10^{-2}}$) }}\tabularnewline
\midrule 
 & \multicolumn{2}{c}{{\scriptsize{}$\boldsymbol{\textbf{Hausdorff}(\widetilde{\underline{d}})}$}} & \multicolumn{2}{c}{{\scriptsize{}$\boldsymbol{\textbf{EMD}(\widetilde{\underline{d}})}$}} & \multicolumn{2}{c}{{\scriptsize{}$\boldsymbol{\textbf{OSPA}(\widetilde{\underline{d}})}$}}\tabularnewline
\midrule 
 & \textbf{\scriptsize{}IoU}  & \textbf{\scriptsize{}GIoU}  & \textbf{\scriptsize{}IoU}  & \textbf{\scriptsize{}GIoU}  & \textbf{\scriptsize{}IoU}  & \textbf{\scriptsize{}GIoU}\tabularnewline
\midrule 
\textbf{\scriptsize{}Ground truth reference}  & {\scriptsize{}$11.3\thinspace(9.85)$} & {\scriptsize{}$11.0\thinspace(5.41)$} & {\scriptsize{}$3.63\thinspace(2.46)$} & {\scriptsize{}$5.90\thinspace(3.28)$} & {\scriptsize{}$0.536\thinspace(0.617)$} & {\scriptsize{}$0.538\thinspace(0.609)$}\tabularnewline
\midrule 
\textbf{\scriptsize{}Approximate truth reference}  & {\scriptsize{}$11.3\thinspace(9.95)$} & {\scriptsize{}$11.1\thinspace(5.37)$} & {\scriptsize{}$3.68\thinspace(2.46)$} & {\scriptsize{}$5.94\thinspace(3.29)$} & {\scriptsize{}$0.611\thinspace(0.663)$} & {\scriptsize{}$0.582\thinspace(0.636)$}\tabularnewline
\bottomrule
\end{tabular}
\end{table*}

% \subsection{The Meaningfulness of Metric Criteria}
\revise{Note that mathematical consistency and/or reliability are not sufficient
to warrant meaningful performance evaluation. Consider the simple
sanity check for people detection in \Fig \ref{fig:OSPA_intuition-1}. Detector
A achieves an IoU error of 0.25 for each of the 3 objects in the scene,
while detector B incurs an IoU error of 0.7 even with only one object.
Reiterating our previous discussion, unequivocally, detector A performs
better than B. A naive metric such as the unnormalized OSPA distance
(no dividing by the number of objects) is mathematically consistent
(because the normalizing factor does not affect the metric axioms)
and reliable (because there are no parameters). However, according
to this metric B (0.7 total IoU error) has smaller prediction error
than A (0.75 total IoU error), \ie B performs better A, which is
nonsensical. In contrast, a mathematically inconsistent criterion
like F1 is more meaningful, confirming (for a 0.5 threshold) that
A performs better than B, and even if the threshold is varied, would
never declare B to be the better.}

\revise{When the number of detected object is correct, it is obvious that
a criterion should not assign a larger error to a scenario with an
accurate prediction than a (different) scenario with an inaccurate
prediction. Hence, it is necessary to sanity-test a criterion across
different scenarios, along the line of the example in \Fig \ref{fig:OSPA_intuition-1}.}

\revise{To this extent, we present a sanity test that assesses the criterion's
meaningfulness across different scenarios numbered from 1 to 10. In
scenario $k$, the number of true objects is $2^{k}$. The objects
are 10 pixels by 10 pixels squares, evenly spaced so that the nearest object is
more than 20 pixels away. The prediction set is the true set with each object
shifted to the left by $2^{-0.5k}$ pixels. Since the predicted cardinality
is correct, unequivocally, scenario 1 must have larger prediction
error than scenario 2 and so on as the localization error decreases
from scenario 1 to scenario 10.}

\begin{figure}[H]
\centering{}\includegraphics[width=0.45\textwidth]{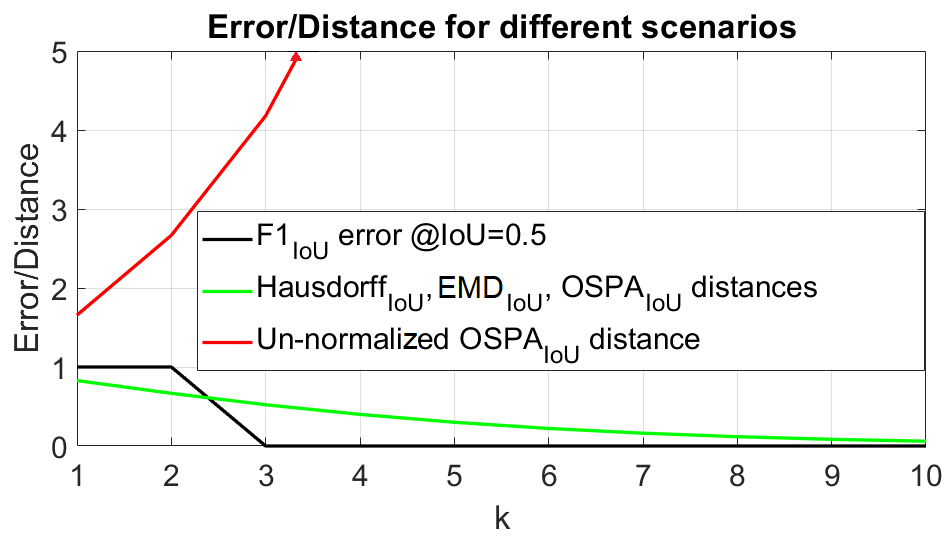}
\caption{The distances between true and predictions sets in scenario $k$ of
the sanity test with only dislocation of the centroid.\label{fig:special_sanity_test}}
\end{figure}

\revise{\Fig \ref{fig:special_sanity_test} plots the $\text{F1}_{\text{IoU}}$
prediction error ($1-\textrm{F1}_{\text{IoU}}$), $\text{OSPA}_{\text{IoU}}$,
un-normalized $\text{OSPA}_{\text{IoU}}$, $\text{EMD}_{\text{IoU}}$,
and $\text{Hausdorff}_{\text{IoU}}$ distances for each scenario.
Note that the $\text{EMD}_{\text{IoU}}$, $\text{Hausdorff}_{\text{IoU}}$
and $\text{OSPA}_{\text{IoU}}$ distances exhibit identical behavior
that corroborate with physical intuition as they decrease with better
performance. The $\text{F1}_{\text{IoU}}$ distance can only take
the value of either 0 or 1, and is not granular enough to distinguish
the prediction errors in scenarios 1, 2 and 3 to 10. Nonetheless,
it still shows the general trend of improving performance. In contrast,
the un-normalized $\text{OSPA}_{\text{IoU}}$ metric\footnote{This distance takes same the form as \Eq \ref{eq:OSPA-dist} but
without the normalizing constant $1/n$.} produces non-sensical prediction error that increases drastically
with unequivocally better performance.}

\section{Assessing 
%Performance 
Criteria via Sanity Tests\label{sec:exp_sanity} }

In Section \ref{sec:guideline}, we suggested guidelines to certify
the trustworthiness of a performance criterion via its reliability,
meaningfulness and mathematical consistency. In this experiment, we
use sanity tests (Section \ref{subsec:Meaningfulness}) to examine
the meaningfulness of different performance criteria for bounding
box multi-object detection and multi-object tracking. Tests on instance-level
segmentation are omitted as bounding boxes can be interpreted as masks,
with both having similar properties in terms of similarity measure.
The sanity test for each task is performed via 100 randomly sampled
ground truth and 100 predictions sets of pre-determined ranking for
each ground truth (totalling 10000 Monte Carlo trials for each task). The construction of the tests (each trial) are briefly described in the following, details can be found in Sections 4.1 and 4.2 of the appendix.

\begin{figure*}
\centering{}\includegraphics[width=0.95\textwidth]{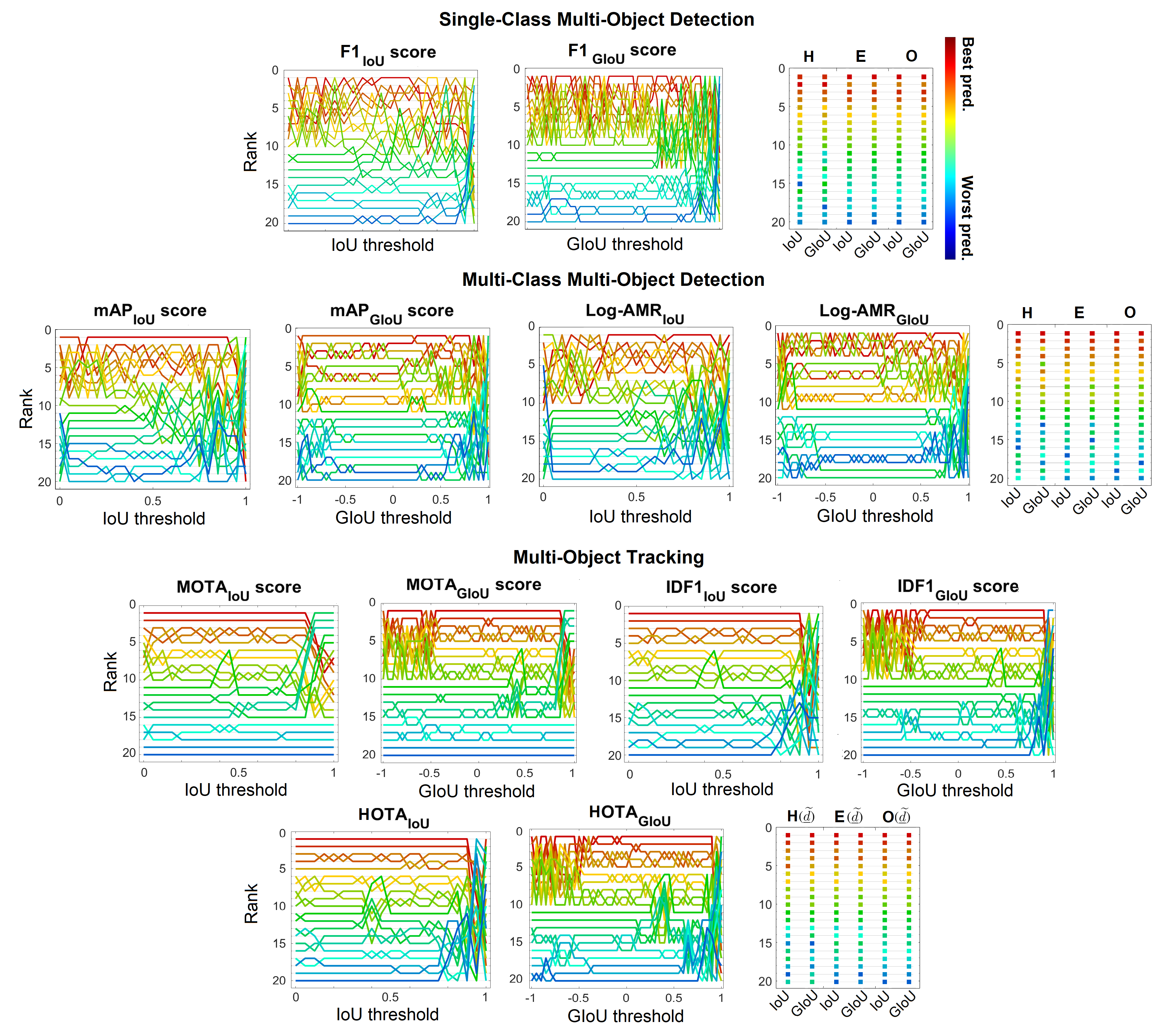}
\caption{Ranks of prediction sets (for a sample reference set) according to
various traditional criteria over a range of IoU/GIoU thresholds, and according to Hausdorff (\textbf{H}),
EMD (\textbf{E}), OSPA (\textbf{O}) metrics. The pre-determined
ranks are color-coded from worst (blue) to best (red).\label{fig:sanity_scores}}
\end{figure*}
\begin{table*}
\centering{}\caption{Monte Carlo means (and standard deviations) of normalized Kendall-tau
ranking errors of various criteria at certain thresholds. The subscripts
of IoU/GIoU indicate the threshold values; ``optimal'' threshold
is the one with best ranking accuracy; ``M-partial'' indicates
that the evaluation is done via averaging the score/rate over the
range 0.5 to 0.95 in steps of 0.05. ``M-full'' indicates that
the evaluation is done via averaging the score/rate over the entire
range of the base-measure (excluded two extreme thresholds). \label{tab:sanity_Man}}
\begin{tabular}{|c|cccc|cccc|}
\hline 
 & {\scriptsize{}$\textbf{IoU}_{\boldsymbol{0.5}}$} & {\scriptsize{}$\textbf{IoU}_{\textbf{optimal}}$} & {\scriptsize{}$\textbf{IoU}_{\textbf{M-partial}}$} & {\scriptsize{}$\textbf{IoU}_{\textbf{M-full}}$} & {\scriptsize{}$\textbf{GIoU}_{\boldsymbol{0}}$} & {\scriptsize{}$\textbf{GIoU}_{\textbf{optimal}}$} & {\scriptsize{}$\textbf{GIoU}_{\textbf{M-partial}}$} & {\scriptsize{}$\textbf{GIoU}_{\textbf{M-full}}$}\tabularnewline
\hline 
\multicolumn{9}{|c|}{\textbf{\scriptsize{}Single-Class Multi-Object Detection: Normalized
Kendall-tau ranking error (in units of $\mathbf{10^{-2}}$) }}\tabularnewline
\hline 
\textbf{\scriptsize{}F1}{\scriptsize{} } & {\scriptsize{}$10.0\thinspace\left(8.82\right)$} & {\scriptsize{}$7.33\thinspace\left(5.17\right)$} & {\scriptsize{}$6.68\thinspace(9.39)$} & {\scriptsize{}$2.15\thinspace(1.51)$} & {\scriptsize{}$7.89\thinspace\left(3.05\right)$} & {\scriptsize{}$7.69\thinspace\left(4.90\right)$} & {\scriptsize{}$7.49\thinspace\left(9.36\right)$} & \textbf{\scriptsize{}$\mathbf{2.17\thinspace\left(1.36\right)}$}\tabularnewline
\hline 
{\scriptsize{}$\textbf{Hausdorff}$} & \multicolumn{4}{c|}{{\scriptsize{}$17.8\thinspace\left(9.87\right)$}} & \multicolumn{4}{c|}{{\scriptsize{}$22.4\thinspace\left(11.1\right)$}}\tabularnewline
\hline 
{\scriptsize{}$\textbf{EMD}$} & \multicolumn{4}{c|}{{\scriptsize{}$3.88\thinspace\left(1.96\right)$}} & \multicolumn{4}{c|}{{\scriptsize{}$5.16\thinspace\left(3.03\right)$}}\tabularnewline
\hline 
{\scriptsize{}$\textbf{OSPA}$} & \multicolumn{4}{c|}{{\scriptsize{}$\mathbf{1.97\thinspace\left(1.48\right)}$}} & \multicolumn{4}{c|}{{\scriptsize{}$2.22\thinspace\left(1.43\right)$}}\tabularnewline
\hline 
\multicolumn{9}{|c|}{\textbf{\scriptsize{}Multi-Object Multi-Class Detection: Normalized
Kendall-tau ranking error (in units of $\mathbf{10^{-2}}$) }}\tabularnewline
\hline 
{\scriptsize{}$\textbf{mAP}$ } & {\scriptsize{}$10.0\thinspace\left(8.90\right)$} & {\scriptsize{}$7.08\thinspace\left(5.56\right)$} & {\scriptsize{}$7.52\thinspace(8.71)$}  & {\scriptsize{}$3.62\thinspace(2.65)$}  & {\scriptsize{}$9.41\thinspace\left(3.81\right)$} & {\scriptsize{}$7.39\thinspace\left(5.51\right)$} & {\scriptsize{}$8.27\thinspace\left(8.71\right)$} & {\scriptsize{}$4.86\thinspace(3.00)$} \tabularnewline
\hline 
{\scriptsize{}$\textbf{Log-AMR}$} & {\scriptsize{}$9.91\thinspace\left(5.97\right)$} & {\scriptsize{}$8.42\thinspace\left(5.29\right)$} & {\scriptsize{}$6.75\thinspace\left(3.42\right)$} & {\scriptsize{}$4.31\thinspace(2.30)$}  & {\scriptsize{}$16.5\thinspace\left(6.57\right)$} & {\scriptsize{}$8.80\thinspace\left(5.46\right)$} & {\scriptsize{}$7.33\thinspace\left(3.69\right)$} & {\scriptsize{}$4.95\thinspace(2.83)$} \tabularnewline
\hline 
{\scriptsize{}$\textbf{Hausdorff}$} & \multicolumn{4}{c|}{{\scriptsize{}$5.43\thinspace\left(2.71\right)$}} & \multicolumn{4}{c|}{{\scriptsize{}$6.39\thinspace\left(2.88\right)$}}\tabularnewline
\hline 
{\scriptsize{}$\textbf{EMD}$} & \multicolumn{4}{c|}{{\scriptsize{}$2.80\thinspace\left(1.83\right)$}} & \multicolumn{4}{c|}{{\scriptsize{}$3.50\thinspace\left(2.26\right)$}}\tabularnewline
\hline 
{\scriptsize{}$\textbf{OSPA}$ } & \multicolumn{4}{c|}{{\scriptsize{}$\mathbf{1.86\thinspace\left(1.64\right)}$}} & \multicolumn{4}{c|}{{\scriptsize{}$\mathbf{2.41\thinspace\left(1.90\right)}$}}\tabularnewline
\hline 
\multicolumn{9}{|c|}{\textbf{\scriptsize{}Multi-Object Tracking: Normalized Kendall-tau
ranking error (in units of $\mathbf{10^{-2}}$) }}\tabularnewline
\hline 
{\scriptsize{}$\textbf{MOTA}$ } & {\scriptsize{}$5.18\thinspace\left(5.51\right)$} & {\scriptsize{}$1.42\thinspace\left(1.60\right)$} & {\scriptsize{}$7.64\thinspace\left(7.74\right)$} & {\scriptsize{}$3.26\thinspace\left(3.90\right)$} & {\scriptsize{}$2.20\thinspace\left(2.28\right)$} & {\scriptsize{}$1.00\thinspace\left(1.39\right)$} & {\scriptsize{}$8.46\thinspace\left(8.22\right)$} & {\scriptsize{}$0.872\thinspace\left(0.806\right)$}\tabularnewline
\hline 
{\scriptsize{}$\textbf{IDF1}$} & {\scriptsize{}$3.47\thinspace\left(3.51\right)$} & {\scriptsize{}$1.84\thinspace\left(1.63\right)$} & {\scriptsize{}$3.04\thinspace\left(3.00\right)$} & {\scriptsize{}$1.38\thinspace\left(1.54\right)$} & {\scriptsize{}$2.82\thinspace\left(2.47\right)$}  & {\scriptsize{}$1.24\thinspace\left(1.68\right)$} & {\scriptsize{}$3.93\thinspace\left(3.62\right)$} & {\scriptsize{}$0.676\thinspace\left(0.920\right)$}\tabularnewline
\hline 
{\scriptsize{}$\textbf{HOTA}$} & {\scriptsize{}$4.11\thinspace\left(4.17\right)$} & {\scriptsize{}$2.95\thinspace\left(2.57\right)$} & {\scriptsize{}$3.56\thinspace\left(3.70\right)$} & {\scriptsize{}$1.45\thinspace\left(1.63\right)$ } & {\scriptsize{}$4.34\thinspace\left(3.21\right)$} & {\scriptsize{}$4.24\thinspace\left(3.19\right)$} & {\scriptsize{}$4.19\thinspace\left(4.18\right)$} & {\scriptsize{}$1.02\thinspace\left(1.26\right)$ }\tabularnewline
\hline 
{\scriptsize{}$\boldsymbol{\mathbf{Hausdorff}(\underline{\tilde{d}})}$}  & \multicolumn{4}{c|}{{\scriptsize{}$12.0\thinspace\left(9.64\right)$}} & \multicolumn{4}{c|}{{\scriptsize{}$10.6\thinspace\left(5.27\right)$}}\tabularnewline
\hline 
{\scriptsize{}$\boldsymbol{\mathbf{EMD}(\underline{\tilde{d}})}$} & \multicolumn{4}{c|}{{\scriptsize{}$3.53\thinspace\left(2.38\right)$}} & \multicolumn{4}{c|}{{\scriptsize{}$5.80\thinspace\left(3.29\right)$}}\tabularnewline
\hline 
{\scriptsize{}$\boldsymbol{\mathbf{OSPA}(\underline{\tilde{d}})}$} & \multicolumn{4}{c|}{{\scriptsize{}$\mathbf{0.518\thinspace\left(0.580\right)}$}} & \multicolumn{4}{c|}{{\scriptsize{}$\mathbf{0.539\thinspace\left(0.577\right)}$}}\tabularnewline
\hline 
\end{tabular}
\end{table*}

\begin{figure}[h!]
\begin{centering}
\includegraphics[width=0.45\textwidth]{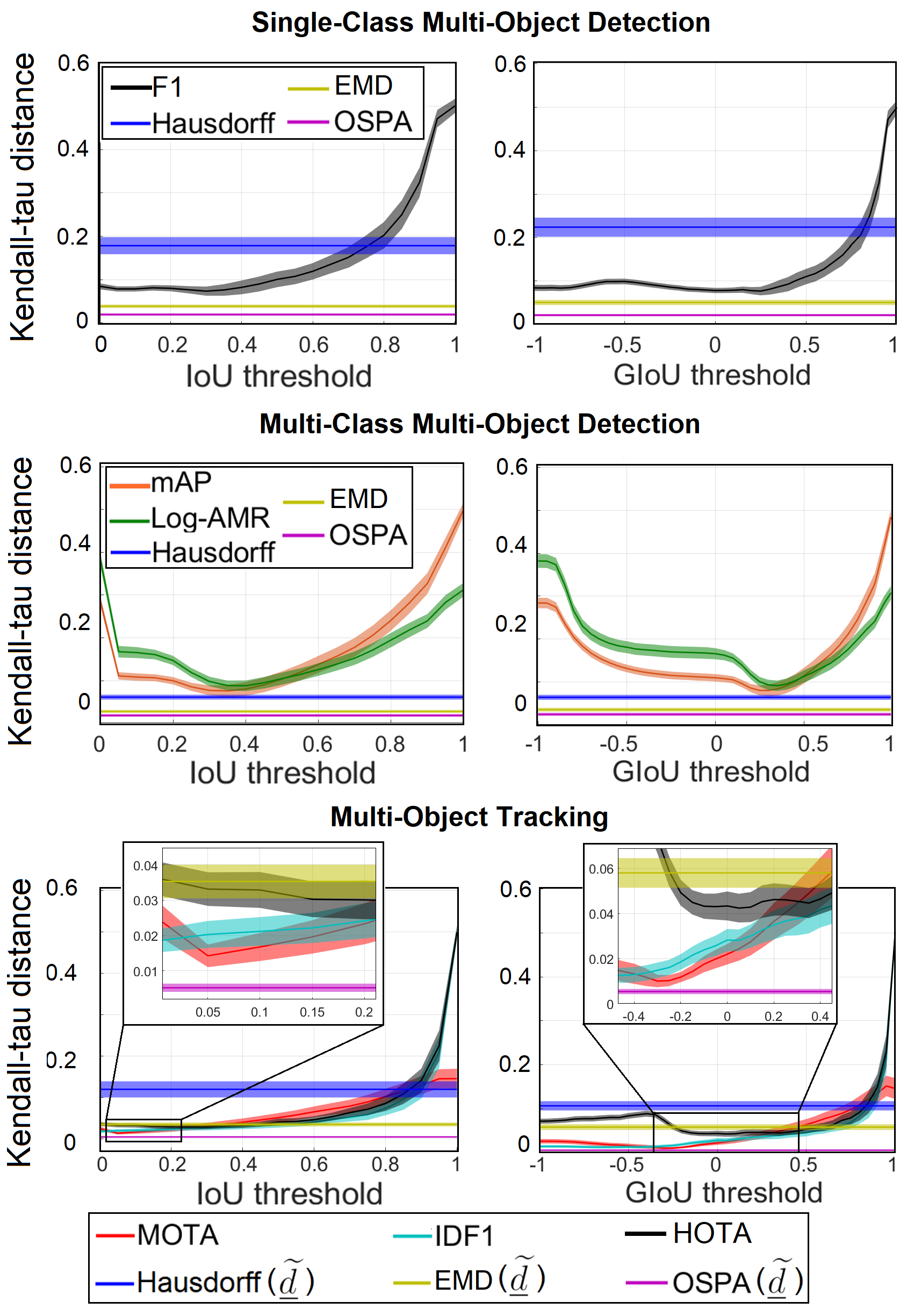} 
\par\end{centering}
\caption{Monte Carlo means of normalized Kendall-tau ranking errors for various criteria at different thresholds, in \textbf{detection
tests} and \textbf{tracking test}. Shaded area around each curve indicates
0.2-sigma bound. \label{fig:sanity_Man}}
\end{figure}

\subsection{Sanity Test for Multi-Object Detection\label{subsec:exp_detection_sanity}}
We first sample a set of bounding
boxes for the reference set, and then perturb
this set to form 20 prediction sets with pre-determined ranks.
The lower the prediction set is ranked: the higher the disturbance in locations and sizes, the higher the number of missed objects, false positives. Additionally, for multi-class detection test, the lower the prediction set is ranked: the higher number of predicted objects with incorrect classes and the lower the detection confidence scores for objects with correct class. In the multi-class detection test, the evaluation score/rate/distance is averaged across all classes.

\subsection{Sanity Test for Multi-Object Tracking\label{subsec:exp_tracking_sanity}}
First, we simulate the initial states of the tracks by  generating a random number of random bounding boxes
at random instances in the 100 time-step window. We then simulate the track lengths randomly from the interval $\{50,...,100\}$ and, accordingly, propagate the initial states
in time via the constant velocity model to simulate a reference set (of tracks). We generate 20 predictions sets (of tracks) with pre-determined ranks by perturbing the reference set. The simulated numbers of missed objects at each time step and false tracks increase from the best prediction set to the worst. Simulated false tracks randomly appear in the scene during their active periods while their sizes vary without any dynamics. Identities swapping events are simulated so that the lower rank prediction sets have, at the same level of mutual IoU, more tracks identity swapping.

\subsection{Results and Discussions\label{subsec:exp_sanity_discuss}}
For completeness, we use both IoU and GIoU metrics for performance
criteria in our experiment. \Fig \ref{fig:sanity_scores} shows traditional
performance criteria producing ranking orders switching severely across
different IoU/GIoU thresholds. In general, more meaningful
criteria should incur smaller ranking errors. Hence, \Fig \ref{fig:sanity_Man}
further confirms that the ranking accuracy (meaningfulness) of these
criteria also vary considerably across the range of IoU/GIoU thresholds,
albeit generally better at low thresholds. \Tab \ref{tab:sanity_Man}
shows that ranking performance at mid-scale threshold is usually not optimal, while the optimal threshold varies depending on the characteristics of the data. It also shows that partially marginalizing the parameters may produce less meaningful
rankings compared to the optimal threshold in the detection test (see
mAP score with IoU). While marginalizing over the entire range of threshold seems to improve the ranking performance, especially, for single-class multi-object detection and multi-object tracking tests, there is nothing to guarantee this in general. Given its insensitivity to the cardinality error, Hausdorff metric tends to have worse ranking performance than other criteria. In contrast, EMD and OSPA metrics show improved ranking performance compared to traditional criteria, with OSPA being the better metric because it also captures the intuition of traditional criteria (but without thresholding). \revise{Further, ranking results using the OSPA metric at different cut-off values are given in the appendix Section 4.}

\section{Real Benchmark Datasets Ranking \label{sec:exp_real}}

This section presents some observations on the traditional benchmarks
and suggested metrics, in the context of how they rank various real
detectors and trackers on public datasets.

\begin{figure*}
\centering{}\includegraphics[width=0.9\textwidth]{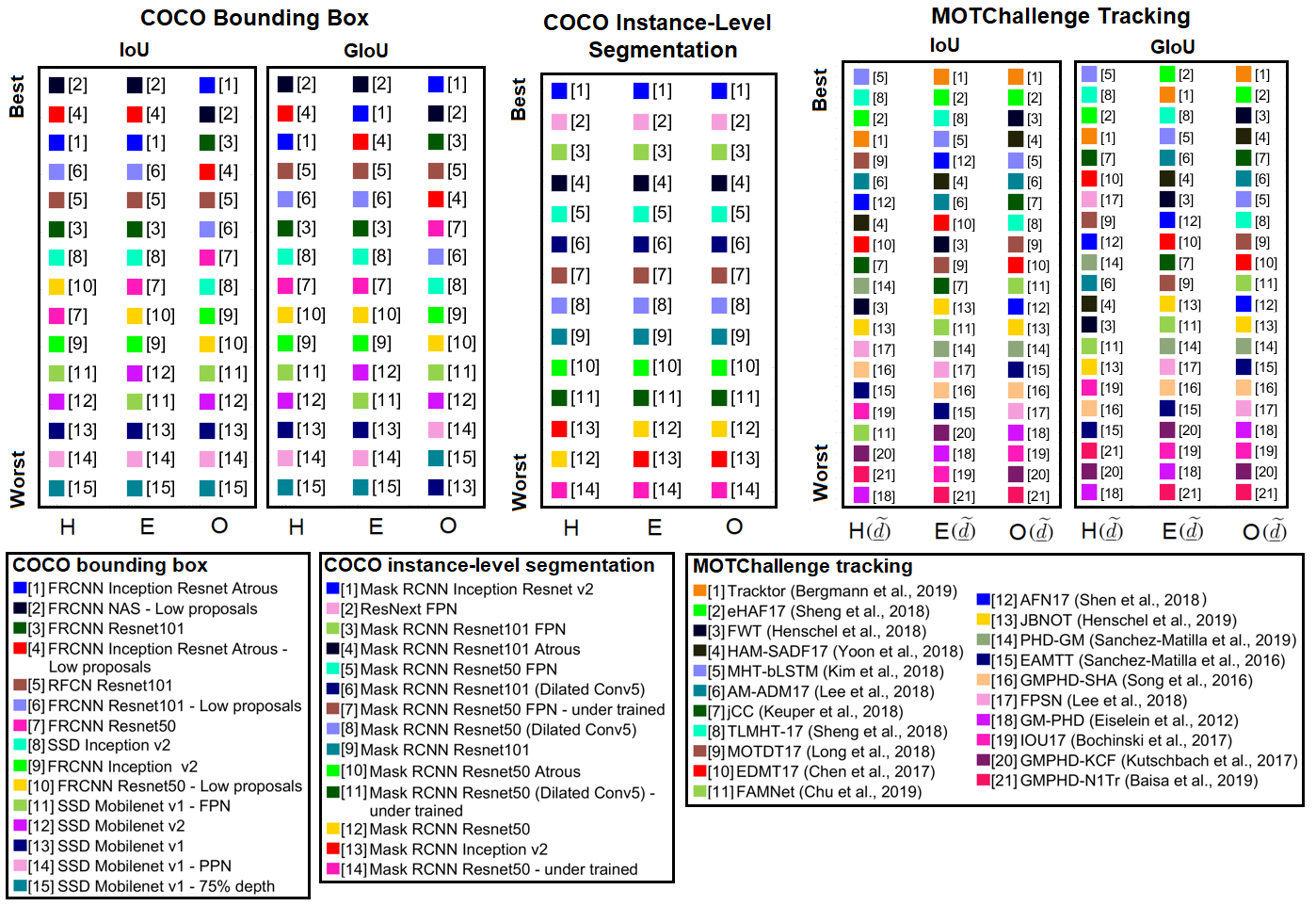}
\caption{Ranks of real algorithms via \textbf{H}: Hausdorff, \textbf{E}: EMD
and \textbf{O}: OSPA metrics on public datasets in COCO bounding box
detection, COCO instance-level segmentation and MOTChallenge tracking
experiments.\label{fig:real_ranks}\vspace{-10pt}}
\end{figure*}
\begin{figure*}
\centering{}\includegraphics[width=0.9\textwidth]{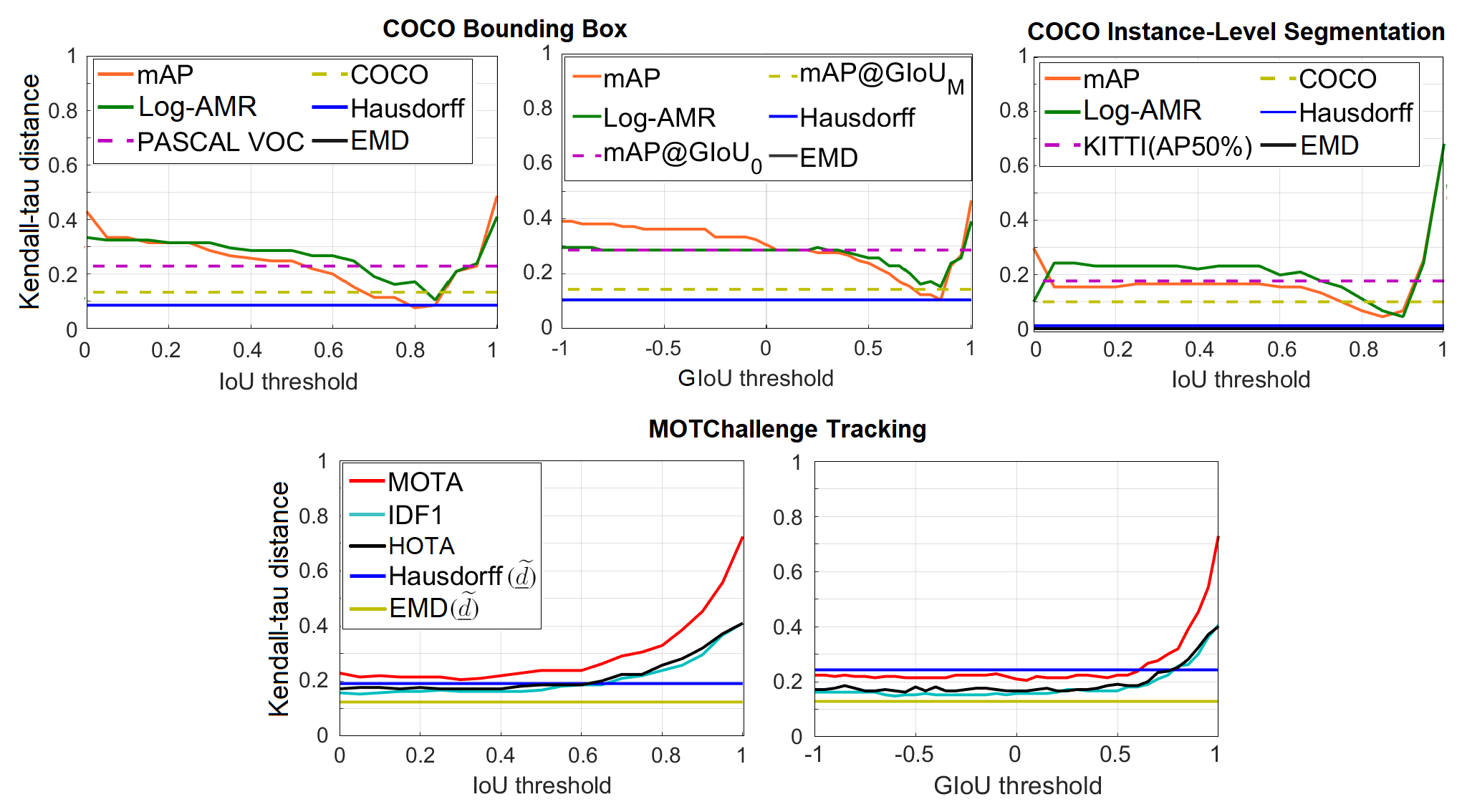}
\caption{Normalized Kendall-tau distances between rankings of OSPA and of other
performance criteria. \textbf{PASCAL VOC} and \textbf{COCO} benchmarks
use the mAP calculations of \cite{Everingham:2012:VOC} and \cite{coco}
respectively. \textbf{KITTI(AP50\%)} is mAP calculated at 0.5 IoU
overlap.\label{fig:real_kendall}}
\end{figure*}
\textbf{\emph{COCO 2017 validation set:}} For bounding box detection,
we use different detection models including Faster-RCNN \cite{M-FRCNN},
Single Shot Detector (SSD) \cite{M-SSD} and Regional based Fully
Convolutional Networks (RFCN) \cite{M-RFCN} with different backbones
(Inception Network \cite{M-Inception,M-Inceptionv2}, Residual Network
(ResNet) \cite{M-ResNet}, Inception ResNet \cite{M-InceptionResnet}
with atrous pooling strategy \cite{M-Atrous}, Neural Architecture
Search (NAS) \cite{M-NAS}, Mobilenets \cite{M-MobileNets}, Mobilenets
v2 \cite{M-MobileNets_v2}, Feature Pyramid Network (FPN) \cite{M-FPN}
and Pooling Pyramid Network (PPN) \cite{M-PPN}) to detect objects.
For instance-level segmentation, we use the Mask-RCNN \cite{M-Mask-RCNN}
model with different network structures (FPN, ResNet, Inception ResNet)
and ResNext model \cite{M-ResNext} (with FPN) to produce predictions.

\textbf{\emph{MOTChallenge (MOT17) dataset:}} This experiment ranks
predictions from 21 trackers \cite{T-AFN17,T-AM-ADM17,T-EAMTT,T-EDMT17,T-eHAF17,T-FAMNet,T-FPSN,T-FWT,T-GM-PHD,T-GMPHD-KCF,T-GMPHD-N1Tr,T-GMPHD-SHA,T-HAM-SADF17,T-IOU17,T-JBNOT,T-jCC,T-MHT-bLSTM,T-MOTDT17,T-PHD-GM,T-TLMHT-17,T-tracktor}
on the MOT17 \cite{MOTChallenge:arxiv:2016} leaderboard, according
to various criteria. The tracking results are obtained by applying
the trackers to track human in 7 training sequences and each with
3 detection methods.

\textbf{\emph{Results and discussion:}} The rankings of established
algorithms via traditional criteria are shown in Fig.\ref{fig:teaser}.
For a given task, each ranked algorithm is represented by a unique
color. Rankings for log-AMR, IDF1, and HOTA are given in the appendix
(Section 5). \Fig \ref{fig:real_ranks} shows the rankings of these
algorithms via the suggested metrics. Observe from \Fig \ref{fig:real_ranks},
that the same metric with IoU and GIoU base-distances (for bounding
boxes detection and multi-object tracking) produce similar ranking
order. In addition, rankings amongst different metrics also tend to
be similar to each other, especially in the segmentation task. Analogous to \Fig\ref{fig:sanity_Man}, \Fig \ref{fig:real_kendall}
shows the differences between the rankings of traditional and metric
criteria, in terms of the normalized Kendall-tau distance from the
OSPA rankings (given they have the lowest ranking discrepancy as shown
in \Fig\ref{fig:sanity_Man}). The behaviors of performance criteria
shown in \Fig \ref{fig:real_kendall} corroborate their behaviors
in the sanity tests (\Fig\ref{fig:sanity_Man} and \Tab\ref{tab:sanity_Man}).

In the detection and segmentation tasks, the difference between EMD
and OSPA rankings is smaller than that between Hausdorff and OSPA rankings.
The ranking distances (from OSPA) are large at low and high extreme
thresholds for mAP and log-AMR. The difference between COCO benchmark
(averaging mAP over IoU between 0.5 and 0.95) and OSPA rankings is
smaller than that between PASCAL VOC/KITTI(AP50\%) benchmark (mAP with
IoU of 0.5) and OSPA rankings. The mAP ranking distance at its optimal
threshold (respecting to OSPA rankings) is smaller than the distance
between COCO and OSPA rankings. These behaviors agree with the sanity
test results. The IoU thresholds at which mAP and log-AMR rankings
are the closest to of OSPA occur at around 0.8 for both detection
and segmentation tasks; in the sanity test, this threshold is around
0.4. This can be explained by the variation in prediction sets
quality. In the MOTChallenge experiment, the differences between IDF1/HOTA
and OSPA rankings are similar and lower than that between MOTA and OSPA rankings
at all thresholds. MOTA, IDF1 and HOTA rankings diverge from those of
OSPA at the high extreme threshold while being closer at low thresholds.
With the IoU base-distance, Hausdorff and EMD rankings are close to those of OSPA. With the GIoU base-distance, the difference between Hausdorff and OSPA rankings
is higher than those of between OSPA and IDF1/HOTA/EMD rankings.
These trends are similar to the sanity test results in \Fig\ref{fig:sanity_Man}. \revise{For completeness, rankings of different algorithms on real benchmark datasets evaluated with the OSPA metric at different cut-off values are also provided in the appendix Section 5.}

\section{Conclusions}

We have suggested the notion of trustworthiness for performance evaluation
criteria in basic vision problems by requiring them to be mathematically
consistent, meaningful and reliable. We also suggested some metrics
for sets of shapes as mathematically consistent and reliable alternatives
over the (neither mathematically consistent nor reliable) traditional
criteria, and assessed their meaningfulness. Our experiments indicated
that metrics which capture the intuition behind traditional criteria
are more meaningful than other metrics and the traditional criteria.
This also means that the most meaningful metric is indeed the most
trustworthy because it is also mathematically consistent and reliable
(by default). While our study is by no means comprehensive, we hope
it paves the way towards a richer and versatile set of performance
evaluation tools for computer vision.

\section{Acknowledgments}

This work was supported by the Western Australia DSC Collaborative Research Funding Scheme (2020) and the
Australian Research Council under Discovery Project DP170104584.

% \ifCLASSOPTIONcaptionsoff   
% 	\newpage 
% \fi

\clearpage
%%%%%%%%%%%%%%%%%%%%%%%%%%%%%%%%%%%%%%%%%%%%%%%%%%%%%%%%%%%%%%%%%%%%%%%%%%%%%%%%%%%%%%%%%
\section*{Appendix}
\setcounter{section}{0}
\section{On Traditional Performance Criteria\label{sec:trad_criteria}}

In this section, we show that criteria based on the notion of true
positives violate the triangle inequality and identity property. For
a similarity measure $s$, we define its corresponding dissimilarity
measure between a reference set $\{x\}$ and a prediction set $\{y\}$
as $d_{s}(\{x\},\{y\})=1-s(\{x\},\{y\})$. For traditional set similarity
measures, this form of dissimilarity measure has the same property
as the abstract counterpart defined in the 1-D counter example at
the end of Section 3.3 of the main text. If $x$ and $y$ are bounding
boxes, the distance $\left|x-y\right|$ can be defined as IoU or GIoU
distance (denoted $d_{IoU}(x,y)$ or $d_{GIoU}(x,y)$).

\begin{figure}[ht!]
\centering{}\includegraphics[width=0.2\textwidth]{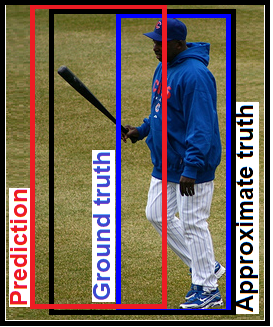}
\caption{Ground truth, approximate truth and prediction bounding boxes for
demonstration of the inconsistency of the traditional criteria.\label{fig:cascaded_rects} }
\end{figure}

\emph{F1-score:} For the example in \Fig~\ref{fig:cascaded_rects}, we can assume
that there exists an IoU (or GIoU) distance threshold $\theta$ such
that (i) the bounding box $x$ can be considered as a true positive
for the bounding box $y$ (\ie $d_{IoU}(x,y)<\theta$), (ii) the
bounding box $y$ can be considered as a true positive for the bounding
box $z$ (\ie $d_{IoU}(y,z)<\theta$), (iii) but the bounding box
$x$ is a false positive for the bounding box $z$ (\ie $d_{IoU}(x,z)>\theta$).
Therefore, in both pairs of scenarios $(x,y)$ and $(y,z)$, the precision,
recall and consequently F1 score values are equal to one, \ie $d_{F1}(\left\{ x\right\} ,\left\{ y\right\} )=d_{F1}(\left\{ y\right\} ,\left\{ z\right\} )=0$.
However, in the pair scenario $(x,z)$, precision, recall and consequently
$F1$ scores are equal to zero, \ie $d_{F1}(\left\{ x\right\} ,\left\{ z\right\} )=1$.
Therefore, F1 score, as dissimilarity measure, does not fulfill the
following metric properties: 
\begin{itemize}
\item (Identity) $d_{F1}(\left\{ x\right\} ,\left\{ y\right\} )=d_{F1}(\left\{ y\right\} ,\left\{ z\right\} )=0$,
but $x\neq y\neq z$ ; 
\item (Triangle inequality) $\underbrace{d_{F1}(\left\{ x\right\} ,\left\{ z\right\} )}_{1}>\underbrace{d_{F1}(\left\{ x\right\} ,\left\{ y\right\} )}_{0}+\underbrace{d_{F1}(\left\{ y\right\} ,\left\{ z\right\} )}_{0}$. 
\end{itemize}
By altering the reference and prediction sets, it can be shown that
the value of precision and recall are switched. However, $F1$ is
symmetrical between the precision and recall and therefore it has
the symmetry property.

\emph{Average Precision (AP):} For the example in \Fig~\ref{fig:cascaded_rects}, with one prediction
and reference in each scenario, AP is turned into the calculation
of the precision only~\footnote{there is a single prediction with an arbitrary score. Therefore, there
exists no range for the confidence score.}. Following the same argument given for F1, precision,  $p$, is equal 1 for the pair
scenarios $(x,y)$ and $(y,z)$, but $p=0$ for the pair scenario
$(x,z)$. Consequently, $d_{AP}$ does not fulfill identity and triangle
inequality as 
\begin{itemize}
\item (Identity) $d_{AP}(\left\{ x\right\} ,\left\{ y\right\} )=d_{AP}(\left\{ y\right\} ,\left\{ z\right\} )=0$,
but $x\neq y\neq z$ ; 
\item (Triangle inequality) $\underbrace{d_{AP}(\left\{ x\right\} ,\left\{ z\right\} )}_{1}>\underbrace{d_{AP}(\left\{ x\right\} ,\left\{ y\right\} )}_{0}+\underbrace{d_{AP}(\left\{ y\right\} ,\left\{ z\right\} )}_{0}$. 
\end{itemize}
The approximated $\widetilde{AP}$ dissimilarity measure also trivially
violates the above metric properties in the same example. Moreover,
AP as area under precision-recall curve in exact form
is symmetrical, but this property cannot be guaranteed in the approximation,
\ie, 
\begin{itemize}
\item $d_{\widetilde{AP}}(\left\{ x\right\} ,\left\{ y\right\} )\neq d_{\widetilde{AP}}(\left\{ y\right\} ,\left\{ x\right\} )$
$\forall x,y\in\mathbb{X}$, where $\mathbb{X}$ is the space of all
possible predictions. 
\end{itemize}
Note that, as mAP is the average of AP over all classes, it is also
not a (mathematical) metric.

\emph{Log-Average Miss Rate (log-AMR):} We define the dissimilarity measure form of log-AMR as itself, \ie,
$d_{\text{Log-AMR}}=AMR$. From the formulation, this dissimilarity
measure has the same property as the abstract dissimilarity measure
defined in the 1-D counter example in the main text (for the pair of two singleton
sets). In \Fig \ref{fig:cascaded_rects}, for the pair scenarios
$(x,y)$ and $(y,z)$, both the miss rate and false positive per image rate (FPPI) rate are zero hence
$d_{\text{Log-AMR}}(\left\{ x\right\} ,\left\{ y\right\} )=d_{\text{Log-AMR}}(\left\{ y\right\} ,\left\{ z\right\} )=0$.
For the pair scenario $(x,z)$, the miss rate and FPPI rate are both
1 hence $d_{\text{Log-AMR}}(\left\{ x\right\} ,\left\{ z\right\} )=1$.
Therefore, the triangle inequality and identity property do not hold. 
\begin{itemize}
\item (Identity) $d_{\text{Log-AMR}}(\left\{ x\right\} ,\left\{ y\right\} )=d_{\text{Log-AMR}}(\left\{ y\right\} ,\left\{ z\right\} )=0$,
but $x\neq y\neq z$ ; 
\item (Triangle inequality) $\underbrace{d_{\text{Log-AMR}}(\left\{ x\right\} ,\left\{ z\right\} )}_{1}>\underbrace{d_{\text{Log-AMR}}(\left\{ x\right\} ,\left\{ y\right\} )}_{0}+\underbrace{d_{\text{Log-AMR}}(\left\{ y\right\} ,\left\{ z\right\} )}_{0}$. 
\end{itemize}
In addition, as the averaging step to calculate log-AMR is carried
out over a finite samples of FPPI rate, the symmetrical property cannot
be guaranteed, \ie, 
\begin{itemize}
\item $d_{\text{Log-AMR}}(\left\{ x\right\} ,\left\{ y\right\} )\neq d_{\text{Log-AMR}}(\left\{ y\right\} ,\left\{ x\right\} )$
$\forall x,y\in\mathbb{X}$, where $\mathbb{X}$ is the space of all
possible predictions. 
\end{itemize}
Further, AP and log-AMR rely on the greedy assignment to match
the true to the predicted objects. This approach is indeed sub-optimal
as the score and the geometrical similarity of the objects are treated
independently, where the geometrical matches are conditioned on the
order of the confidence score. To this extent, in Section 4.1,
via our proposed IoU/GIoU extension to confidence score, we introduce
a new approach to compute AP and log-AMR optimally which is shown
to produce more meaningful predictions ranks in the experiment in
Section \ref{sec:extended_exp_optimal_mAP}.

\emph{MOTA:} \revise{Consider unit-length tracks}, following the same argument as above, the bounding box (as a single
frame track) $x$ can be considered as a true positive for the track
$y$ ($FP_{t}=FN_{t}=IDSW_{t}=0$ and $d_{MOTA}(\left\{ x\right\} ,\left\{ y\right\} )=0$),
and the track $y$ can be considered as a true positive for the track
$z$ ($FP_{t}=FN_{t}=IDSW_{t}=0$ and $d_{MOTA}(\left\{ y\right\} ,\left\{ z\right\} )=0$) \revise{(where $FP_{t}$, $FN_{t}$, and $IDSW_{t}$ are respectively the numbers of false positive, false negative and ID switches at time $t$)}.
However, the track $x$ is considered as false positive for the track
$z$; therefore, there is one false positive and false negative ($FP_{t}=FN_{t}=1$
and $d_{MOTA}(\left\{ y\right\} ,\left\{ z\right\} )=2$). Consequently,
$MOTA$ does not fulfill metric properties, \ie, 
\begin{itemize}
\item (Identity) $d_{MOTA}(\left\{ x\right\} ,\left\{ y\right\} )=d_{MOTA}(\left\{ y\right\} ,\left\{ z\right\} )=0$,
but $x\neq y\neq z$ ; 
\item (Triangle inequality) $\underbrace{d_{MOTA}(\left\{ x\right\} ,\left\{ z\right\} )}_{2}>\underbrace{d_{MOTA}(\left\{ x\right\} ,\left\{ y\right\} )}_{0}+\underbrace{d_{MOTA}(\left\{ y\right\} ,\left\{ z\right\} )}_{0}.$ 
\end{itemize}
Due to its sequential process to indicate ID switches over time, it
can be also shown that $MOTA$ does not fulfill the symmetry property,
\ie, 
\begin{itemize}
\item $d_{MOTA}(\left\{ x\right\} ,\left\{ y\right\} )\neq d_{MOTA}(\left\{ y\right\} ,\left\{ x\right\} )$
$\forall x,y\in\mathbb{T}$ where $\mathbb{T}$ is the space of all
possible predicted tracks. 
\end{itemize}
\emph{IDF1:} Similar
to the MOTA example, IDF1 dissimilarity measure between pairs of single-frame
tracks $(x,y)$ and $(y,z)$ are $d_{IDF1}(\left\{ x\right\} ,\left\{ y\right\} )=d_{IDF1}(\left\{ y\right\} ,\left\{ z\right\} )=0$
as the numbers of false negative ID and false positive ID are 0 and
the number of true positive ID is 1. For the pair of single-frame
track $(x,z)$ the IDF1 dissimilarity measure is $d_{IDF1}(\left\{ x\right\} ,\left\{ z\right\} )=1$
as the number of true positive ID is 1 and there are no false positive
ID and false negative ID. Hence the IDF1 in dissimilarity measure
form violates the following metric properties: 
\begin{itemize}
\item (Identity) $d_{IDF1}(\left\{ x\right\} ,\left\{ y\right\} )=d_{IDF1}(\left\{ y\right\} ,\left\{ z\right\} )=0$,
but $x\neq y\neq z$ ; 
\item (Triangle inequality) $\underbrace{d_{IDF1}(\left\{ x\right\} ,\left\{ z\right\} )}_{1}>\underbrace{d_{IDF1}(\left\{ x\right\} ,\left\{ y\right\} )}_{0}+\underbrace{d_{IDF1}(\left\{ y\right\} ,\left\{ z\right\} )}_{0}.$ 
\end{itemize}
\emph{HOTA:} For the HOTA score defined in the main text, given that $x$ is
matched with $y$, hence $\sum_{c\in\{TP\}}\mathcal{A}(c)=1$ (as
$TP=\{c_{xy}\}$, \revise{where $c_{xy}$ denotes a true positive match between $x$ and $y$}) hence $d_{HOTA}(\{x\},\{y\})=0$. Similarly,
$d_{HOTA}(\{y\},\{z\})=0$ as $y$ is also matched with
$z$. However, as $x$ is not matched with $z$ then $d_{HOTA}(\{x\},\{z\})=1$
(as $TP=\emptyset$). Hence the $\textrm{HOTA}$
in dissimilarity measure form violates the following metric properties: 
\begin{itemize}
\item (Identity) $d_{HOTA}(\left\{ x\right\} ,\left\{ y\right\} )=d_{HOTA}(\left\{ y\right\} ,\left\{ z\right\} )=0$,
but $x\neq y\neq z$ ; 
\item (Triangle inequality) $\underbrace{d_{HOTA}(\left\{ x\right\} ,\left\{ z\right\} )}_{1}>\underbrace{d_{HOTA}(\left\{ x\right\} ,\left\{ y\right\} )}_{0}+\underbrace{d_{HOTA}(\left\{ y\right\} ,\left\{ z\right\} )}_{0}.$
\end{itemize}

\begin{figure}[H]
\centering{}\includegraphics[width=0.2\textwidth]{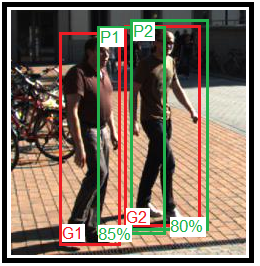}
\caption{As prediction P1 has higher confidence score than P2, it is considered
first and as it has more overlapping with G2, it is matched to G2.
P2 is not matched to any ground truth as G2 has already been taken by P1.
Although it would be more intuitive if P2 is matched to G2 and P1
is matched to G1.\label{fig:greedy_assignment} }
\end{figure}

\emph{Greedy assignment} is used for for mAP and log-AMR calculations.
However, as a sub-optimal algorithm, the greedy assignment is not
intuitive in some scenarios, \ie see the below \Fig \ref{fig:greedy_assignment}.

\section{Ranking Reliability Indicators\label{sec:further_criteria} }

% \revise{Reliability guideline can also be extended to evaluation tasks where the users would like to emphasize on certain aspects of the evaluation. For such tasks, the intent of the evaluation is usually reflected through some parameter choices of the performance criteria. In those cases, the change in ranking should be robust to the change in parameters to guarantee predictable behaviours of the performance criteria. For example, the IoU thresholds in traditional criteria can be used to reflect the intent of the user to the weighting between localization and cardinality errors. Hence, a small change in the threshold (reflecting a small change in the user intent) should not incur large change in the produced rankings. If ranking order is highly sensitive to the change in parameters, one could not reliably select parameters that reflect their evaluation intent.}

In this section, we provide details on three intuitive indicators
that can be used to measure the robustness of a performance criterion
respecting to the variation of parameters. While there are many alternatives
to measure the \textit{ranking consistency}, we are particularly interested in the
\textit{purity} of the ranking order, its \textit{distortion level}
and \textit{sensitivity} to the change of parameter.

Specifically, to measure the purity of the ranks across $m$ independent
parameters, we calculate the average number of ranking switches per
predictions set. For an $m$-D vector $\varrho^{(i)}$ of the ranks
of prediction set $i^{th}$ across $m$ parameters, the number of
ranking switches is given by $R_{S}^{(i)}=\left|\{\varrho^{(i)}[j]:j\in\{1,...,m\}\}\right|-1$.
The \textit{average ranking switches per set} is given by $\overline{R_{S}}=\sum_{i=1}^{K}R_{S}^{(i)}/K$,
where $K$ is the number of predictions sets in consideration.

On the other hand, the degree of distortion of the ranks is reflected
in the standard deviation of the elements of $\varrho^{(i)}$. For
the $i^{th}$ set, the ranking distortion is defined as $R_{std}^{(i)}=\textrm{std}(\varrho^{(i)})$
and \textit{average} \textit{ranking distortion per set} as $\overline{R_{std}}=\sum_{i=1}^{K}R_{std}{}^{(i)}/K$,
where $\textrm{std}(\cdot)$ is the function to calculate the standard
deviation of elements of the vector in its argument.

To indicate the ranking consistency given the sequential nature of
the thresholds, we can measure the sensitivity of the ranking order
against the change of parameters via taking its first order derivative
with respect to the thresholds. In particular, let $\varsigma_{t_{1}}^{(1)},...,\varsigma_{t_{m}}^{(K)}$
be the ranking vectors (tuple of the ranks) of methods $1$ to $K$
across $m$ thresholds from $t_{1}$ (sequentially) to $t_{m}$, the
\textit{average ranking sensitivity} across the set of these $m$
thresholds is defined as $\overline{R_{Sen}}=\sum_{i=1}^{K}\sum_{j=1}^{m-1}\left|\left(\varsigma_{t_{j}}^{(i)}-\varsigma_{t_{j+1}}^{(i)}\right)/\left((t_{j+1}-t_{j})\times(m-1)\times K\right)\right|$.
If the thresholds are evenly spaced the factor $(t_{j+1}-t_{j})$
can be omitted.

% \subsection{Manhattan distance for measuring meaningfulness}

% The intent of measuring the meaningfulness of a criterion is to measure
% the discrepancy between the ranking order provided by the performance
% criterion and the true ranking order. While the Kendall-tau distance
% provided in Section 3.2 of the main text is one obvious and intuitive
% metric for this exercise, another approach is to consider the 2 ranking
% orders as 2 vectors then directly measuring the distance between them.
% A natural metric for such task is the Manhattan distance which reflects
% the total deviation from one ranking order to the other. For completeness,
% in addition to the Kendall-tau results in the main text, in Subsection
% \ref{sec:extended_sanity_results}, we also show the meaningfulness
% of criteria via Manhattan distance.

\section{Further Discussions on Metrics\label{sec:further_metric}}

In the main text, we propose an extension of IoU/GIoU to accommodate
the confidence score implicitly in the calculation (Section 4.1)
and the use of (mathematical) metrics as alternatives for the traditional
performance criteria (Section 4.2). In this section, we present
detailed implementation of the proposed IoU/GIoU extension and further
discussions on the \revise{optimal sub-pattern assignment (OSPA) metric}.

\subsection{Metric for Shapes and Confidence Score\label{sec:extended_GI}}

Traditional IoU and GIoU measures only reflect the similarity between
shapes geometrically but not the confidence scores of the predictions. In the main text, we propose a new method to calculate IoU/GIoU by
extending the shapes to an extra dimension to accommodate the confidence
score (via taking Cartesian product between the shape and corresponding
score). \revise{Specifically, for a set of bounding boxes $\mathbb{B}\subset\mathbb{R}^{N}$ ($N=4$ for 2-D bounding boxes) and the set of confidence
score $\mathbb{S}=(0,1]$, the set of (confidence score) augmented bounding boxes is $\mathbb{B}\times\mathbb{S}$ (where `$\times$' denotes the Cartesian product operation between sets). Visually, for 2-D bounding box scenario, the augmented bounding box is a rectangular box in 3-D. Computing IoU/GIoU distance between augmented bounding boxes can be performed similarly as for standard IoU/GIoU with steps given in Alg. \ref{Alg. Extension_IoU_GIoU}.} This extension of IoU/GIoU to the
confidence score inherits all mathematical properties discussed in
\cite{005}.

As discussed previously, current implementations of AP and log-AMR
rely on the greedy assignment to determine the truth-to-prediction
matches which do not guarantee the optimality of the matches. Basing
on the IoU/GIoU extension, we propose an alternative strategy to compute AP
(mAP) and log-AMR (can be extended to other criteria relying on greedy
assignment). Particularly, we first calculate the pair-wise similarity
scores between true and predicted objects via the IoU/GIoU extension.
We then propose the use of optimal assignment algorithm to determine
the matches. Given the optimal matches and a threshold value, we can
determine the numbers of true positives, false positives, false negatives
and then sort them in the order from the highest to the lowest confidence
score. Subsequently, the standard computation for AP or log-AMR is
carried out. As this approach takes into account both the confidence
score and the geometrical similarity together, the assignment is indeed
optimal. In Section \ref{sec:extended_exp_optimal_mAP}, we show
that it produces more meaningful ranking order compared to the greedy
assignment approach.

\noindent 
\begin{algorithm}[h!]
\noindent {\small{}\rule[0.5ex]{1\columnwidth}{0.5pt}}{\small\par}

\textsf{\textbf{\footnotesize{}Input:}}\textsf{\footnotesize{} }two
arbitrary \textit{N-}D convex shapes, $x$, $y$ and their corresponding
confidence score, $0<s_{x}\leq1$ and $0<s_{y}\leq1$.

\textsf{\textbf{\footnotesize{}Output: }}Standard IoU/GIoU distsance,
$d_{IoU}(x,y)$, $d_{GIoU}(x,y)$; extended IoU/GIoU distance, $d_{\widetilde{IoU}}(x,s_{x},y,s_{y})$,
$d_{\widetilde{GIoU}}(x,s_{x},y,s_{y})$

{\footnotesize{}\smallskip{}
}{\footnotesize\par}

\noindent {\footnotesize{}\rule[0.5ex]{1\columnwidth}{0.5pt}}{\footnotesize\par}

{\footnotesize{}\smallskip{}
}{\footnotesize\par}

\noindent {\footnotesize{}}%
\begin{tabular*}{0.8\columnwidth}{@{\extracolsep{\fill}}l}
For $x$ and $y$, find the smallest enclosing convex object $C$,
then\tabularnewline
\quad{}$IoU=\frac{\left|x\cap y\right|}{\left|x\cup y\right|}$,\tabularnewline
\quad{}$d_{IoU}=1-IoU$,\tabularnewline
\quad{}$GIoU=IoU-\frac{\left|C\setminus(x\cup y)\right|}{\left|C\right|}$,\tabularnewline
\quad{}$d_{GIoU}=\frac{1-GIoU}{2}$.\tabularnewline
\revise{Construct $V_{x}=(x,s_{x})$ and $V_{y}=(y,s_{y})$, the \textit{(N+1)-}D shapes} \tabularnewline
\revise{which are
augmented bounding boxes in $\mathbb{B}\times\mathbb{S}$}.\tabularnewline
For $V_{x}$ and $V_{y}$, find the smallest enclosing convex object
$V_{C}$, then\tabularnewline
\quad{}$\widetilde{IoU}=\frac{\left|V_{x}\cap V_{y}\right|}{\left|V_{x}\cup V_{y}\right|}$,\tabularnewline
\quad{}$d_{\widetilde{IoU}}=1-\widetilde{IoU}$,\tabularnewline
\quad{}$\widetilde{GIoU}=\widetilde{IoU}-\frac{\left|V_{C}\setminus(V_{x}\cup V_{y})\right|}{\left|V_{C}\right|}$,\tabularnewline
\quad{}$d_{\widetilde{GIoU}}=\frac{1-\widetilde{GIoU}}{2}$. \tabularnewline
\end{tabular*}{\footnotesize\par}

{\small{}\caption{IoU/GIoU extension to confidence score\label{Alg. Extension_IoU_GIoU}}
}{\small\par}
\end{algorithm}
\begin{comment}
\begin{algorithm}[!ht]
\caption{}
 \textbf{input: }two arbitrary \textit{N-}D convex shapes, $x$, $y$
and their corresponding confidence score, $0<s_{x}\leq1$ and $0<s_{y}\leq1$.

\textbf{output:} Standard IoU/GIoU distsance, $d_{IoU}(x,y)$, $d_{GIoU}(x,y)$;
extended IoU/GIoU distance, $d_{\widetilde{IoU}}(x,s_{x},y,s_{y})$,
$d_{\widetilde{GIoU}}(x,s_{x},y,s_{y})$

For $x$ and $y$, find the smallest enclosing convex object $C$,
then

\qquad{}$IoU=\frac{\left|x\cap y\right|}{\left|x\cup y\right|}$,

\qquad{}$d_{IoU}=1-IoU$,

\qquad{}$GIoU=IoU-\frac{\left|C\setminus(x\cup y)\right|}{\left|C\right|}$,

\qquad{}$d_{GIoU}=\frac{1-GIoU}{2}$.

Construct $V_{x}$ and $V_{y}$, the \textit{(N+1)-}D shapes which are
Cartesian products of $x$ and $y$ with $s_{x}$ and $s_{y}$ respectively.

For $V_{x}$ and $V_{y}$, find the smallest enclosing convex object
$V_{C}$, then

\qquad{}$\widetilde{IoU}=\frac{\left|V_{x}\cap V_{y}\right|}{\left|V_{x}\cup V_{y}\right|}$,

\qquad{}$d_{\widetilde{IoU}}=1-\widetilde{IoU}$,

\qquad{}$\widetilde{GIoU}=\widetilde{IoU}-\frac{\left|V_{C}\setminus(V_{x}\cup V_{y})\right|}{\left|V_{C}\right|}$,

\qquad{}$d_{\widetilde{GIoU}}=\frac{1-\widetilde{GIoU}}{2}$. 
\end{algorithm}
\end{comment}

\subsection{Optimal Sub-Pattern Assignment Metric\label{sec:point_pattern_metrics}}

Consider a metric space $(\mathcal{\mathbb{W}},\underline{d})$, where
 $\underline{d}:\mathcal{\mathcal{\mathbb{W}}\times}\mathcal{\mathbb{W}}\rightarrow[0,\infty)$
is the\emph{ base-distance} between the elements of $\mathcal{\mathbb{W}}$. In its general form, the OSPA distance
of order $p\geq1$, and cut-off $c>0$, between two point patterns
$X=\{x_{1},...,x_{m}\}$ and $Y=\{y_{1},...,y_{n}\}$ is defined by
\cite{OSPA1} 
\begin{multline}
d_{\mathtt{O}}^{(p,c)}(X,Y)=\\
\left(\frac{1}{n}\left(\min_{\pi\in\Pi_{n}}\sum_{i=1}^{m}\underline{d}^{(c)}\left(x_{i},y_{\pi(i)}\right)^{p}+c^{p}\left(n-m\right)\right)\right)^{\frac{1}{p}},\label{eq:OSPA-dist-c}
\end{multline}
if $n\geq m>0$, and $d_{\mathtt{O}}^{(p,c)}(X,Y)=d_{\mathtt{O}}^{(p,c)}(Y,X)$
if $m>n>0$, where $\Pi_{n}$ is the set of permutations of $\left\{ 1,2,...,n\right\} $,
$\underline{d}^{(c)}(x,y)=\min\left(c,\underline{d}\left(x,y\right)\right)$.
Further $d_{\mathtt{O}}^{(p,c)}(X,Y)=c$ if one of the set is empty,
and $d_{\mathtt{O}}^{(p,c)}(\emptyset,\emptyset)=0$. The order $p$
plays the same role as per the Wasserstein distance \revise{discussed in the main text}, and is taken
to be 1 in this work. The cut-off parameter $c$ provides a weighting
between cardinality and location errors. A large $c$ emphasizes cardinality
error while a small $c$ emphasizes location error. However, a small
$c$ also decreases the sensitivity to the separation between the
points due to the saturation of $\underline{d}^{(c)}$ at $c$.
\begin{figure}[h]
\centering{}\includegraphics[scale=0.4]{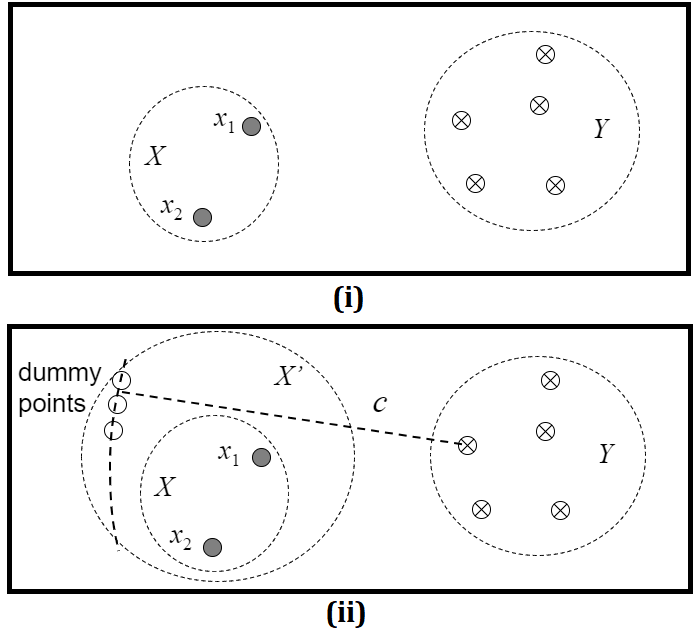}
\caption{OSPA distance between $X$ and $Y$ as the average distance between
the best pairing of the points of $X'$ and $Y$. \label{fig:OSPA_points-1}}
\end{figure}
The general OSPA distance above yields the following base-distance
between two tracks tracks $f$ and $g$: 
\begin{align*}
\underline{d}^{\left(c\right)}\left(f,g\right)= & \sum\limits _{t\in\mathcal{D}_{f}\cup\mathcal{D}_{g}}\!\frac{d_{\mathtt{O}}^{\left(c\right)}\left(\left\{ f\left(t\right)\right\} ,\left\{ g\left(t\right)\right\} \right)}{\left|\mathcal{D}_{f}\cup\mathcal{D}_{g}\right|},
\end{align*}
if $\mathcal{D}_{f}\cup\mathcal{D}_{g}\neq\emptyset$, and $\underline{d}^{\left(c\right)}\left(f,g\right)=0$
if $\mathcal{D}_{f}\cup\mathcal{D}_{g}=\emptyset$, where $d_{\mathtt{O}}^{\left(c\right)}$
denotes the OSPA distance (the order parameter $p$ is redundant because
only sets of at most one element are considered) \cite{OSPA2}. Note
that, apart from the tracking error over the entire scenario, the
OSPA\textsuperscript{(2)} distance (OSPA distance with the above
base-distance) between two sets of tracks can be plotted against time.
Two algorithms with similar OSPA\textsuperscript{(2)} errors over
the entire scenario, may exhibit different OSPA\textsuperscript{(2)}
error curves over time. The monitoring of the tracking performance
over time is important for the analysis/diagnosis of tracking algorithms.
We refer the interested reader to \cite{OSPA2} for more details.

The OSPA distance treats a cardinality error as if the set with smaller
cardinality contained an additional (dummy) point separated from the
remaining set by a base-distance of at least $c$. For an IoU/GIoU
base-distance, such dummy point does not exist when the cut-off $c>1$,
because the largest possible separation between any two points in
$\mathbb{W}$ is 1. Hence, there is no physical meaning in penalizing
a cardinality error with an IoU/GIoU base-distance of $c>1$. On the
other hand, \revise{for evaluation tasks where the users do not give any preference to either localization or cardinality error}, to ensure sensitivity to all IoU/GIoU base-distance separations,
we require $c\geq1$. Consequently, for an IoU/GIoU base-distance,
the best cut-off choice for the OSPA distance is $c=1$, as per \Eq
3 of Section 4 of the main text. 

\revise{For evaluation tasks where it is important to emphasize on either localization or cardinality error, a cut-off $c<1$ can be used. The smaller the value of $c$, the less sensitive to localization error since any pairs with base-distance greater than $c$ is counted as a cardinality mismatch (distance saturated at $c$).
Indeed, this cut-off parameter can be interpreted in a similar light to the IoU threshold in traditional criteria. However, unlike traditional criteria, localization error can also be measured for matched pairs with base-distance lower than the cut-off. Traditional criteria can only count matched pairs as true positives without penalizing the actual localization error. For completeness, in Sections 4.3 and 5 (of this appendix), we also show the error and the corresponding rankings produced by OSPA metric at different cut-off values.}

\section{Further Details on Sanity Tests\label{sec:extended_sanity_results}}
In the main text, we briefly discuss how we set up the sanity tests. In this section, we detail the constructions of the sanity tests and provide further insights on the results.
\subsection{Sanity Test for Multi-Object Detection\label{subsec:exp_detection_sanity}}
We first uniformly sample a reference set of $N_{D}$ bounding boxes
(capped at maximum 40 boxes) with centroid range $[-200,200]\times[-200,200]$
and size range $[20,40]$. We then generate $20$ sets of predictions
(produced by 20 hypothetical detectors) by perturbing the reference
set. In this test, the perturbations are dislocation of centroid,
scaling of size, mis-detections, state-dependent falses, and random
falses. In the multi-class test, each predicted bounding box has an
additional confidence score between $0$ and $1$, and each true box
is assigned a random enumerated class between $1$ and $5$ (true
boxes have a score of one). The additional perturbations for the multi-class
test include the mis-classifications and the reduction of confidence
score (from 1) for the correctly predicted object (class).

To simulate dislocation, we assign each reference box with an enumerated
label. For the box with enumerated label $n$ in the $k^{th}$ prediction
set, we set its centroid dislocation magnitude to $d^{(k)}(n)=a^{(k)}n$,
where $a^{(k)}$ is a unique constant. The centroid dislocation vector
is set to 
\begin{eqnarray}
\left[\begin{array}{c}
\Delta_{x}^{(n,k)}\\
\Delta_{y}^{(n,k)}
\end{array}\right] & = & \left[\begin{array}{c}
ud^{(k)}(n)\\
\sqrt{\left(d^{(k)}(n)\right)^{2}-\left(\Delta_{x}^{(n,k)}\right)^{2}}
\end{array}\right],
\end{eqnarray}
where $u$ is a random number between $0$ and $1$. Next we sample
a random 2-D vector $u_{2}$ whose elements lie between $0$ and $1$.
If $u_{2}[1]<0.5$ then $\Delta_{x}^{(n,k)}=-\Delta_{x}^{(n,k)}$
and if $u_{2}[2]<0.5$ then $\Delta_{y}^{(n,k)}=-\Delta_{y}^{(n,k)}$.
In the multi-class detection experiment, its confidence score is scaled
by $1-r^{(k)}(n)$, where $r^{(k)}(n)=b^{(k)}n$, and $b^{(k)}$ is
a constant associated with the prediction set $k$. For $20$ sets
of predictions, we use $a^{(k)}=D[k]/N_{D},$where $D$ is a $20$-D
vector whose elements are evenly spaced (in ascending order) numbers
from $10$ to $20$. Similarly, $b^{(k)}=S[k]/N_{D}$ where $S$ is
another $20$-D vectors whose elements are evenly spaced (in ascending
order) numbers from $0.2$ to $0.8$. In this test, each box has a
small random disturbance on their size.

Perturbation involving falses and mis-detections are introduced from
the $11^{th}$ prediction set. For each experiment, we sample the
10-D vectors, $P_{D}$, $P_{C}$ uniformly within the range $[0.5,0.95]$,
$F_{S}$ uniformly within the range $[0.05,0.5]$, and $F_{R}$ from
Poisson distributions with respective rates $1,2,...,10$. The elements of $P_{D}$, $P_{C}$ are then sorted in descending
order and elements of $F_{S}$, $F_{R}$ are sorted in ascending order.
To simulate state-dependent falses in detector $k$, we first set
the number of falses to $N_{F_{R}}^{(k)}=\textrm{round}(N_{D}F_{S}[k])$
(where $\textrm{round}(\cdot)$ rounds its argument to the nearest
non-negative whole number). If an object is chosen to have state-dependent
false, we generate a false object with the same dislocation magnitude
and confidence score as the corresponding predicted object. Amongst
the remaining objects (not having state-dependent falses), we simulate
mis-detection by discarding the $N_{M}^{(k)}=\max((N_{D}-N_{F_{R}}^{(k)})(1-P_{D}[k]),0)$
objects with the largest enumerated labels, \ie, objects with the
highest distortion magnitudes and lowest confidence scores. The $F_{R}[k]$
false positive boxes are sampled using the same procedure as that
per the reference boxes. For the multi-class test, we choose the $N_{C}^{(k)}=\max((N_{D}-N_{F_{R}}^{(k)}-N_{M}^{(k)})(1-P_{C}[k]),0)$
objects with largest enumerated labels to be mis-classified objects.

\begin{figure*}[h!]
\centering{}\includegraphics[width=1\textwidth]{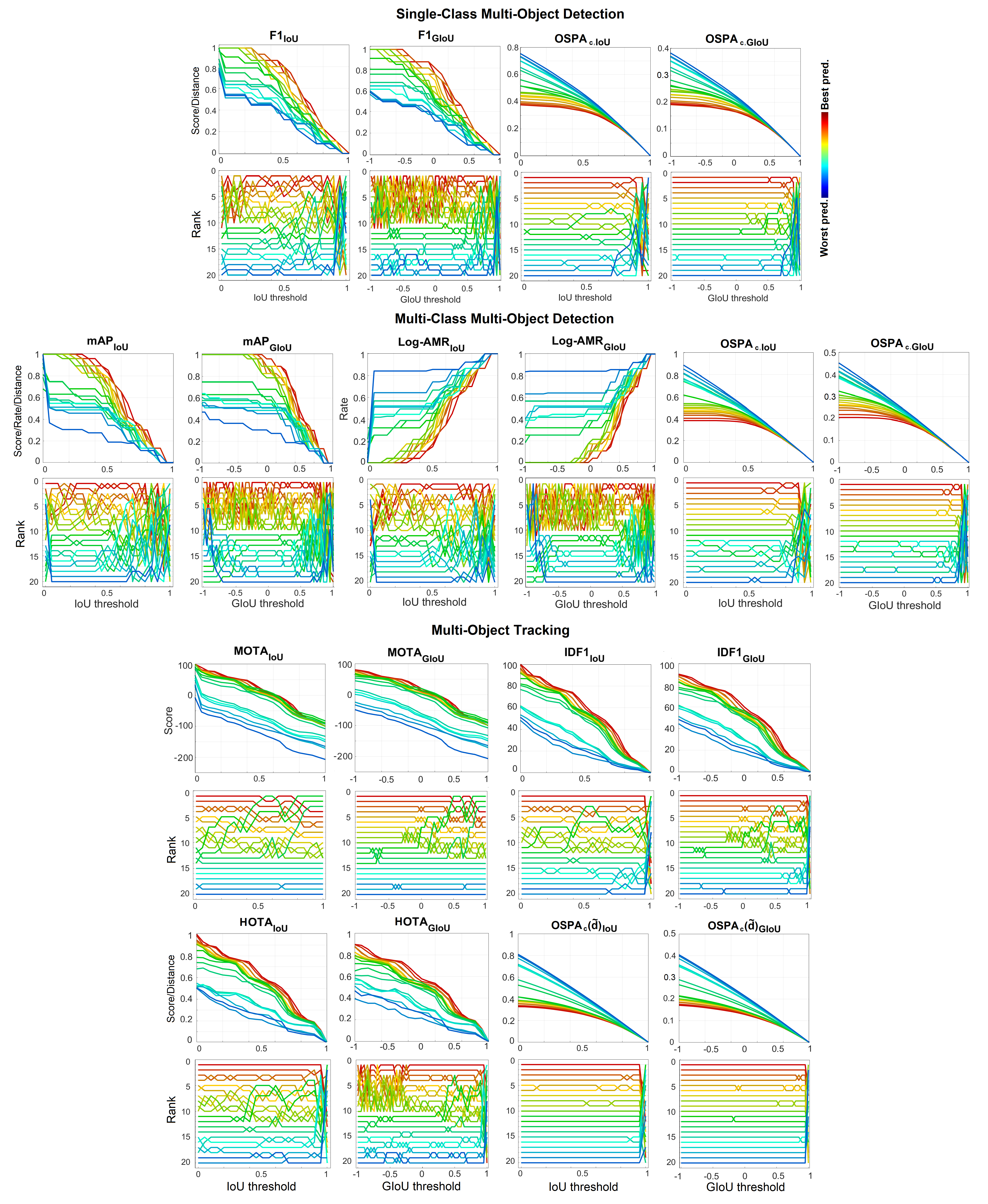}
\caption{Scores/rates and corresponding ranks of prediction sets (for a sample
reference set) according to various criteria over a range of IoU/GIoU
thresholds. The pre-determined ranks are color-coded from worst (blue)
to best (red).\label{fig:sanity_score_rank} } 
\end{figure*}

\subsection{Sanity Test for Multi-Object Tracking\label{subsec:exp_tracking_sanity}}

For tracking sanity tests, we set the tracking window to $100$ time
steps and the number of tracks, $N_{T}$, in the reference set is
randomly sampled between $5$ and $30$. The states of the tracks
are sampled from the space of bounding boxes and each track is assigned
an enumerated label (1 to $N_{T}$). The length of the reference tracks
are sampled between $50$ and $100$ time steps. The initial time
of the track is then sampled between $1$ and the latest possible
initial time step conditioned on its length. The initial centroids
of the tracks are sampled from the region $[-200,200]\times[-200,200]$.

For the initial size, we set a linear correlation between the height
and the sampled initial y-coordinate of the tracks such that the height
is limited within the range $[20,40]$ and the higher the y-coordinate
the lower the height. After the height is generated, the width is
then generated by multiplying the height with a random number drawn
from the interval $[0.5,1.5]$. To generate the initial velocity of
the tracks we sample the course angles and speeds uniformly from the
intervals $[0,360]$ and $[1,5]$. 

After initialization, the centroids of the tracks follow a constant
velocity model. To simulate the effect of in-out camera in real tracking
scenarios, we vary the heights of the tracks linearly with their y-velocity,
and cap minimum height at $20$. The width is kept unchanged through
time.

In this test, we generate $20$ sets of predictions (from 20 hypothetical
trackers). The error types considered here are the dislocation of
centroids, size errors, missed tracks, tracks identities confusion
(swapping) and false tracks (both state-dependent and random). Following
the multi-object detection test, for a track with label $n$, predicted
by the $k^{th}$ tracker, the centroid dislocation magnitude (at each
time step) is set to $\tau^{(k)}(n)=\alpha^{(k)}n$, where $\alpha^{(k)}=T[k]/N_{T}$,
$T$ is a $20$-D vector, whose elements are evenly spaced numbers
between $20$ and $40$ (in ascending order). The centroid dislocation
vector (at every instances that the track exists) is set to 
\begin{eqnarray}
\left[\begin{array}{c}
\Delta_{x}^{(n,k)}\\
\Delta_{y}^{(n,k)}
\end{array}\right] & = & \left[\begin{array}{c}
u\tau^{(k)}(n)\\
\sqrt{\left(\tau^{(k)}(n)\right)^{2}-\left(\Delta_{x}^{(n,k)}\right)^{2}}
\end{array}\right],
\end{eqnarray}
where $u$ is a random number between $0$ and $1$. Next we sample
a 2-D vector $u_{2}$ whose elements lie between $0$ and $1$. If
$u_{2}[1]<0.5$ then $\Delta_{x}^{(n,k)}=-\Delta_{x}^{(n,k)}$, and
if $u_{2}[2]<0.5$ then $\Delta_{y}^{(n,k)}=-\Delta_{y}^{(n,k)}$.
We also add small uniform noise to the sizes of objects.

\begin{figure*}
\begin{align}
s(I^{(i,j)},P_{id}[k])=\begin{cases}
0 & I^{(i,j)}\leq15\\
2\times\left(\frac{I^{(i,j)}-15}{P_{id}[k]-15}\right)^{2} & 15\leq I^{(i,j)}\leq\frac{15+P_{id}[k]}{2}\\
1-2\times\left(\frac{I^{(i,j)}-15}{P_{id}[k]-15}\right)^{2} & \frac{15+P_{id}[k]}{2}\leq I^{(i,j)}\leq P_{id}[k]\\
1 & I^{(i,j)}\geq P_{id}[k]
\end{cases}.\label{eq:idswap_likelihood}
\end{align}
\vspace{-5pt}
\end{figure*}

Similar to the detection sanity test, for the first 10 predictions
sets we only perturb individual tracks. From the $11^{th}$ set, perturbations
involving false tracks and missed tracks are introduced. For each
experiment, we sample 10-D vectors $P_{fr}$, $P_{sft}$, and $P_{id}$ uniformly
within the range $[0.05,1]$,
and $P_{rft}$ from Poisson distributions with respective rates $1,2,...,10$.
The elements of $P_{fr}$, $P_{sft}$ and $P_{rft}$ are then sorted
in ascending order and $P_{id}$ in descending order. To simulate
state-dependent falses in tracker $k$, we first set number of tracks
with state-dependent falses to $N_{sft}=N_{T}N_{sft}[k]$. If a track
is chosen (randomly) to have state-dependent falses, we generate an
extra track with the same dislocation magnitude from the truth for
each time step of the predicted track. At time $t$, we simulate missed
track instances by discarding the $N_{fr}^{(t,k)}=N^{(t,k)}P_{fr}[k]$
instances of tracks with the highest enumerated labels, \ie worst
prediction in terms of dislocation. To simulate false tracks, we
introduce $P_{rft}[k]$ additional tracks with a fixed length of $10$
time steps. The initial times are chosen randomly, and the initialization
of these false tracks is carried out as per the reference tracks.
During their active time, the false tracks appear randomly in the tracking region while their sizes vary within the
range $[20,40]$ without any dynamics.

To simulate identities swapping of detected tracks at each time step,
we first calculate the mutual IoU for all pairs of tracks. For a mutual
IoU of $I^{(i,j)}$ between tracks instances labeled $i$ and $j$,
their likelihood of swapping identities is given by \Eq \ref{eq:idswap_likelihood}.
The tracks labels (at current time $t$) are swapped if this likelihood
is above 0.5 and the swapping is performed in order from the pair
with highest mutual IoU to the lowest. 

\subsection{Further Discussion and Results\label{sec:extended_discussion_sanity_results}}

For completeness, both the scores and corresponding ranks produced
by criteria studied in our sanity tests are shown in \Fig \ref{fig:sanity_score_rank}. \revise{Further, we also include OSPA distances and corresponding ranks with different cut-off values. To distinguish this version of OSPA from the one without parameter, we use an additional subscript $c$, \ie $\text{OSPA}_{c}$.}

\revise{For traditional criteria, the ranking plots show a high number ranking switches. Conversely, the score plots demonstrate the rough changes of the score values across the range of thresholds in one trial of the sanity test. In contrast, we observe less ranking switches for $\text{OSPA}_{c}$ metric. Further, $\text{OSPA}_{c}$ distance decreases smoothly when the IoU/GIoU threshold increases (cut-off value decreases). This predictable behavior of $\text{OSPA}_{c}$ allows the users to reliably choose a cut-off threshold that reflects their evaluation intents.}

\revise{In \Tab \ref{tab:kendall_supp}, we show the meaningfulness of $\text{OSPA}_{c}$ (evaluated using Kendall-tau metric) with different threshold settings. Note that a measure results from averaging $\text{OSPA}_{c}$ distances over some thresholds may not be a mathematical metric. Nonetheless, in general $\text{OSPA}_{c}$ is more meaningful than other measures at different threshold settings. Further, the results show that the meaningfulness of $\text{OSPA}_{c}$ metric at the maximum threshold is almost optimal. This demonstrates the advantage of using maximum cut-off value of 1 for generic evaluation tasks discussed in the main text.}

\revise{The ranking reliability indicators shown in \Tab \ref{tab:sanity_cons} confirm the plots in \Fig \ref{fig:sanity_score_rank} which show $\text{OSPA}_{c}$ metric is more reliable than the traditional criteria. This is because $\text{OSPA}_{c}$ metric can also penalize the localization error for matches with base-distances below the cut-off values.}

\begin{table*}

\caption{Monte Carlo means (and standard deviations) of normalized Kendall-tau
ranking errors of various criteria at certain thresholds. The subscripts
of IoU/GIoU indicate the threshold values; \textquotedblleft optimal\textquotedblright{}
threshold is the one with best ranking accuracy; \textquotedblleft M-partial\textquotedblright{}
indicates that the evaluation is done via averaging the score/rate
over the range 0.5 to 0.95 in steps of 0.05. \textquotedblleft M-full\textquotedblright{}
indicates that the evaluation is done via averaging the score/rate
over the entire range of the base-measure (excluded two extreme thresholds). \label{tab:kendall_supp}}

\centering{}%
\begin{tabular}{|>{\centering}p{1.7cm}|>{\centering}p{1.5cm}>{\centering}p{1.5cm}>{\centering}p{1.5cm}>{\centering}p{1.5cm}|>{\centering}p{1.5cm}ccc|}
\hline 
 & {\scriptsize{}$\textbf{IoU}_{\boldsymbol{0.5}}$} & {\scriptsize{}$\textbf{IoU}_{\textbf{optimal}}$} & {\scriptsize{}$\textbf{IoU}_{\textbf{M-partial}}$} & {\scriptsize{}$\textbf{IoU}_{\textbf{M-full}}$} & {\scriptsize{}$\textbf{GIoU}_{\boldsymbol{0}}$} & {\scriptsize{}$\textbf{GIoU}_{\textbf{optimal}}$} & {\scriptsize{}$\textbf{GIoU}_{\textbf{M-partial}}$} & {\scriptsize{}$\textbf{GIoU}_{\textbf{M-full}}$}\tabularnewline
\hline 
\multicolumn{9}{|c|}{\textbf{\scriptsize{}Single-Class Multi-Object Detection: Normalized
Kendall-tau ranking error (in units of $\mathbf{10^{-2}}$)}}\tabularnewline
\hline 
\textbf{\scriptsize{}F1} & {\scriptsize{}$10.0\thinspace\left(8.82\right)$} & {\scriptsize{}$7.33\thinspace\left(5.17\right)$} & {\scriptsize{}$6.68\thinspace(9.39)$} & {\scriptsize{}$2.15\thinspace(1.51)$} & {\scriptsize{}$7.89\thinspace\left(3.05\right)$} & {\scriptsize{}$7.69\thinspace\left(4.90\right)$} & {\scriptsize{}$7.49\thinspace\left(9.36\right)$} & \textbf{\scriptsize{}$\mathbf{2.17\thinspace\left(1.36\right)}$}\tabularnewline
\hline 
\textbf{\scriptsize{}$\boldsymbol{\text{OSPA}_{c}}$} & {\scriptsize{}$5.19\thinspace\left(6.12\right)$} & {\scriptsize{}$\mathbf{1.97\thinspace\left(1.48\right)}$} & {\scriptsize{}$5.69\thinspace(5.81)$} & {\scriptsize{}$2.37\thinspace(1.67)$} & {\scriptsize{}$2.38\thinspace\left(1.67\right)$} & {\scriptsize{}$2.18\thinspace\left(1.51\right)$} & {\scriptsize{}$6.18\thinspace\left(5.52\right)$} & \textbf{\scriptsize{}$2.18\thinspace\left(1.48\right)$}\tabularnewline
\hline 
{\scriptsize{}$\textbf{Hausdorff}$} & \multicolumn{4}{c|}{{\scriptsize{}$17.8\thinspace\left(9.87\right)$}} & \multicolumn{4}{c|}{{\scriptsize{}$22.4\thinspace\left(11.1\right)$}}\tabularnewline
\hline 
{\scriptsize{}$\textbf{EMD}$} & \multicolumn{4}{c|}{{\scriptsize{}$3.88\thinspace\left(1.96\right)$}} & \multicolumn{4}{c|}{{\scriptsize{}$5.16\thinspace\left(3.03\right)$}}\tabularnewline
\hline 
{\scriptsize{}$\textbf{OSPA}$} & \multicolumn{4}{c|}{{\scriptsize{}$\mathbf{1.97\thinspace\left(1.48\right)}$}} & \multicolumn{4}{c|}{{\scriptsize{}$2.22\thinspace\left(1.43\right)$}}\tabularnewline
\hline 
\multicolumn{9}{|c|}{\textbf{\scriptsize{}Multi-Object Multi-Class Detection: Normalized
Kendall-tau ranking error (in units of $\mathbf{10^{-2}}$)}}\tabularnewline
\hline 
{\scriptsize{}$\textbf{mAP}$} & {\scriptsize{}$10.0\thinspace\left(8.90\right)$} & {\scriptsize{}$7.08\thinspace\left(5.56\right)$} & {\scriptsize{}$7.52\thinspace(8.71)$} & {\scriptsize{}$3.62\thinspace(2.65)$} & {\scriptsize{}$9.41\thinspace\left(3.81\right)$} & {\scriptsize{}$7.39\thinspace\left(5.51\right)$} & {\scriptsize{}$8.27\thinspace\left(8.71\right)$} & {\scriptsize{}$4.86\thinspace(3.00)$}\tabularnewline
\hline 
{\scriptsize{}$\textbf{Log-AMR}$} & {\scriptsize{}$9.91\thinspace\left(5.97\right)$} & {\scriptsize{}$8.42\thinspace\left(5.29\right)$} & {\scriptsize{}$6.75\thinspace\left(3.42\right)$} & {\scriptsize{}$4.31\thinspace(2.30)$} & {\scriptsize{}$16.5\thinspace\left(6.57\right)$} & {\scriptsize{}$8.80\thinspace\left(5.46\right)$} & {\scriptsize{}$7.33\thinspace\left(3.69\right)$} & {\scriptsize{}$4.95\thinspace(2.83)$}\tabularnewline
\hline 
\textbf{\scriptsize{}$\boldsymbol{\text{OSPA}_{c}}$} & {\scriptsize{}$4.38\thinspace\left(5.72\right)$} & {\scriptsize{}$\mathbf{1.84\thinspace\left(1.60\right)}$} & {\scriptsize{}$4.56\thinspace\left(5.37\right)$} & {\scriptsize{}$2.06\thinspace(1.64)$} & {\scriptsize{}$2.27\thinspace\left(2.65\right)$} & {\scriptsize{}$\mathbf{1.86\thinspace\left(1.63\right)}$} & {\scriptsize{}$4.78\thinspace\left(5.21\right)$} & {\scriptsize{}$1.97\thinspace(1.68)$}\tabularnewline
\hline 
{\scriptsize{}$\textbf{Hausdorff}$} & \multicolumn{4}{c|}{{\scriptsize{}$5.43\thinspace\left(2.71\right)$}} & \multicolumn{4}{c|}{{\scriptsize{}$6.39\thinspace\left(2.88\right)$}}\tabularnewline
\hline 
{\scriptsize{}$\textbf{EMD}$} & \multicolumn{4}{c|}{{\scriptsize{}$2.80\thinspace\left(1.83\right)$}} & \multicolumn{4}{c|}{{\scriptsize{}$3.50\thinspace\left(2.26\right)$}}\tabularnewline
\hline 
{\scriptsize{}$\textbf{OSPA}$} & \multicolumn{4}{c|}{{\scriptsize{}$1.86\thinspace\left(1.64\right)$}} & \multicolumn{4}{c|}{{\scriptsize{}$2.41\thinspace\left(1.90\right)$}}\tabularnewline
\hline 
\multicolumn{9}{|c|}{\textbf{\scriptsize{}Multi-Object Tracking: Normalized Kendall-tau
ranking error (in units of $\mathbf{10^{-2}}$)}}\tabularnewline
\hline 
{\scriptsize{}$\textbf{MOTA}$} & {\scriptsize{}$5.18\thinspace\left(5.51\right)$} & {\scriptsize{}$1.42\thinspace\left(1.60\right)$} & {\scriptsize{}$7.64\thinspace\left(7.74\right)$} & {\scriptsize{}$3.26\thinspace\left(3.90\right)$} & {\scriptsize{}$2.20\thinspace\left(2.28\right)$} & {\scriptsize{}$1.00\thinspace\left(1.39\right)$} & {\scriptsize{}$8.46\thinspace\left(8.22\right)$} & {\scriptsize{}$0.872\thinspace\left(0.806\right)$}\tabularnewline
\hline 
{\scriptsize{}$\textbf{IDF1}$} & {\scriptsize{}$3.47\thinspace\left(3.51\right)$} & {\scriptsize{}$1.84\thinspace\left(1.63\right)$} & {\scriptsize{}$3.04\thinspace\left(3.00\right)$} & {\scriptsize{}$1.38\thinspace\left(1.54\right)$} & {\scriptsize{}$2.82\thinspace\left(2.47\right)$} & {\scriptsize{}$1.24\thinspace\left(1.68\right)$} & {\scriptsize{}$3.93\thinspace\left(3.62\right)$} & {\scriptsize{}$0.676\thinspace\left(0.920\right)$}\tabularnewline
\hline 
{\scriptsize{}$\textbf{HOTA}$} & {\scriptsize{}$4.11\thinspace\left(4.17\right)$} & {\scriptsize{}$2.95\thinspace\left(2.57\right)$} & {\scriptsize{}$3.56\thinspace\left(3.70\right)$} & {\scriptsize{}$1.45\thinspace\left(1.63\right)$} & {\scriptsize{}$4.34\thinspace\left(3.21\right)$} & {\scriptsize{}$4.24\thinspace\left(3.19\right)$} & {\scriptsize{}$4.19\thinspace\left(4.18\right)$} & {\scriptsize{}$1.02\thinspace\left(1.26\right)$}\tabularnewline
\hline 
\textbf{\scriptsize{}$\boldsymbol{\text{OSPA}_{c}(\tilde{\underline{d}})}$} & {\scriptsize{}$0.869\thinspace\left(0.800\right)$} & {\scriptsize{}$\mathbf{0.518\thinspace\left(0.580\right)}$} & {\scriptsize{}$0.942\thinspace\left(0.843\right)$} & {\scriptsize{}$0.660\thinspace\left(0.682\right)$} & {\scriptsize{}$0.675\thinspace\left(0.670\right)$} & {\scriptsize{}$\mathbf{0.529\thinspace\left(0.567\right)}$} & {\scriptsize{}$1.00\thinspace\left(0.872\right)$} & {\scriptsize{}$0.558\thinspace\left(0.585\right)$}\tabularnewline
\hline 
{\scriptsize{}$\boldsymbol{\mathbf{Hausdorff}(\underline{\tilde{d}})}$} & \multicolumn{4}{c|}{{\scriptsize{}$12.0\thinspace\left(9.64\right)$}} & \multicolumn{4}{c|}{{\scriptsize{}$10.6\thinspace\left(5.27\right)$}}\tabularnewline
\hline 
{\scriptsize{}$\boldsymbol{\mathbf{EMD}(\underline{\tilde{d}})}$} & \multicolumn{4}{c|}{{\scriptsize{}$3.53\thinspace\left(2.38\right)$}} & \multicolumn{4}{c|}{{\scriptsize{}$5.80\thinspace\left(3.29\right)$}}\tabularnewline
\hline 
{\scriptsize{}$\boldsymbol{\mathbf{OSPA}(\underline{\tilde{d}})}$} & \multicolumn{4}{c|}{{\scriptsize{}$\mathbf{0.518\thinspace\left(0.580\right)}$}} & \multicolumn{4}{c|}{{\scriptsize{}$0.539\thinspace\left(0.577\right)$}}\tabularnewline
\hline 
\end{tabular}
\end{table*}
% \begin{figure}[H]
% \centering{}\includegraphics[width=0.45\textwidth]{supp_figures/sanity_man}
% \caption{Mean Manhattan ranking errors (from the true ranking) of various criteria
% at different thresholds, over all Monte Carlo trials. Shaded area
% around each curve indicates 0.2-sigma bound. \label{fig:sanity_Man} }
% \end{figure}
\begin{table*}
\begin{centering}
\caption{Monte Carlo means (and standard deviations) of ranking consistency
indicators over the entire range of IoU/GIoU threshold in the sanity
tests. \label{tab:sanity_cons}}
\par\end{centering}
\begin{centering}
\begin{tabular}{ccccc}
\toprule 
\multicolumn{5}{c}{\textbf{\scriptsize{}Single-Class Multi-Object Detection}}\tabularnewline
\midrule 
 & \textbf{\scriptsize{}$\boldsymbol{\text{F1}_{\text{IoU}}}$} & \textbf{\scriptsize{}$\boldsymbol{\text{OSPA}_{c,\text{IoU}}}$} & {\scriptsize{}$\boldsymbol{\textbf{F1}_{\textbf{GIoU}}}$} & \textbf{\scriptsize{}$\boldsymbol{\text{OSPA}_{c,\text{GIoU}}}$}\tabularnewline
\midrule 
{\scriptsize{}$\boldsymbol{\overline{R_{S}}}$}  & {\scriptsize{}$7.73\thinspace(1.08)$}  & {\scriptsize{}$\mathbf{5.17\thinspace(1.44)}$}  & {\scriptsize{}$8.98\thinspace(1.10)$}  & {\scriptsize{}$5.39\thinspace(1.56)$} \tabularnewline
\midrule 
{\scriptsize{}$\boldsymbol{\overline{R_{std}}}$}  & {\scriptsize{}$3.43\thinspace(0.670)$}  & {\scriptsize{}$2.94\thinspace(0.745)$}  & {\scriptsize{}$2.91\thinspace(0.499)$}  & {\scriptsize{}$\mathbf{2.22\thinspace(0.630)}$} \tabularnewline
\midrule 
{\scriptsize{}$\boldsymbol{\overline{R_{Sen}}}$}  & {\scriptsize{}$4.33\thinspace(1.22)$}  & {\scriptsize{}$2.14\thinspace(1.46)$}  & {\scriptsize{}$3.69\thinspace(0.665)$}  & {\scriptsize{}$\mathbf{1.12\thinspace(0.793)}$} \tabularnewline
\bottomrule
\end{tabular}
\par\end{centering}
\smallskip{}

\begin{centering}
\begin{tabular}{ccccccc}
\toprule 
\multicolumn{7}{c}{\textbf{\scriptsize{}Multi-Class Multi-Object Detection}}\tabularnewline
\midrule 
 & {\scriptsize{}$\boldsymbol{\textbf{mAP}_{\textbf{IoU}}}$}  & {\scriptsize{}$\boldsymbol{\textbf{Log-AMR}_{\textbf{IoU}}}$}  & \textbf{\scriptsize{}$\boldsymbol{\text{OSPA}_{c,\text{IoU}}}$} & {\scriptsize{}$\boldsymbol{\textbf{mAP}_{\textbf{GIoU}}}$}  & {\scriptsize{}$\boldsymbol{\textbf{Log-AMR}_{\textbf{GIoU}}}$} & \textbf{\scriptsize{}$\boldsymbol{\text{OSPA}_{c,\text{GIoU}}}$}\tabularnewline
\midrule 
{\scriptsize{}$\boldsymbol{\overline{R_{S}}}$}  & {\scriptsize{}$8.13\thinspace(1.20)$}  & {\scriptsize{}$7.89\thinspace(1.09)$}  & {\scriptsize{}$\mathbf{4.28\thinspace(1.70)}$}  & {\scriptsize{}$10.0\thinspace(1.31)$}  & {\scriptsize{}$10.4\thinspace(1.48)$}  & {\scriptsize{}$4.51\thinspace(1.97)$} \tabularnewline
\midrule 
{\scriptsize{}$\boldsymbol{\overline{R_{std}}}$}  & {\scriptsize{}$3.53\thinspace(0.664)$}  & {\scriptsize{}$3.25\thinspace(0.602)$}  & {\scriptsize{}$2.54\thinspace(0.855)$}  & {\scriptsize{}$3.19\thinspace(0.534)$}  & {\scriptsize{}$3.27\thinspace(0.588)$}  & {\scriptsize{}$\mathbf{1.93\thinspace(0.739)}$} \tabularnewline
\midrule 
{\scriptsize{}$\boldsymbol{\overline{R_{Sen}}}$}  & {\scriptsize{}$4.38\thinspace(1.42)$}  & {\scriptsize{}$4.18\thinspace(1.48)$}  & {\scriptsize{}$1.74\thinspace(1.59)$}  & {\scriptsize{}$3.86\thinspace(1.13)$}  & {\scriptsize{}$4.12\thinspace(1.40)$}  & {\scriptsize{}$\mathbf{0.952\thinspace(0.980)}$} \tabularnewline
\bottomrule
\end{tabular}
\par\end{centering}
\smallskip{}

\centering{}%
\begin{tabular}{ccccccccc}
\toprule 
\multicolumn{9}{c}{\textbf{\scriptsize{}Multi-Object Tracking}}\tabularnewline
\midrule 
 & {\scriptsize{}$\boldsymbol{\textbf{MOTA}_{\textbf{IoU}}}$}  & {\scriptsize{}$\boldsymbol{\textbf{IDF1}_{\textbf{IoU}}}$}  & {\scriptsize{}$\boldsymbol{\textbf{HOTA}_{\textbf{IoU}}}$}  & \textbf{\scriptsize{}$\boldsymbol{\text{OSPA}_{c}(\tilde{\underline{d}})_{\text{IoU}}}$} & {\scriptsize{}$\boldsymbol{\textbf{MOTA}_{\textbf{GIoU}}}$}  & \multicolumn{1}{c}{{\scriptsize{}$\boldsymbol{\textbf{IDF1}_{\textbf{GIoU}}}$} } & {\scriptsize{}$\boldsymbol{\textbf{HOTA}_{\textbf{GIoU}}}$}  & \textbf{\scriptsize{}$\boldsymbol{\text{OSPA}_{c}(\tilde{\underline{d}})_{\text{GIoU}}}$}\tabularnewline
\midrule 
{\scriptsize{}$\boldsymbol{\overline{R_{S}}}$}  & {\scriptsize{}$2.65\thinspace(1.23)$}  & {\scriptsize{}$3.87\thinspace(1.28)$}  & {\scriptsize{}$4.44\thinspace(1.28)$}  & {\scriptsize{}$\mathbf{1.64\thinspace(0.485)}$}  & {\scriptsize{}$3.31\thinspace(1.41)$}  & {\scriptsize{}$4.53\thinspace(1.37)$} & {\scriptsize{}$7.14\thinspace(1.18)$}  & {\scriptsize{}$1.68\thinspace(0.501)$}\tabularnewline
\midrule 
{\scriptsize{}$\boldsymbol{\overline{R_{std}}}$}  & {\scriptsize{}$1.14\thinspace(0.719)$}  & {\scriptsize{}$2.15\thinspace(0.633)$}  & {\scriptsize{}$2.27\thinspace(0.641)$}  & {\scriptsize{}$1.56\thinspace(0.232)$}  & {\scriptsize{}$1.13\thinspace(0.664)$}  & {\scriptsize{}$1.72\thinspace(0.492)$} & {\scriptsize{}$2.22\thinspace(0.442)$}  & {\scriptsize{}$\mathbf{1.13\thinspace(0.169)}$}\tabularnewline
\midrule 
{\scriptsize{}$\boldsymbol{\overline{R_{Sen}}}$}  & {\scriptsize{}$0.733\thinspace(0.290)$}  & {\scriptsize{}$1.46\thinspace(0.718)$}  & {\scriptsize{}$1.73\thinspace(0.745)$}  & {\scriptsize{}$0.479\thinspace(0.162)$}  & {\scriptsize{}$0.546\thinspace(0.183)$}  & {\scriptsize{}$0.902\thinspace(0.380)$} & {\scriptsize{}$2.11\thinspace(0.439)$}  & {\scriptsize{}$\mathbf{0.247\thinspace(0.0834)}$}\tabularnewline
\bottomrule
\end{tabular}
\end{table*}

\begin{figure*}
\centering{}\includegraphics[width=\textwidth]{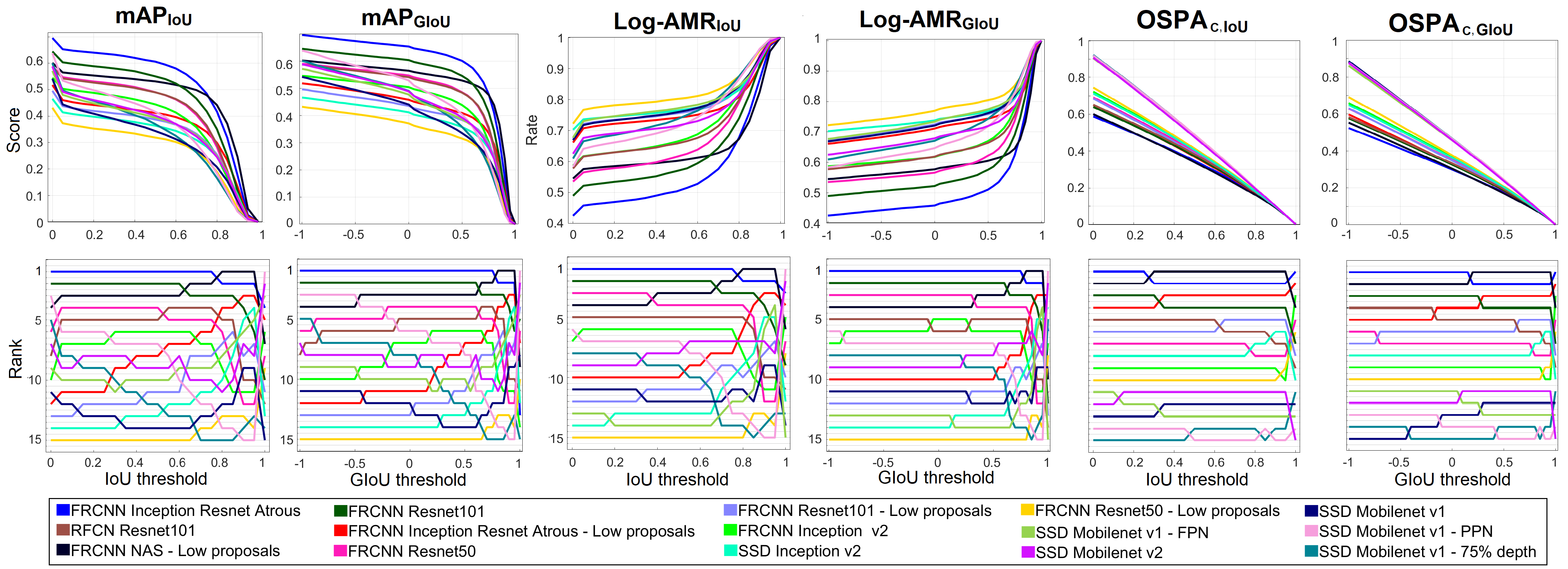}
\caption{Score/rate and ranks of predictions sets according to mAP, log-AMR
over range of IoU/GIoU thresholds in COCO bounding
box detection experiment.\label{fig:COCO_box_score_rank} }
\end{figure*}

\begin{figure*}[h]
\begin{centering}
\includegraphics[width=0.7\textwidth]{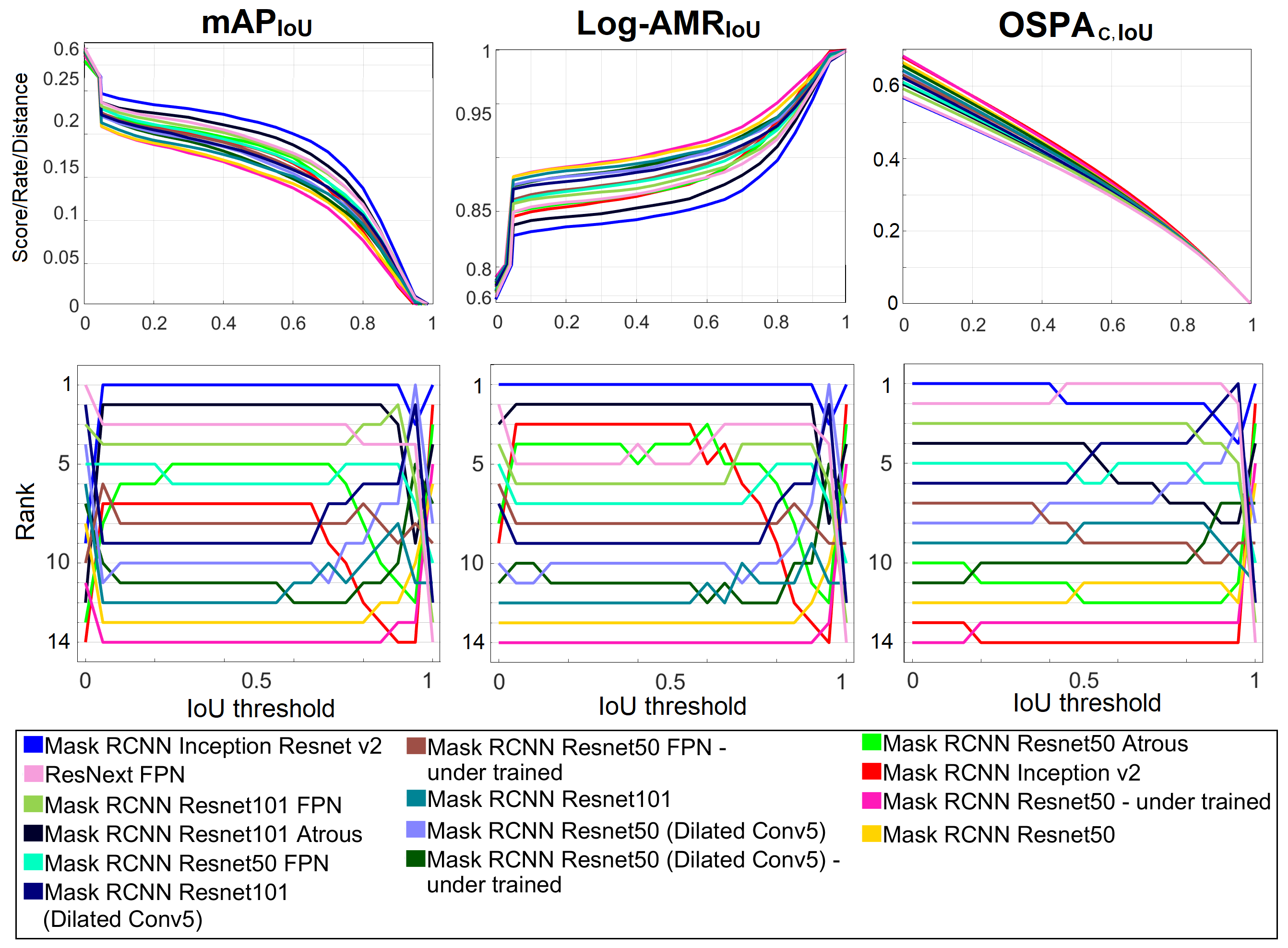} 
\par\end{centering}
\caption{Score/rate and ranks of predictions sets according to mAP, log-AMR
over range of IoU thresholds in COCO instance
level segmentation experiment.\label{fig:COCO_mask_score_rank} }
\end{figure*}

\begin{figure*}
\centering{}\includegraphics[width=0.95\textwidth]{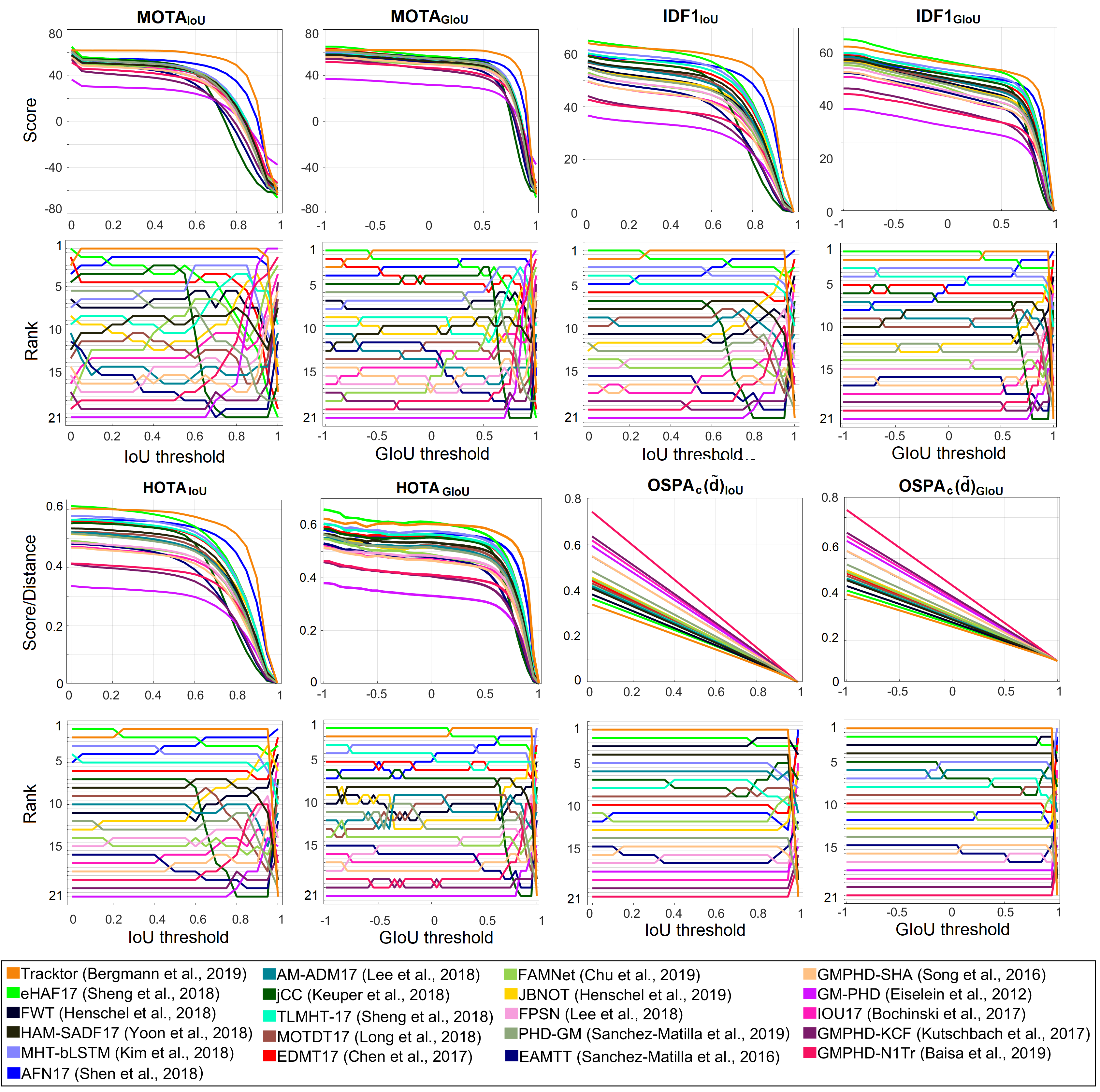}
\caption{Score/rate and ranks of predictions sets according to MOTA, IDF1, and HOTA over
range of IoU/GIoU thresholds in MOT17 tracking
experiment.\label{fig:MOT17_score_rank}}
\end{figure*}

\section{Further on Real Benchmark Datasets Ranking Comparisons \label{sec:extended_exp_real_res} }

In this section, we detail the results on real dataset experiments,
\ie COCO detection with bounding box, COCO instance-level segmentation
and MOTChallenge multi-object tracking to supplement Section 6 of
the main text. \revise{In addition to the ranking plots provided in the main text, we also show the scores and variation of ranks across different thresholds in Figs. \ref{fig:COCO_box_score_rank}, \ref{fig:COCO_mask_score_rank}, and \ref{fig:MOT17_score_rank}. Evaluation results with $\text{OSPA}_{c}$ metric are also included for completeness.}

For the COCO bounding box detection experiment,
we observe that the ranks gradually change over thresholds ranges.
For example, the ``SSD Mobilenet v1 - 75\% depth'' performs relative
well at low threshold but gradually gets worse when the threshold
increases or the ``FRCNN Inception Resnet Atrous - Low Proposals''
performs worse at low thresholds but gets better at higher thresholds. In general, at low thresholds, we observe the ranks are quite stable
but from value of 0.6 onward (both IoU and GIoU) the ranks start to
switch more frequently. This observation is also confirmed in the
log-AMR plot in \Fig 7 of \cite{dollar2011pedestrian}. Conversely, \revise{less ranking variation is observed for $\text{OSPA}_{c}$ metric.} \revise{For COCO instance-level segmentation experiment, \Fig \ref{fig:COCO_mask_score_rank} also shows less drastic changes in rankings for $\text{OSPA}_{c}$ metric compared to traditional criteria.}

\begin{table*}
\begin{centering}
\caption{Ranking consistency indicators over the entire range of IoU/GIoU threshold
in real benchmark experiments.\label{tab:real_rank_cons}}
\par\end{centering}
\begin{centering}
\begin{tabular}{ccccccc}
\toprule 
\multicolumn{7}{c}{\textbf{\scriptsize{}COCO Bounding Box Detection}}\tabularnewline
\midrule 
 & {\scriptsize{}$\textbf{mAP}_{\textbf{IoU}}$} & {\scriptsize{}$\textbf{Log-AMR}_{\textbf{IoU}}$} & {\scriptsize{}$\textbf{OSPA}_{c,\textbf{IoU}}$} & {\scriptsize{}$\textbf{mAP}_{\textbf{GIoU}}$} & {\scriptsize{}$\textbf{Log-AMR}_{\textbf{GIoU}}$} & {\scriptsize{}$\textbf{OSPA}_{c,\textbf{GIoU}}$}\tabularnewline
\midrule 
{\scriptsize{}$\boldsymbol{\overline{R_{S}}}$} & {\scriptsize{}$5.33$} & {\scriptsize{}$4.53$} & {\scriptsize{}$\mathbf{1.93}$} & {\scriptsize{}$5.73$} & {\scriptsize{}$4.27$} & {\scriptsize{}$2.27$}\tabularnewline
\midrule 
{\scriptsize{}$\boldsymbol{\overline{R_{std}}}$} & {\scriptsize{}$2.14$} & {\scriptsize{}$1.98$} & {\scriptsize{}$0.775$} & {\scriptsize{}$1.76$} & {\scriptsize{}$1.45$} & {\scriptsize{}$\mathbf{0.774}$}\tabularnewline
\midrule 
{\scriptsize{}$\boldsymbol{\overline{R_{Sen}}}$} & {\scriptsize{}$0.827$} & {\scriptsize{}$0.673$} & {\scriptsize{}$0.267$} & {\scriptsize{}$0.387$} & {\scriptsize{}$0.373$} & {\scriptsize{}$\mathbf{0.143}$}\tabularnewline
\bottomrule
\end{tabular}
\par\end{centering}
\smallskip{}

\begin{centering}
\begin{tabular}{cccc}
\toprule 
\multicolumn{4}{c}{\textbf{\scriptsize{}COCO Instance-Level Segmentation}}\tabularnewline
\midrule 
 & {\scriptsize{}$\textbf{mAP}_{\textbf{IoU}}$} & {\scriptsize{}$\textbf{Log-AMR}_{\textbf{IoU}}$} & {\scriptsize{}$\textbf{OSPA}_{c,\textbf{IoU}}$}\tabularnewline
\midrule 
{\scriptsize{}$\boldsymbol{\overline{R_{S}}}$} & {\scriptsize{}$4.21$} & {\scriptsize{}$3.71$} & {\scriptsize{}$\mathbf{3.00}$}\tabularnewline
\midrule 
{\scriptsize{}$\boldsymbol{\overline{R_{std}}}$} & {\scriptsize{}$2.06$} & {\scriptsize{}$1.72$} & {\scriptsize{}$\mathbf{1.62}$}\tabularnewline
\midrule 
{\scriptsize{}$\boldsymbol{\overline{R_{Sen}}}$} & {\scriptsize{}$0.950$} & {\scriptsize{}$0.836$} & {\scriptsize{}$\mathbf{0.514}$}\tabularnewline
\bottomrule
\end{tabular}
\par\end{centering}
\smallskip{}

\centering{}%
\begin{tabular}{ccccccccc}
\toprule 
\multicolumn{9}{c}{\textbf{\scriptsize{}MOT17 Multi-Object Tracking}}\tabularnewline
\midrule 
 & {\scriptsize{}$\textbf{MOTA}_{\textbf{IoU}}$} & {\scriptsize{}$\textbf{IDF1}_{\textbf{IoU}}$} & {\scriptsize{}$\textbf{HOTA}_{\textbf{IoU}}^{(\alpha)}$} & {\scriptsize{}$\textbf{OSPA}_{c}(\tilde{\underline{d}})_{\textbf{IoU}}$} & {\scriptsize{}$\textbf{MOTA}_{\textbf{GIoU}}$} & {\scriptsize{}$\textbf{IDF1}_{\textbf{GIoU}}$} & {\scriptsize{}$\textbf{HOTA}_{\textbf{GIoU}}^{(\alpha)}$} & {\scriptsize{}$\textbf{OSPA}_{c}(\tilde{\underline{d}})_{\textbf{GIoU}}$}\tabularnewline
\midrule 
{\scriptsize{}$\boldsymbol{\overline{R_{S}}}$} & {\scriptsize{}$6.05$} & {\scriptsize{}$3.90$} & {\scriptsize{}$4.05$} & {\scriptsize{}$\mathbf{1.90}$} & {\scriptsize{}$6.48$} & {\scriptsize{}$4.10$} & {\scriptsize{}$4.52$} & {\scriptsize{}$2.00$}\tabularnewline
\midrule 
{\scriptsize{}$\boldsymbol{\overline{R_{std}}}$} & {\scriptsize{}$3.52$} & {\scriptsize{}$2.00$} & {\scriptsize{}$2.04$} & {\scriptsize{}$1.37$} & {\scriptsize{}$2.80$} & {\scriptsize{}$1.57$} & {\scriptsize{}$1.65$} & {\scriptsize{}$\mathbf{1.05}$}\tabularnewline
\midrule 
{\scriptsize{}$\boldsymbol{\overline{R_{Sen}}}$} & {\scriptsize{}$0.938$} & {\scriptsize{}$0.695$} & {\scriptsize{}$0.695$} & {\scriptsize{}$0.405$} & {\scriptsize{}$0.526$} & {\scriptsize{}$0.352$} & {\scriptsize{}$0.526$} & {\scriptsize{}$\mathbf{0.210}$}\tabularnewline
\bottomrule
\end{tabular}
\end{table*}

In the MOTChallenge experiment, the ranks switch frequently across
different thresholds.
Especially, it is noticeable that the ``jCC'' method changes the
rank dramatically after threshold of 0.5 on MOTA (IoU) measure. In general, we
observe higher number of ranking switches at the high extreme of the thresholds
ranges which indicates the criteria are more unreliable at high thresholds. \revise{Conversely, we observe the ranking orders are relatively stable across different cut-off values of $\text{OSPA}_{c}$ metric.}

% \Tab \ref{tab:real_rank_cons} shows the consistency indicators for
% all experiments. In COCO bounding box detection experiment, log-AMR
% performs relatively more reliable compared to the mAP although it
% is slightly more sensitive to the change in thresholds compared to
% mAP with IoU base-measure. In COCO segmentation experiment, the ranking
% consistency shows log-AMR is more reliable than the mAP score. In
% MOTChallenge tracking experiment, it is shown that HOTA
% with GIoU base-measure is relatively more reliable compared to MOTA
% and IDF1 but it has slightly more ranking switches than IDF1 with
% IoU base-measure.

\revise{From the results shown in \Tab \ref{tab:real_rank_cons} we observe that  $\text{OSPA}_{c}$ metric is more reliable than the traditional criteria, which quantitatively confirms the observations in Figs. \ref{fig:COCO_box_score_rank}, \ref{fig:COCO_mask_score_rank}, and \ref{fig:MOT17_score_rank}.} \vspace{-10pt}

\begin{figure*}
\centering{}\includegraphics[width=0.95\textwidth]{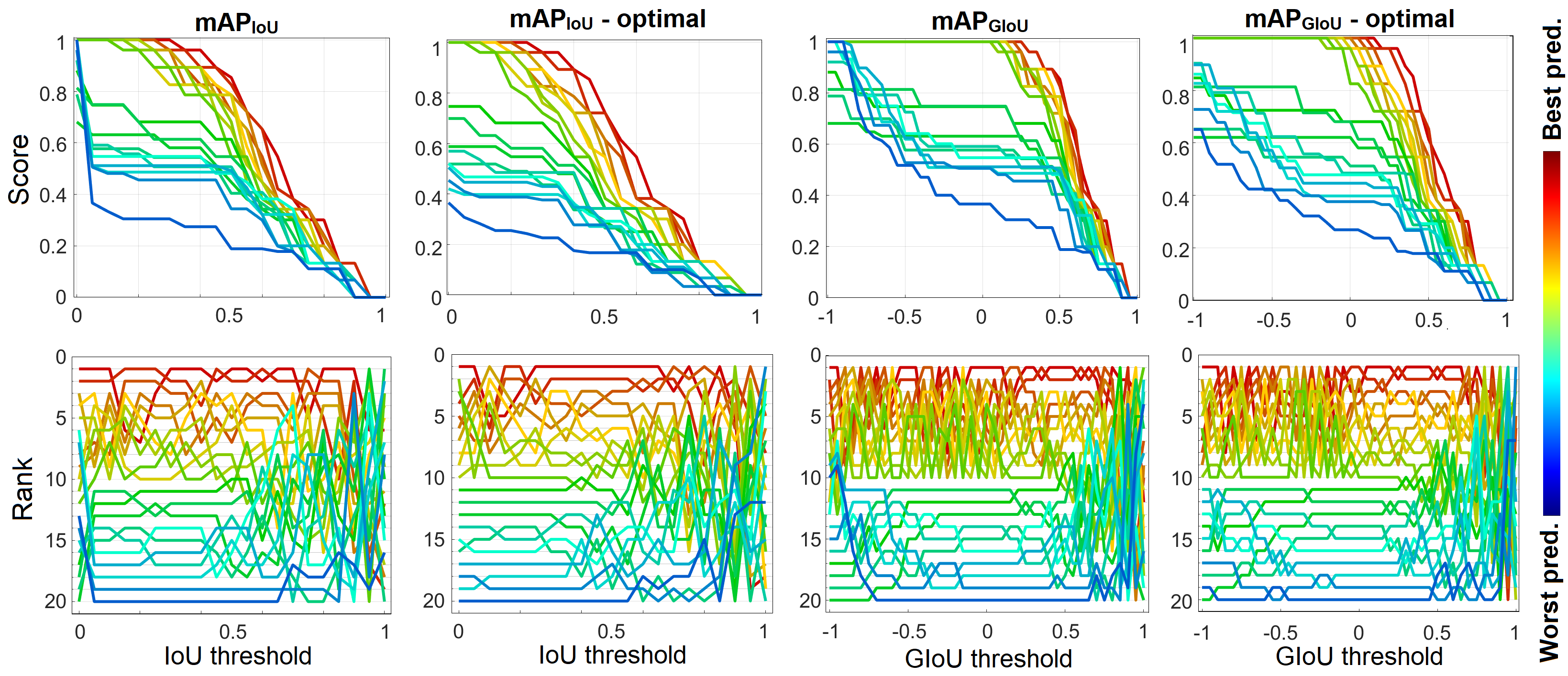}
\caption{mAP scores with greedy and optimal assignment approaches (top row)
and the corresponding ranks of predictions (bottom row) over a range
of IoU/GIoU thresholds in one trial of the multi-class multi-object
detection sanity test. The pre-determined ranks are color-coded from
worst (blue) to best (red). \label{fig:mAP_optimal_score_rank} }
\end{figure*}
\begin{figure*}

\begin{centering}
\includegraphics[width=0.95\textwidth]{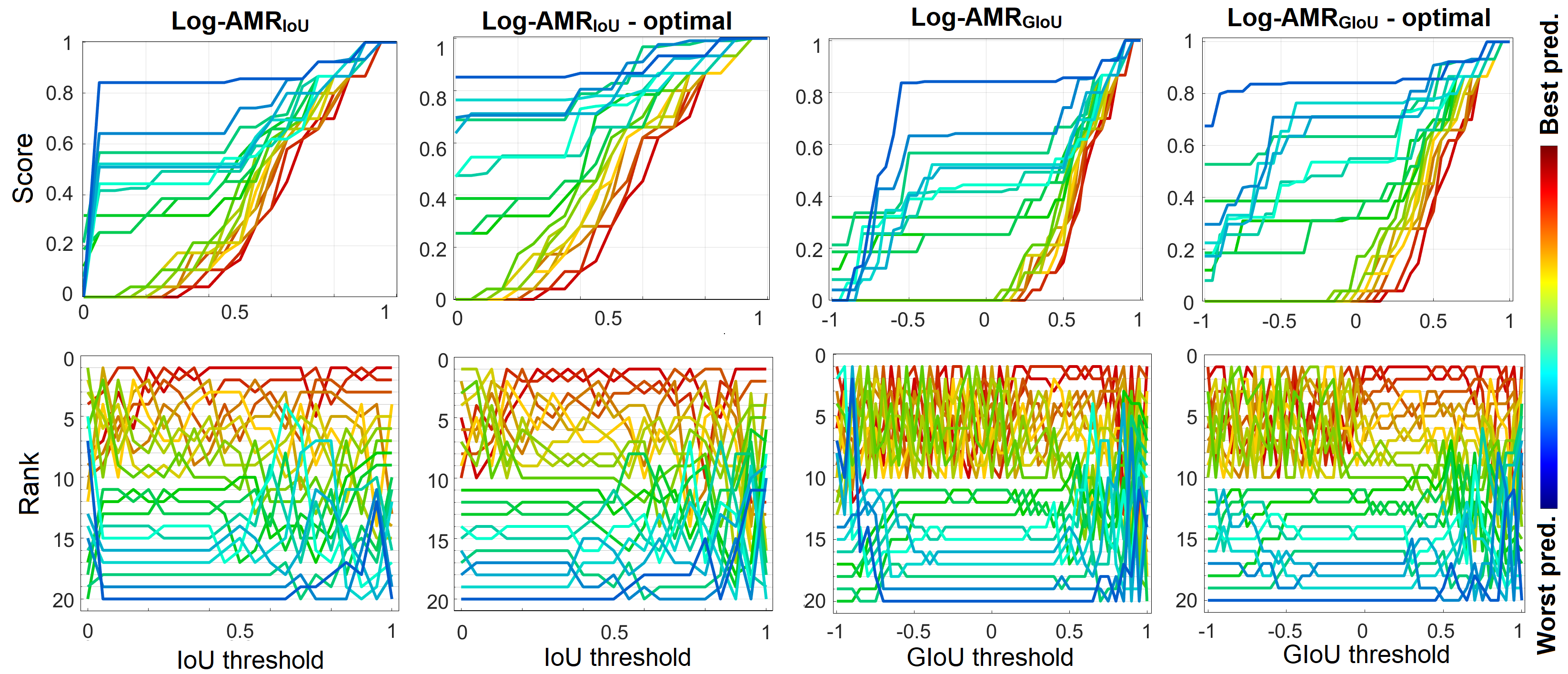} 
\par\end{centering}
\caption{Log-AMR with greedy and optimal assignment approaches (top row) and
the corresponding ranks of predictions (bottom row) over a range of
IoU/GIoU thresholds in one trial of the multi-class multi-object detection
sanity test. The pre-determined ranks are color-coded from worst (blue)
to best (red).\label{fig:AMR_optimal_score_rank} }

\end{figure*}

\section{Optimal Assignment for mAP and Log-AMR\label{sec:extended_exp_optimal_mAP} }

In this experiment, we construct the sanity test in the like-wise
manner to the mentioned multi-class multi-object detection experiment (Section \ref{subsec:exp_detection_sanity}).
We then evaluate the predictions sets on the standard mAP, log-AMR
criteria (with greedy assignment) and their corresponding optimal
assignment approach. In Figs. \ref{fig:mAP_optimal_score_rank} and
\ref{fig:AMR_optimal_score_rank}, by visual inspection, it
is observed that the ranks switch severely for both greedy and optimal
assignments approaches in a particular trial. However, in \Tab \ref{tab:optimal_rank_cons}
it is confirmed that the optimal assignment approach is more reliable
than the greedy counterpart.

\begin{figure*}
\centering{}\includegraphics[width=0.95\textwidth]{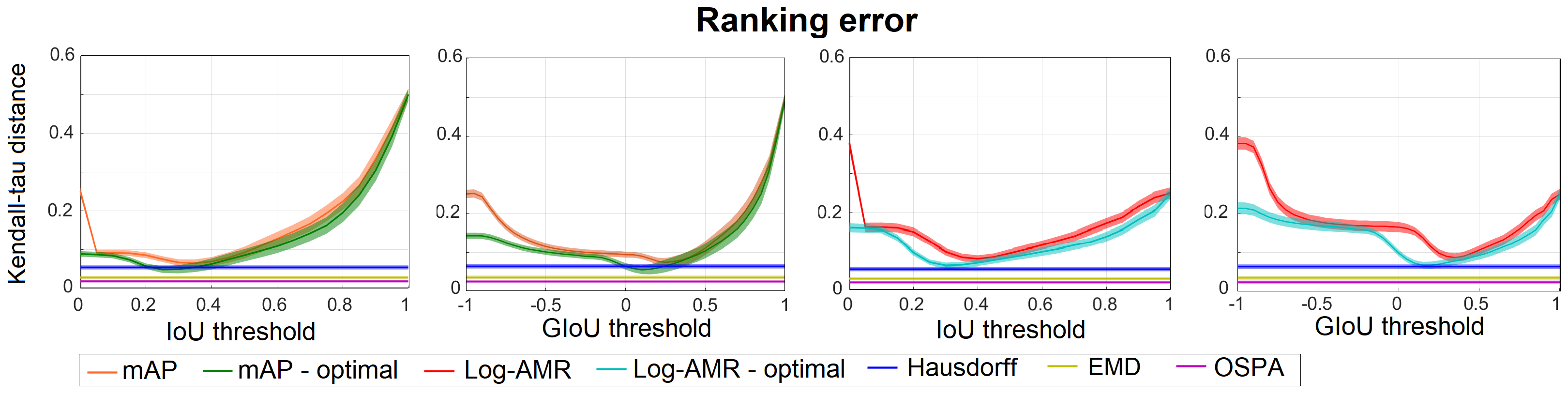}
\caption{Monte Carlo mean normalized Kendall-tau ranking errors
(from the true ranking) of various criteria at different thresholds,
in multi-class multi-object detection
test with greedy and optimal assignment approaches. Shaded area around
each curve indicates 0.2-sigma bound. We also show the results for
metric criteria for reference. \label{fig:optimal_Man} }
\end{figure*}

In terms of the meaningfulness of the ranks, the optimal assignment
approach is better than the greedy one in terms of ranking accuracy
as shown in \Fig \ref{fig:optimal_Man}. For the proposed approach,
while it is competitive to the Hausdorff metric, it is still less
meaningful than the EMD and OSPA metrics. \Tab \ref{tab:optimal_Man}
further confirms that optimal is better than greedy assignment approach
as it produces more meaningful ranking order. For both greedy and
optimal assignment approaches, the partial marginalization of thresholds does
not always produce more meaningful ranking order compared to the optimal
threshold. However, marginalizing over the whole range of thresholds seems to improve the ranking performance overall.

\begin{table*}
\caption{Monte Carlo means (and standard deviations) of the ranking consistency indicators
for mAP and log-AMR with greedy and optimal assignment approaches. \label{tab:optimal_rank_cons}}

\centering{}%
\begin{tabular}{ccccc}
\toprule 
\multicolumn{5}{c}{\textbf{\scriptsize{}mAP and Log-AMR with Optimal Assignment}}\tabularnewline
\midrule 
 & {\scriptsize{}$\textbf{mAP}_{\textbf{IoU}}$ } & {\scriptsize{}$\textbf{mAP}_{\textbf{IoU-optimal}}$ } & {\scriptsize{}$\textbf{mAP}_{\textbf{GIoU}}$ } & {\scriptsize{}$\textbf{mAP}_{\textbf{GIoU-optimal}}$}\tabularnewline
\midrule 
{\scriptsize{}$\boldsymbol{\overline{R_{S}}}$}  & {\scriptsize{}$8.13\thinspace(1.20)$} & \textbf{\scriptsize{}$\mathbf{7.37\thinspace(1.52)}$}{\scriptsize{} } & {\scriptsize{}$10.0\thinspace(1.31)$} & \textbf{\scriptsize{}$9.01\thinspace(1.66)$}\tabularnewline
\midrule 
{\scriptsize{}$\boldsymbol{\overline{R_{std}}}$}  & {\scriptsize{}$3.53\thinspace(0.664)$} & \textbf{\scriptsize{}$3.18\thinspace(0.816)$}{\scriptsize{} } & {\scriptsize{}$3.19\thinspace(0.534)$} & \textbf{\scriptsize{}$\mathbf{2.82\thinspace(0.657)}$}\tabularnewline
\midrule 
{\scriptsize{}$\boldsymbol{\overline{R_{Sen}}}$}  & {\scriptsize{}$4.38\thinspace(1.42)$} & \textbf{\scriptsize{}$3.88\thinspace(1.65)$}{\scriptsize{} } & {\scriptsize{}$3.86\thinspace(1.13)$} & \textbf{\scriptsize{}$\mathbf{3.39\thinspace(1.20)}$}\tabularnewline
\midrule 
 & {\scriptsize{}$\textbf{Log-AMR}_{\textbf{IoU}}$ } & {\scriptsize{}$\textbf{Log-AMR}_{\textbf{IoU-optimal}}$ } & {\scriptsize{}$\textbf{Log-AMR}_{\textbf{GIoU}}$ } & {\scriptsize{}$\textbf{Log-AMR}_{\textbf{GIoU-optimal}}$}\tabularnewline
\midrule 
{\scriptsize{}$\boldsymbol{\overline{R_{S}}}$}  & {\scriptsize{}$7.89\thinspace(1.09)$} & \textbf{\scriptsize{}$\mathbf{6.96\thinspace(1.30)}$}{\scriptsize{} } & {\scriptsize{}$10.4\thinspace(1.48)$} & \textbf{\scriptsize{}$9.07\thinspace(1.66)$}\tabularnewline
\midrule 
{\scriptsize{}$\boldsymbol{\overline{R_{std}}}$}  & {\scriptsize{}$3.25\thinspace(0.602)$} & \textbf{\scriptsize{}$2.73\thinspace(0.698)$}{\scriptsize{} } & {\scriptsize{}$3.27\thinspace(0.588)$} & \textbf{\scriptsize{}$\mathbf{2.70\thinspace(0.650)}$}\tabularnewline
\midrule 
{\scriptsize{}$\boldsymbol{\overline{R_{Sen}}}$}  & {\scriptsize{}$4.18\thinspace(1.48)$} & \textbf{\scriptsize{}$3.52\thinspace(1.64)$}{\scriptsize{} } & {\scriptsize{}$4.12\thinspace(1.40)$} & \textbf{\scriptsize{}$\mathbf{3.47\thinspace(1.40)}$}\tabularnewline
\bottomrule
\end{tabular}
\end{table*}
\begin{table*}

\caption{Monte Carlo means (and standard deviations) of normalized Kendall-tau
ranking errors of mAP and log-AMR with greedy and optimal assignment
approaches at certain thresholds. The subscripts of IoU/GIoU indicate
the threshold values; ``optimal'' threshold is the one with best
ranking accuracy; \textquotedblleft M-partial\textquotedblright{}
indicates that the evaluation is done via averaging the score/rate
over the range 0.5 to 0.95 in steps of 0.05. \textquotedblleft M-full\textquotedblright{}
indicates that the evaluation is done via averaging the score/rate
over the entire range of the base-measure (excluded two extreme thresholds).
We also show the results for metric criteria for reference. \label{tab:optimal_Man}}

\begin{centering}
\begin{tabular}{|c|cccc|cccc|}
\hline 
\multicolumn{9}{|c|}{\textbf{\scriptsize{}mAP and Log-AMR with Optimal Assignment: Normalized
Kendall-tau ranking error (in units of $10^{-2}$)}}\tabularnewline
\hline 
 & \textbf{\scriptsize{}$\boldsymbol{\text{IoU}_{0.5}}$}{\scriptsize{} } & {\scriptsize{}$\boldsymbol{\mathbf{\text{IoU}_{\text{optimal}}}}$}  & {\scriptsize{}$\boldsymbol{\mathbf{\text{IoU}_{\text{M-partial}}}}$}  & {\scriptsize{}$\boldsymbol{\mathbf{\text{IoU}_{\text{M-full}}}}$}  & {\scriptsize{}$\mathbf{\boldsymbol{\text{GIoU}_{0}}}$}  & {\scriptsize{}$\boldsymbol{\mathbf{\text{GIoU}_{\text{optimal}}}}$}  & {\scriptsize{}$\boldsymbol{\mathbf{\text{GIoU}_{\text{M-partial}}}}$} & {\scriptsize{}$\boldsymbol{\mathbf{\text{GIoU}_{\text{M-full}}}}$}\tabularnewline
\hline 
\textbf{\scriptsize{}mAP}{\scriptsize{} } & {\scriptsize{}$10.0\thinspace\left(8.90\right)$} & {\scriptsize{}$7.08\thinspace\left(5.56\right)$} & {\scriptsize{}$7.52\thinspace(8.71)$}  & {\scriptsize{}$3.62\thinspace(2.65)$}  & {\scriptsize{}$9.41\thinspace\left(3.81\right)$} & {\scriptsize{}$7.39\thinspace\left(5.51\right)$} & {\scriptsize{}$8.27\thinspace\left(8.71\right)$} & {\scriptsize{}$4.86\thinspace(3.00)$} \tabularnewline
\hline 
\textbf{\scriptsize{}mAP-optimal}{\scriptsize{} } & {\scriptsize{}$9.16\thinspace\left(9.23\right)$}  & {\scriptsize{}$5.36\thinspace\left(5.76\right)$}  & {\scriptsize{}$6.34\thinspace(8.80)$}  & {\scriptsize{}$\mathbf{2.31\thinspace(1.75)}$} & {\scriptsize{}$10.4\thinspace\left(9.50\right)$}  & {\scriptsize{}$5.45\thinspace\left(5.52\right)$} & {\scriptsize{}$6.87\thinspace\left(9.17\right)$} & {\scriptsize{}$\mathbf{3.23\thinspace\left(4.56\right)}$}\tabularnewline
\hline 
\textbf{\scriptsize{}Log-AMR}{\scriptsize{} } & {\scriptsize{}$9.91\thinspace\left(5.97\right)$} & {\scriptsize{}$8.42\thinspace\left(5.29\right)$} & {\scriptsize{}$6.75\thinspace\left(3.42\right)$} & {\scriptsize{}$4.31\thinspace(2.30)$}  & {\scriptsize{}$16.5\thinspace\left(6.57\right)$} & {\scriptsize{}$8.80\thinspace\left(5.46\right)$} & {\scriptsize{}$7.33\thinspace\left(3.69\right)$} & {\scriptsize{}$4.95\thinspace(2.83)$} \tabularnewline
\hline 
\textbf{\scriptsize{}Log-AMR-optimal}{\scriptsize{} } & {\scriptsize{}$8.64\thinspace\left(5.31\right)$}  & {\scriptsize{}$6.72\thinspace\left(4.55\right)$}  & {\scriptsize{}$5.64\thinspace\left(2.61\right)$}  & {\scriptsize{}$3.36\thinspace(1.55)$} & {\scriptsize{}$9.30\thinspace\left(5.33\right)$}  & {\scriptsize{}$6.78\thinspace\left(4.58\right)$}  & {\scriptsize{}$5.89\thinspace\left(2.70\right)$} & {\scriptsize{}$3.73\thinspace\left(1.81\right)$}\tabularnewline
\hline 
\textbf{\scriptsize{}Hausdorff}{\scriptsize{} } & \multicolumn{4}{c|}{{\scriptsize{}$5.43\thinspace\left(2.71\right)$}} & \multicolumn{4}{c|}{{\scriptsize{}$6.39\thinspace\left(2.88\right)$}}\tabularnewline
\hline 
\textbf{\scriptsize{}EMD}{\scriptsize{} } & \multicolumn{4}{c|}{{\scriptsize{}$2.80\thinspace\left(1.83\right)$}} & \multicolumn{4}{c|}{{\scriptsize{}$3.50\thinspace\left(2.26\right)$}}\tabularnewline
\hline 
\textbf{\scriptsize{}OSPA}{\scriptsize{} } & \multicolumn{4}{c|}{{\scriptsize{}$1.86\thinspace\left(1.64\right)$}} & \multicolumn{4}{c|}{{\scriptsize{}$2.41\thinspace\left(1.90\right)$}}\tabularnewline
\hline 
\end{tabular}
\par\end{centering}
\end{table*}
% \clearpage
\clearpage
\clearpage
\nobalance

\bibliographystyle{IEEEtran}
\bibliography{references/IEEEabrv,references/reflib,references/rebuttal}

\end{document}